\pdfoutput=1

\documentclass[11pt]{article}

\usepackage[]{acl}

\usepackage{times}
\usepackage{latexsym}

\usepackage[T1]{fontenc}

\usepackage[utf8]{inputenc}

\usepackage{microtype}

\usepackage{inconsolata}

\usepackage{graphicx}

\usepackage{microtype}
\usepackage{graphicx}
\usepackage{subfigure}
\usepackage{booktabs} 

\usepackage{hyperref}

\usepackage{amsmath}
\usepackage{amssymb}
\usepackage{mathtools}
\usepackage{amsthm}
\usepackage{subcaption} 
\usepackage[table,xcdraw]{xcolor}
\usepackage[normalem]{ulem}
\usepackage{tcolorbox}
\usepackage{caption}
\usepackage{graphicx}
\usepackage[utf8]{inputenc}
\usepackage[capitalize,noabbrev]{cleveref}
\usepackage{bbm}

\theoremstyle{plain}
\newtheorem{theorem}{Theorem}[section]

\theoremstyle{definition}
\newtheorem{definition}[theorem]{Definition}
\newtheorem{assumption}[theorem]{Assumption}

\newtheorem{claim}[theorem]{Argument}
\theoremstyle{remark}

\usepackage{tcolorbox}
\usepackage{varwidth}
\usepackage{lipsum} 
\usepackage[textsize=tiny]{todonotes}

\newcommand{\framework}{\mbox{\textsc{DiverseAgentEntropy}}}

\newcommand{\vpara}[1]{\noindent \textbf{#1 }}

\title{Rethinking LLM Uncertainty: A Multi-Agent Approach to Estimating Black-Box Model Uncertainty}

\author{
\textbf{Yu Feng}$^{1}$\thanks{Corresponding email: \texttt{fengyu1@seas.upenn.edu, hphu@amazon.com}. 
Work done during Yu's internship at AWS AI Labs, when all authors were employed there. 
Code available at \url{https://github.com/amazon-science/DiverseAgentEntropy}.} \quad
\textbf{Phu Mon Htut}$^{2}$ \quad
\textbf{Zheng Qi}$^{2}$ \quad
\textbf{Wei Xiao}$^{2}$ \quad
\textbf{Manuel Mager}$^{3}$ \\[6pt]
\textbf{Nikolaos Pappas}$^{2}$ \quad
\textbf{Kishaloy Halder}$^{2}$ \quad
\textbf{Yang Li}$^{2}$ \quad
\textbf{Yassine Benajiba}$^{4}$ \quad
\textbf{Dan Roth}$^{1,4}$\\[8pt]
$^1$University of Pennsylvania \quad
$^2$AWS AI Labs \\
$^3$Johannes Gutenberg University of Mainz \quad
$^4$Oracle AI
}

\begin{document}
\maketitle
\begin{abstract}
Quantifying uncertainty in black-box LLMs is vital for reliable responses and scalable oversight. Existing methods, which gauge a model’s uncertainty through evaluating self-consistency in responses to the target query, can be misleading: an LLM may confidently provide an incorrect answer to a target query, yet give a confident and accurate answer to that same target query when answering a knowledge-preserving perturbation of the query.
We systematically analyze the model behaviors and demonstrate that this discrepancy stems from suboptimal retrieval of parametric knowledge, often due to contextual biases that prevent consistent access to stored knowledge. We then introduce \framework{}, a novel, theoretically-grounded method employing multi-agent interaction across diverse query variations for uncertainty estimation of black-box LLMs. This approach more accurately assesses an LLM's true uncertainty and improves hallucination detection, outperforming existing self-consistency based techniques.
\end{abstract}
\section{Introduction} 
Large language models (LLMs) have shown remarkable abilities to encode and apply real-world knowledge through their learned parameters~\citep{yu2024kola}. However, when crucial knowledge is absent, unreliable, inaccurately stored, or simply not retrieved, even if embedded within the model, LLMs can produce hallucinations~\citep{10.1145/3571730,ravichander2025halogenfantasticllmhallucinations}. In high-stakes areas like medicine, finance, and law, such errors can have serious consequences, making uncertainty estimation as important as predictive accuracy. Looking ahead, the responsible development and deployment of increasingly capable AI systems will require scalable oversight~\citep{bowman2022measuringprogressscalableoversight}, alignment strategies that evolve with model complexity. While external assistance can support this goal, it is often costly and difficult to scale. Ideally, models should be able to assess their own uncertainty to support more autonomous alignment. As models grow more powerful yet remain prone to hallucinations~\citep{nananukul2024haloontologyrepresentingcategorizing}, it becomes increasingly important for users to identify which outputs are trustworthy. In practice, this challenge is compounded by the fact that most users interact with LLMs via APIs~\citep{claude3modelcard, openai2024gpt4technicalreport}, and by the growing prevalence of proprietary models, our focus turns to the black-box setting. This leads us to a central research question: \emph{How can we develop a robust methodology to quantify a model's uncertainty regarding its parametric knowledge, and further enable it to refrain from generating hallucinated responses, without internal model access or external assistance?}

\begin{figure*}[h]
\centering
\scalebox{0.985}{
	\includegraphics[width=1\textwidth]{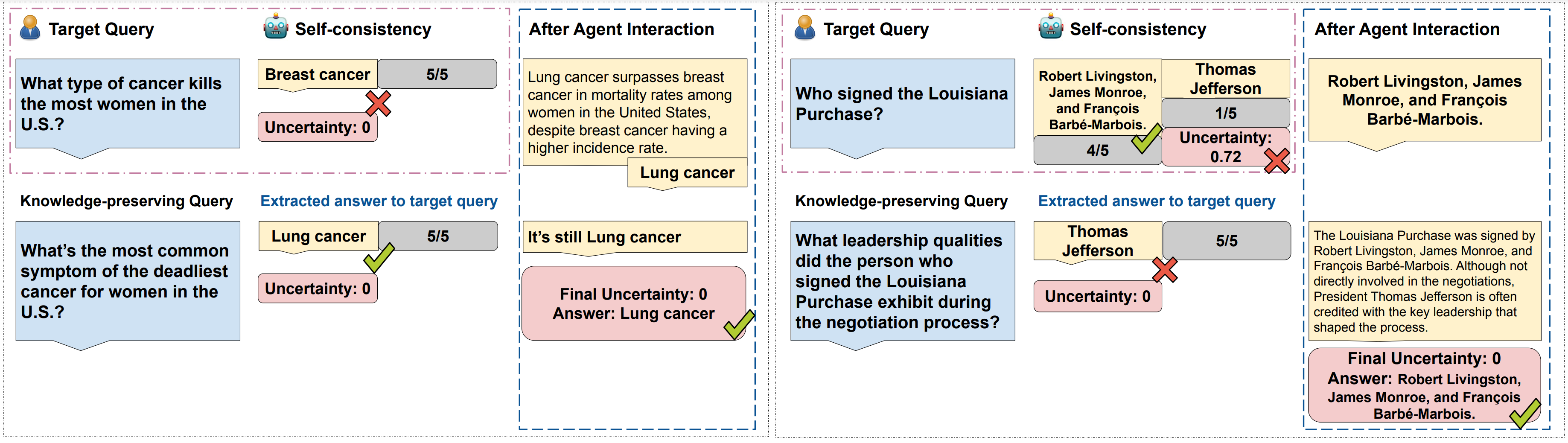}}
	\caption{\label{fig:intro_example} Black-box methods relying on self-consistency (pink box) misestimate model uncertainty due to a mismatch between the uncertainty estimated from the original target query and the model’s actual knowledge, while \framework{} (blue box) recovers true model uncertainty.}
\end{figure*}

Current research on black-box model uncertainty often focuses on evaluating self-consistency for a single target query~\citep{kuhn2023semantic, lin2024generatingconfidenceuncertaintyquantification, nikitin2024kernel}, calculating uncertainty by sampling multiple responses to the same target query and measuring entropy as uncertainty across semantically clustered responses. However, these approaches do not necessarily capture a model's true uncertainty. As in Fig.~\ref{fig:intro_example}, a model may exhibit low uncertainty for an incorrect answer when responding to the target query, while confidently providing an accurate answer to the same target query when answering varied knowledge-perserving queries that require the same underlying knowledge (left). It may also assign inaccurate uncertainty to correct answers (right).

We systematically demonstrate that the model failures shown in Fig.~\ref{fig:intro_example} arise from the suboptimal retrieval of parametric knowledge in LLMs, that is, their inconsistent ability to reliably access knowledge encoded in their parameters~\cite{zhang-etal-2023-sac3, zhao-etal-2024-knowing, chen2024two}. Our analysis further reveals that this limitation is driven by the model’s susceptibility to contextual biases: subtle changes in question, e.g., different phrasing or additional contextual information, can lead to divergent responses~\cite{guo2024biaslargelanguagemodels, gonen-etal-2023-demystifying, sclar2024quantifying}.

 To address this issue and demonstrate a proof of concept, we propose \framework{}, a theoretically grounded framework inspired by recent advances in multi-agent collaboration for factual accuracy~\cite{du2024improving, feng2024donthallucinateabstainidentifying}, self-revision in LLMs~\cite{kadavath2022languagemodelsmostlyknow, shinn2023reflexion} especially under contextual guidance~\cite{sun2023recitationaugmented}, and test-time scaling~\cite{snell2025scaling, muennighoff2025s1simpletesttimescaling}. \framework{} leverages controlled multi-agent interaction, treating diverse, knowledge-preserving variants of a target query, queried through the same model, as distinct agents. These agents collaboratively refine their responses to the target query, enabling more reliable uncertainty estimation. As illustrated in Fig.~\ref{fig:pipeline_example}, the interaction leads agents to converge on the correct answer. Uncertainty is then estimated via the weighted entropy of final responses, with an abstention policy applied when uncertainty is high.

In \S\ref{main_experiment}, we show that our uncertainty metric is more calibrated and  effectively detects hallucinations, outperforming self-consistency-based black-box methods. It achieves higher AUROC, with notable gains on long-tailed and realistic benchmarks (+3.8\% PopQA, +7.5\% FreshQA, +5.2\% TruthfulQA). Sampling across abstention rates also improves accuracy on known questions by 2.5\%. 
The major contributions of our paper are:
\begin{itemize}
    \item Identify and explain a key failure mode in black-box LLM uncertainty estimation caused by inconsistent retrieval of parametric knowledge due to contextual biases (\S\ref{analysis}).
    \item Propose a theoretically grounded framework leveraging multi-agent interaction on diverse, knowledge-preserving queries to better recover true model uncertainty (\S\ref{method} and \S\ref{method_1}).
    \item Demonstrate improved uncertainty estimation and reduced hallucinations, outperforming self-consistency baselines across multiple benchmarks (\S\ref{main_experiment}).
\end{itemize}
\vspace{-9pt}
\section{Related work}
\vpara{Uncertainty Estimation of LMs.} Recent work has quantified LLM uncertainty using entropy over multiple sampled outputs~\citep{shorinwa2024surveyuncertaintyquantificationlarge, kuhn2023semantic_1, kuhn2023semantic, yadkori2024believebelievellm, lin2024generatingconfidenceuncertaintyquantification, aichberger2024how}, but these methods rely solely on self-consistency with the original query, which can be misleading (Fig.~\ref{fig:intro_example}). Some studies explore verbalizing uncertainty~\citep{madaan2023selfrefine, tian-etal-2023-just, xiong2024can}, though \citet{xiong2024can} find LLMs tend to be overconfident. Others examine uncertainty through model activations~\citep{chen2024inside, ch-wang-etal-2024-androids}, which is not feasible in our black-box setting.

\vpara{Consistency Evaluation of LMs.}
While \citet{wang2023selfconsistency} and \citet{manakul-etal-2023-selfcheckgpt} show that self-consistency and sampling aid reasoning and fact-checking, others highlight limitations. \citet{zhang-etal-2023-sac3} and \citet{zhao-etal-2024-knowing} argue that detecting hallucinations requires consistency across semantically equivalent questions. \citet{chen2024two} further show that LLMs often fail to maintain compositional consistency. 

\vpara{Agent interaction for LMs.}Recent works~\citep{xiong-etal-2023-examining, du2024improving, feng2024donthallucinateabstainidentifying} improve factuality in LMs through multi-agent cooperation or debate, primarily using cross-model agents with no self-correction. 
\section{Background on LLM Uncertainty Estimation}
\label{existing_uncertainty}
We aim to quantify the uncertainty of black-box LLMs for 
a target query $q$.
Following~\citet{kuhn2023semantic_1}, we evaluate uncertainty by
treating uncertainty estimation as the problem of predicting whether to trust an answer the model generates.
For a given model $\theta$, for any query $q, p_\theta(y \mid q), y \in \mathcal{Y}$ is the probability of output $y$ given the query $q$ over \emph{all possible} full responses $\mathcal{Y}$ of the query. 
We then provide background on uncertainty estimation, focusing on entropy-based evaluation, as uncertainty is commonly measured by the entropy of predictions in the existing literature~\citep{wellmann2012uncertainties, abdar2021review}. 
For an input query $q$ and output space $\mathcal{Y}$, the total uncertainty for a given model $\theta$ can be approximated by the entropy of the output distribution~\citep{malinin2021uncertainty}:
\begin{equation}
U(q) = H(\mathcal{Y}|q) \approx -\sum_{y_i \in C} p(y_i |q) \log p(y_i |q),
\label{eq_main_approx}
\end{equation}
where $C$ represents all grouped semantically different answers  when a model is queried $n$ times with the same query $q$ . $y_i$ is one possible semantically different answer for $q$. Lower entropy indicates higher model confidence.

Existing self-consistency-based uncertainty estimation methods in the black-box setting~\citep{kuhn2023semantic_1,kuhn2023semantic, lin2024generatingconfidenceuncertaintyquantification, aichberger2024how, nikitin2024kernel}
follow a similar procedure: 1) For a given input $q$, generate $N$ response samples. 2) Calculate similarity scores for these $N$ responses. 3) Compute an uncertainty estimate $U (q)$ using the similarity scores. We explain existing methods in detail in Appendix~\ref{appendix:other_workds}. The model's uncertainty about the answer to a target query $q$ thus depends entirely on the responses sampled specifically for that query. 

\section{\framework{}}
\label{proposed_uncertainty}
\begin{figure*}[t]
\centering
\scalebox{1}{
	\includegraphics[width=0.98\textwidth]{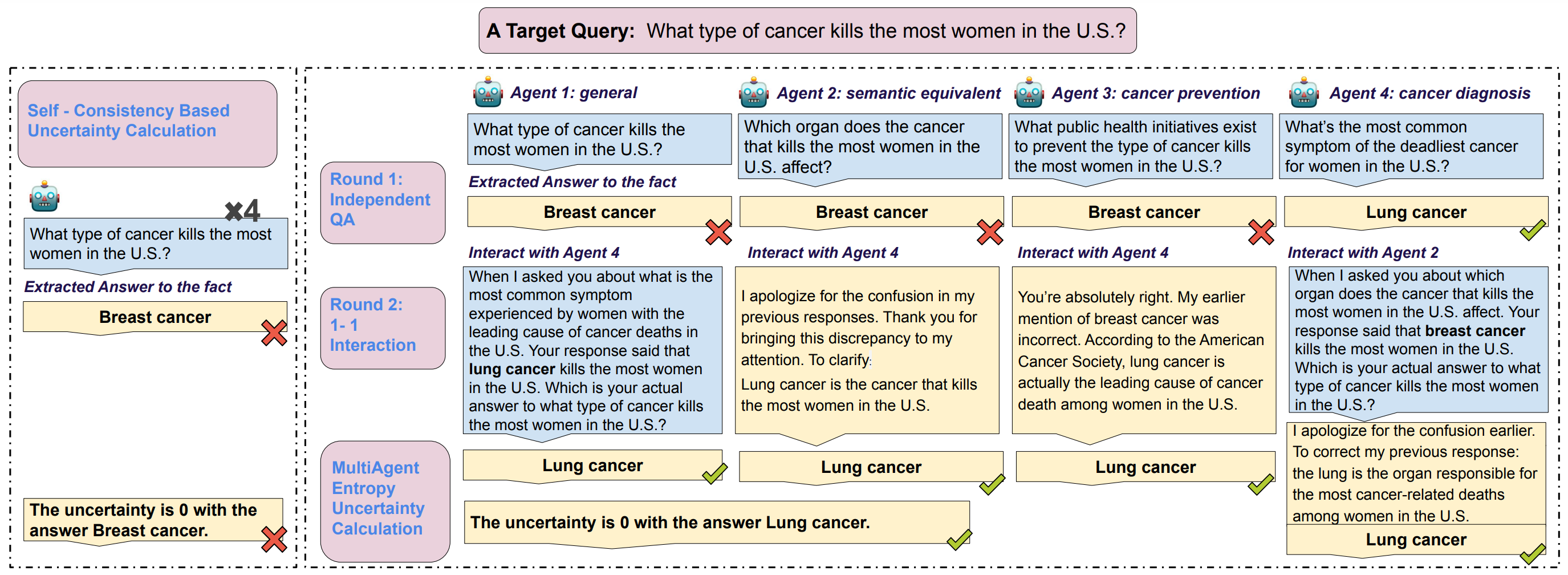}}
	\caption{\label{fig:pipeline_example} \framework{} estimates model uncertainty by enabling multi-agent interactions on diverse knowledge-perserving queries, analyzing uncertainty based on these interactions rather than simple self-consistency.}
\end{figure*}
In this section, we show that black-box methods relying on self-consistency misestimate model uncertainty due to a mismatch between uncertainty estimated from the original query and the model’s actual knowledge (Fig.~\ref{fig:intro_example}). This stems from \emph{suboptimal retrieval of parametric knowledge}~\cite{zhang-etal-2023-sac3, zhao-etal-2024-knowing, chen2024two} due to \emph{susceptibility to contextual biases}~\cite{guo2024biaslargelanguagemodels}. To address this, we introduce a theoretically grounded same-model multi-agent framework to better recover true uncertainty.

\subsection{Preliminaries}
We begin by presenting basic definitions, e.g., knowledge-preserving perturbations in the paper.

\begin{definition}[Target knowledge Derivation]
We define a knowledge function $\mathcal{K}_{\text{target}}: \mathcal{X}\to \Omega$ that maps the query $q$ to the set of target knowledge required for a \emph{complete} answer to $q$. 
\end{definition}
\begin{definition}[Knowledge-Preserving Perturbation of the Target Query]\label{kp-definition}
We say a transformation $T: \mathcal{X}\to \mathcal{X}$ is a 
\emph{knowledge-preserving perturbation} with respect to the target query $q$: if letting $q' = T(q)$, we have $\mathcal{K}_{\text{target}}(q) \in \mathcal{K}_{\text{target}}(q')$, i.e., $q'$ still requires the knowledge of the target query $q$ to answer. Here the perturbation could result in 1) semantically equivalent version of $q$, 2) diverse compositional questions of $q$ which still require the same target knowledge as shown in Fig.~\ref{fig:pipeline_example}. 
\end{definition}

\begin{definition}[Induced Distribution for Target Query $q$]
Let $\mathcal{Y}_{\text{target}}$ be the set of all possible semantically different outcomes for the target query $q$. Define a \emph{extraction} function $
  \phi: \mathcal{Y'} \to \mathcal{Y}_{\text{target}}$ which extracts the answer to the target query $q$ from any full response $y\in\mathcal{Y'}$ given a updated query $q'$. 
A single $y$ might contain multiple contents.  The function $\phi(y)$ extracts only those parts relevant to $q$. 
If $y$ does not address the target knowledge at all, $\phi(y)=\varnothing$. 
We then define an \emph{induced} distribution for the target query $q$ given an updated query $q'$:
\begin{equation}
  \tilde{p}_\theta(z \mid q')
  \;=\;
  \sum_{\,y \in \mathcal{Y'} : \phi(y) = z\,}
  p_\theta(y \mid q'),
  \quad
  z \,\in\, \mathcal{Y}_{\text{target}}.
\end{equation}
Here $\tilde{p}_\theta(z \mid q')$ 
is the total probability that given an updated query $q'$, the LLM's extracted answer from a response $y$ to the target query $q$ is $z$.
\end{definition}

\subsection{Model
Retrieval Challenges and Self-consistency Limitations}
\label{analysis}

We formalize the model’s behavior when it knows the answer to the target query with the following assumption. This helps explain the model’s unsatisfactory retrieval of parametric knowledge under contextual biases.

\begin{assumption}[Fixed True Distribution Under Perfect Knowledge Assumption]
\label{claim:distribution}
As Ground truth independence assumption in~\citet{yadkori2024believebelievellm}, for a target query $q$, define $q^*=\mathcal{K}_{\text{target}}(q)$ as the target knowledge for the query and $z^*$ as the ground truth answer. If the model knows the answer to the target query, the model has a fixed true induced distribution $\tilde{p}_\theta(\cdot \mid q^*)$ as the internal belief of the target knowledge.\footnote{
$q^*$ should be seen as the model's representation of the target knowledge.}
\end{assumption}

\begin{claim}[Distributional Shifts from Contextual Biases]
\label{claim:distribution-shift}
Assume the model knows the answer to the target query (knowledge) and consider a query after \emph{a knowledge-preserving perturbation} $q' = T(q)$ that fails to output the ground truth answer $z^*$\footnote{Here $q$ could also be considered as a knowledge-preserving perturbation about the target knowledge $q^*$.}, even though \(q'\) relies on the same target knowledge, the resulting distribution $\tilde{p}_\theta(\cdot \mid q')$ shifts from $\tilde{p}_\theta(\cdot \mid q^*)$. 
Especially, $\tilde{p}_\theta(z^* \mid q') = \tilde{p}_\theta(z^* \mid q^*) + \epsilon_{x'}(z^*)$, where $\epsilon_{x'}(z^*)$ is the (possibly negative) bias for the ground truth answer. 
Hence, small changes in grammar, formats, or additional contextual information~\cite{guo2024biaslargelanguagemodels, gonen-etal-2023-demystifying, sclar2024quantifying} induce a shift in the distribution of the target knowledge and thus the model fails to consistently retrieve the same response
as in Fig.~\ref{fig:intro_example}. A detailed empirical analysis supporting the argument can be referred to in \S~\ref{exp:retrievability}.
\end{claim}

This also illustrates why self-consistency alone is insufficient for accurately assessing a model's uncertainty regarding the target query: relying solely on a single formulation of the query can lead to contextual biases, resulting in distributional shifts that obscure the model’s true uncertainty.

\subsection{Remedy with Multi-Agent Interaction}
\label{method}

Prior work shows that LLMs improve with revision and diverse context~\citep{kadavath2022languagemodelsmostlyknow, shinn2023reflexion, sun2023recitationaugmented}. Inspired by multi-agent methods~\citep{xiong-etal-2023-examining, du2024improving, feng2024donthallucinateabstainidentifying} to improve factual correctness, we propose a theoretically grounded same-model interaction approach, using knowledge-preserving query perturbations to simulate diverse agents and better recover the model’s true uncertainty.

\begin{claim}[Behaviors of LLMs under Multi-agent Interaction]
\label{claim:behavior}
For a given target query \( q \), the model exhibits one of two behaviors under multi-agent interaction: 1) \textbf{Known answer (knowledge)}: If the model internally represents a fixed, true distribution \( \tilde{p}_\theta(\cdot \mid q^*) \) for the target query (as in Assumption~\ref{claim:distribution}), then, after sufficient interactions, it can recover the correct answer distribution. 2) \textbf{Unknown answer}: If no such fixed distribution exists, interaction leads to divergent responses with no convergence (36\%), convergence to ``I don't know,’’ (54\%) or a wrong answer (10\%)\footnote{The percentage is based on the behaviors under \framework{} of 50 random sampled queries where the model definitely doesn't know the answer. Details in \S\ref{appendix:pilot}.}. In most cases, both the model's interaction dynamics and final output make its lack of knowledge evident.

\end{claim}

We now describe the multi-agent interaction process under diverse queries and show that, when the model knows the answer to the target query, the method can recover the model’s true uncertainty.

\vpara{Diverse Query Generation.} We use \emph{knowledge-preserving perturbations} defined in Definition~\ref{kp-definition} to construct diverse agent profiles. Let \( Q = \{ q_1, q_2, \dots, q_n \} \) be a set of \( n \) \emph{queries}, where \( q_1 = q \) is the original target query and \( \{ q_2, \dots, q_n \} \) are generated via perturbation functions \( \{ T_i \} \). We aim to maximize query diversity; details of the automated generation process ensuring diversity are provided in \S\ref{sec:implementation}. 

\vpara{Controlled Multi-Agent Interaction.} We instantiate \( n \) agents from the same model, where each agent \( A_j \) independently answers a unique query \( q_j \), which serves as its individualized contextual background. We facilitate multiple rounds of controlled one-on-one interaction between agents to help refine their answers to the target query. Each query \( q_i \in Q \) is associated with an \emph{induced belief distribution} \( \tilde{p}^{\,r}_\theta(z \mid q_i) \) at round \( r \), where \( z \in \mathcal{Y}_{\text{target}} \). 
At each round, agent \( A_i \) interacts with another agent \( A_j \) whose belief distribution is most diverse from its own. \( A_i \) is shown query \( q_j \) and its corresponding answer distribution, with \( A_i \)'s own interaction history. Agent \( A_i \) then updates its belief using an internal update function \( f_i \), implicitly defined by the model. Formally:
\begin{equation}
    \tilde{p}^{\,r+1}_\theta(\cdot \mid q_i)
  \;=\;
  f_i\Bigl(
    \tilde{p}^{\,r}_\theta(\cdot \mid q_i),\,
    \tilde{p}^{\,r}_\theta(\cdot \mid q_j)
  \Bigr). 
\end{equation}

\begin{assumption}[Connected Graph]
\label{assump:connected}
Let $G \;=\; (Q, E)$ be a directed graph capturing which queries interact over the rounds.
An edge \( (q_i, q_j) \in E \) indicates that in some round \( r \), the distribution \( \tilde{p}^{\,r+1}_\theta(\cdot \mid q_i) \) was updated using \( \tilde{p}^{\,r}_\theta(\cdot \mid q_j) \).  
We assume \( G \) is \emph{connected} over \( R \) rounds (large \( R \)).
\end{assumption}

\begin{assumption}[Expected Pairwise Contractive Update Property]
\label{assump:contractive-updates} Under perfect knowledge assumption (Scenario 1 in Argument~\ref{claim:behavior}) and Assumption~\ref{assump:connected}, let $\tilde{p}^{\,r}_\theta(z \mid q_j)$ be the belief distribution for query $q_j$ at round $r$. We have $\tilde{p}^{\,r}_\theta(z \mid q_j) = \tilde{p}_\theta(z \mid q^*) + \epsilon_{j}^{r}(z)$. Let $R$ be the total number of update rounds. Suppose there exists a finite constant $C$ (independent of $R$) such that 
\emph{at most} $C$ rounds are ``bad''---in these bad rounds, 
the expected divergence from the true distribution may increase or fail to decrease. In all other (``good'') rounds $r \notin \{\text{bad}\}$, when $D\bigl(
        \tilde{p}_\theta(\cdot \mid q^*),\;
        \tilde{p}^{\,r}_\theta(\cdot \mid q_j))\neq 0$, the expected update is 
\emph{contractive} with respect to the true distribution:
\begin{equation}
\begin{split}
    \mathbb{E}\bigl[
      D\bigl(
        \tilde{p}_\theta(\cdot \mid q^*),\;
        \tilde{p}^{\,r+1}_\theta(\cdot \mid q_j)
      \bigr)
    \bigr]
    \;<\; \\
    \mathbb{E}\bigl[
      D\bigl(
        \tilde{p}_\theta(\cdot \mid q^*),\;
        \tilde{p}^{\,r}_\theta(\cdot \mid q_j)
      \bigr)
    \bigr].
\end{split}
\end{equation}
\end{assumption} 
Here, \( D(\cdot, \cdot) \) denotes a suitable divergence measure (e.g., KL divergence). We validate this assumption in \S\ref{appendix:assump}. Fig.~\ref{fig:num_rounds} also shows that each interaction round aligns the results more closely with the true distribution, improving performance.

\begin{theorem}[Convergence to True Distribution When Known]\label{Theorem} Assume the model knows the answer to the target query, then the final weighted distribution satisfies:
\begingroup
\setlength{\abovedisplayskip}{7pt}
\setlength{\belowdisplayskip}{3pt}
\begin{equation}
\begin{aligned}
\lim_{\substack{n \to \infty \\ r \to \infty}}\sum_{j=0}^n w_j \tilde{p}^{\,r}_\theta(z \mid q_j) 
  &= \tilde{p}_\theta(z \mid q^*), \\
\forall z &\in \mathcal{Y}_{\text{target}}.
\end{aligned}
\end{equation}
\endgroup
\end{theorem}
where $\sum_{j=0}^n w_j=1$. The detailed proof sketch can be referred to in \S\ref{appendix:proof}. We will first use Assumption~\ref{assump:contractive-updates} to prove that we have $\lim_{r \to \infty}\mathbb{E}[\epsilon_{j}^r(z)] = 0 $ 
and each $\epsilon_j^{r}(z)$ has a bounded variance, where $\mathbb{E}[\epsilon_j^{r}(z)]$ is the \emph{expected} error of that specific answer $z$ 
at round $r$ for query $j$. Since the queries are sampled diverse,  $\epsilon_j^{r}(z)$ are weak dependent, putting these conditions together with the law of large number,
\begingroup
\setlength{\abovedisplayskip}{7pt}
\setlength{\belowdisplayskip}{7pt}
\begin{equation}
\begin{split}
\lim_{\substack{n \to \infty \\ r \to \infty}}\sum_{j=0}^n w_j \tilde{p}^{\,r}_\theta(z \mid q_j) 
&= \\ \underbrace{\sum_{j=0}^n w_j}_{=1} \tilde{p}_\theta(z \mid q^*)
&\quad + \underbrace{\sum_{j=0}^n w_j \epsilon_j^{r}(z)}_{\to 0}.
\end{split}
\end{equation}
\endgroup

\subsection{Proposed Metric of \framework{}}
\label{method_1}
Building on the aforementioned theory, we propose \framework{} as an empirical implementation of the multi-agent interaction and corresponding uncertainty estimation. An illustrative example is shown in Fig.~\ref{fig:pipeline_example}.

\vpara{Step 1: Diverse Query Generation.}  Let $Q \;=\; \{\,q_1, \dots, q_n\}$
be a set of \(n\) \emph{queries} obtained via \emph{knowledge-preserving perturbations} $\{T_i\}$ of a target query \(q\) and $q_1 = q$. The set of Q will include the target query q itself, semantically equivalent questions of the target query, and questions about different perspectives as shown in Fig.~\ref{fig:pipeline_example}.

\vpara{Step 2: Multi-Agent Interaction.} 
We create \( n \) agents from the same tested model, where each agent \( A_j \) independently answers \( q_j \). We then conduct controlled one-on-one cross-play interactions using fixed prompts (Fig.~\ref{fig:pipeline_example}). To ensure practicality, each query is answered once per interaction, and we approximate the belief distribution using the top-1 (most probable) distribution. In each round, agent \( A_j \) selects a peer whose answer to the target query differs for interaction, prioritizing agents it has not yet interacted with. During interaction, \( A_j \) sees its own conversation history and is given the peer agent’s query and most recent response. Based on this, \( A_j \) decides whether to retain, or revise from its answer. The interaction ends when: (1) all agents agree on an answer, (2) each agent maintains its answer for two consecutive rounds, or (3) the maximum of \( R^* \) rounds is reached.

\vpara{Step 3: Uncertainty Score Calculation.} In practice, with limited interaction rounds, agents vary in response reliability. We assign each agent \( A_j \) a weight \( w_j \) in the final probability based on how often it changes its answer to the target query. Frequent changes indicate lower reliability and result in a lower weight.
\begingroup
\setlength{\abovedisplayskip}{3pt}
\setlength{\belowdisplayskip}{7pt}
\begin{equation}
w_j = \frac{R-r_j + 1}{\sum_{j=1}^n(R-r_j + 1)}.
\end{equation}
\endgroup
where $j=1,..., n$. We denote $R$ as the final total number of interaction rounds and $r_j$ as the number of rounds where the agent $A_j$ changes its answer during the interaction. We apply Laplace smoothing to avoid zero weight. Denote by $ \mathbbm{1}\{A_j=z_i\}  $ whether $ z_i $ is the final answer to the target query of the agent $A_j$ after the interaction. Therefore,
\begingroup
\setlength{\abovedisplayskip}{3pt}
\setlength{\belowdisplayskip}{5pt}
\begin{equation}
\tilde{p}_\theta(z_i \mid q^*) = \sum_{j=1}^n w_j \mathbbm{1}\{A_j=z_i\}.
\label{eq_prob}
\end{equation}
\endgroup
We then apply Eq.\ref{eq_main_approx} with Eq.\ref{eq_prob} to calculate the final uncertainty as \framework{}. We present all the prompts for the  method in \S\ref{appendix:prompt}.

\vpara{Step 4: Score-based Abstention Policy.} 
We introduce an abstention policy with a threshold parameter: the model abstains if the uncertainty score exceeds the threshold (see \S\ref{appendix:threshold} for method variants) or if the mode response is ``I don't know.'' If the policy does not abstain, the answer with the highest estimated probability is returned.

\section{Experiment}
\label{main_experiment}

\subsection{Experiment Setting}
\label{appendix:experiment_setting}
\vpara{Evaluation Models.} We evaluate on Llama-3-70b-Instruct~\citep{llama3modelcard} and Claude-3-Sonnet~\citep{claude3modelcard} through API calls.

\vpara{Datasets.} See \S\ref{appendix:data} for a detailed description of the datasets. \textbf{Entity-centric QA}: we randomly sample from PopQA
~\citep{mallen-etal-2023-trust} for 1) PopQA popular with popular entities and 2) PopQA less popular with less popular entities. \textbf{False assumption QA}: 3) FalseQA~\citep{hu-etal-2023-wont}. All questions in the dataset contain false assumptions. \textbf{General QA}: 4) TruthfulQA~\citep{lin-etal-2022-truthfulqa}. 5) FreshQA~\citep{vu2023freshllmsrefreshinglargelanguage}.

\vpara{Metrics.}
We evaluate entropy-based uncertainty methods using AUROC, which measures how well uncertainty scores distinguish correct from incorrect outputs. A higher AUROC indicates better performance (1.0 is perfect; 0.5 is random chance).

To assess whether abstention reduces hallucinations, we adopt the setup from \citet{lin-etal-2022-truthfulqa}, evaluating if accuracy improves when the model chooses to answer. We report: \textbf{accuracy}, the percentage of correct responses among the questions where the model does not abstain; \textbf{abstention rate}, the percentage of questions where the model abstains; \textbf{correctness score}, the percentage of correct responses across all questions; and \textbf{truthfulness score}, the percentage of correct or abstained responses across all questions. We further analyze the accuracy–recall (AR) trade-off across methods and datasets, where recall is defined as \( \text{recall} = 1 - \text{abstention rate} \).

\vpara{Baselines.} We first adopt a verbalization baseline, P(True)~\citep{madaan2023selfrefine}. We then adopt five entropy-based black-box uncertainty estimation baselines in \S\ref{existing_uncertainty}  
and the model is prompted to answer 5 times. 2)  Self-consistency with SemanticEntropy (SC SE)~\citep{kuhn2023semantic}. Three baselines with affinity graph~\citep{lin2024generatingconfidenceuncertaintyquantification}: 2) SC with Eccentricity (SC Ecc). 3) SC with the Degree Matrix (SC Degree). 4) SC with Eigenvalues (SC EigV). 5) SC with kernel and von
Neumann entropy (SC KLE)~\citep{nikitin2024kernel}.

We adopt seven baselines for hallucination detection. \textbf{Greedy-based baselines}: 1) Greedy. 2) Self-Refine~\citep{madaan2023selfrefine}. 3) Self-evaluation w many samples~\citep{kadavath2022languagemodelsmostlyknow}. A total of five answers, including the greedy one, are generated, after which the model assesses the greedy sample's validity. 4) Multiple-Recite~\citep{sun2023recitationaugmented}. The model generates multiple related paragraphs from its parametric knowledge before answering the question.
\textbf{Sampling-based baselines:} 5) Self-consistency (SC)~\citep{wang2023selfconsistency}: the model answers the query 5 times, and we accept the majority answer or abstain if no answer appears at least 3 times. 6) Consistency with semantically equivalent questions (SeQ)~\citep{zhang-etal-2023-sac3, zhao-etal-2024-knowing}: the model answers 5 semantically equivalent questions about the same target query. 7) Consistency with diverse questions (DiverseQ): the model answers 5 diverse queries about the same target query. Note that we evaluate the semantic equivalence of answers to cluster responses for all sampling-based baselines.

\textbf{\framework{}  Variants.} We introduce two variants with 5 agents: 1) ${\rm Agent}_L$: abstains if the uncertainty score exceeds the entropy of three answers with probability (0.6, 0.2, 0.2). 2) Agent: a stricter vote, abstains if the score exceeds the entropy of two answers (0.6, 0.4). Details are explained in \S\ref{appendix:threshold}.

\begin{table}[t]
    \centering
    \setlength{\tabcolsep}{8pt}
    \scalebox{0.75}{
    \begin{tabular}{lcc}
        \toprule
        Dataset & Claude-3-Sonnet & Llama-3-70b-Instruct \\
        \midrule
        PopQA pop        & 0.114 & 0.118 \\
        PopQA less pop   & 0.193 & 0.207 \\
        FalseQA              & 0.154 & 0.154 \\
        TruthfulQA           & 0.296 & 0.330 \\
        FreshQA              & 0.167 & 0.175 \\
        \bottomrule
    \end{tabular}
    }
    \caption{The average percentage of incorrect answers to the target query in the first round without agent interaction, evaluated only on instances where all agents agree on the correct answer after interactions.} 
    \label{tab:retriv_error}
\end{table}

\subsection{Evaluation for  Retrieval of Parametric Knowledge}

\label{exp:retrievability}
\vpara{Unsatisfying retrieval of parametric knowledge.} LLMs often fail to reliably and consistently retrieve encoded knowledge. We show that even when the model knows the correct answer, it may initially respond inconsistently across contexts. We examine cases where all agents converge on the same ground truth answer after interaction, signaling that the knowledge is present but not reliably accessed. We conduct a quantitative analysis of the model’s initial performance by measuring the average error rate on target queries in the first round before agent interaction. This captures how often the model fails to retrieve the correct answer initially, before any interaction. As shown in Table~\ref{tab:retriv_error}, LLMs are not consistently reliable, with at least one in five answers incorrect on average (19.1\%). The issue is more pronounced on long-tailed (PopQA less popular) and realistic datasets (FreshQA, TruthfulQA).

\begin{figure*}[t]
    \centering
    \scalebox{0.55}{
    \begin{minipage}{0.8\textwidth}
        \centering
        \includegraphics[width=\linewidth]{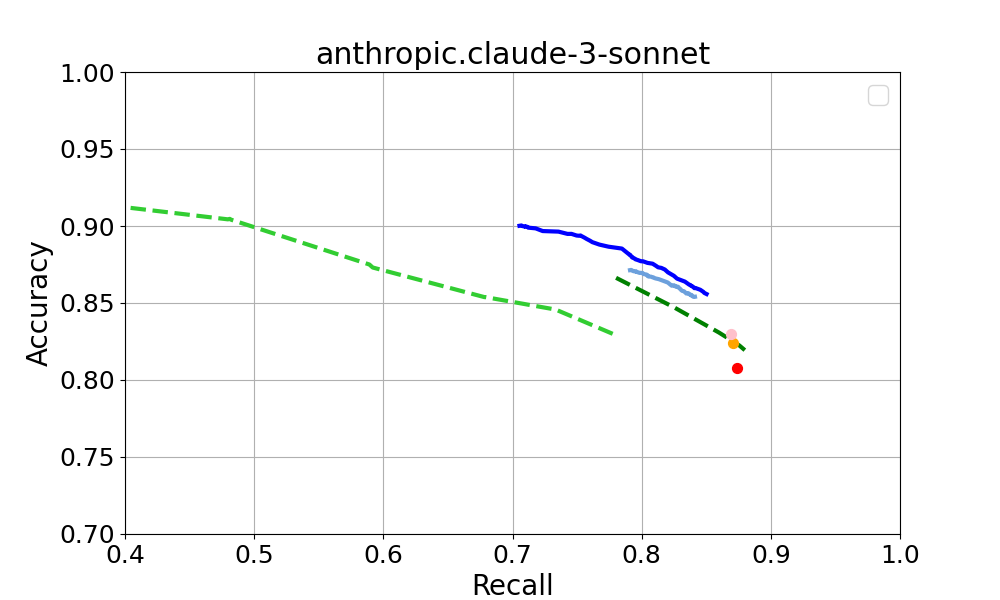}
    \end{minipage}\hfill
    \begin{minipage}{0.8\textwidth}
        \centering
        \includegraphics[width=\linewidth]{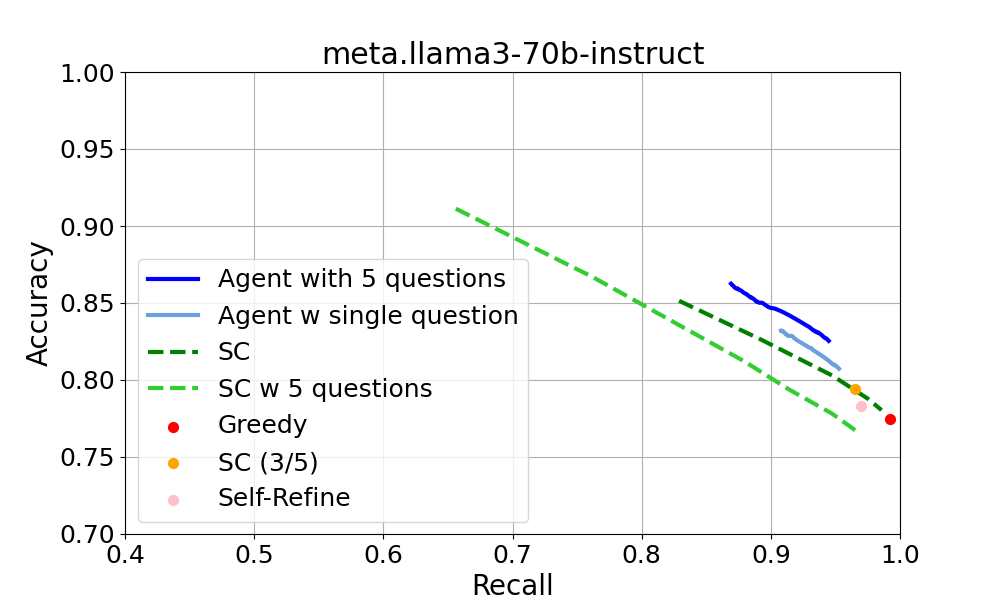}   
    \end{minipage}\hfill}
    \caption{AR-curves across all data. SC refers to SC (SE). SC w 5 questions refers to calculating entropy using the agents' diverse questions without agent interaction. We present the \underline{best} and \dotuline{sub optimal} methods for each method category in Tables~\ref{tab:auroc} and \ref{tab:main}. 
    }
    \label{fig:pr_curves}
\end{figure*}

\begin{table}[t]
\centering
\small
\scalebox{0.8}
{
\begin{tabular}{l@{\hspace{3.5pt}}c@{\hspace{3.5pt}}c@{\hspace{3.5pt}}c@{\hspace{3.5pt}}c@{\hspace{3.5pt}}c@{\hspace{3.5pt}}c}
\toprule
Model & PopQA\_p & PopQA\_l & FalseQA & FreshQA & TruthfulQA & All \\
\midrule
\multicolumn{7}{c}{\textit{Claude-3-Sonnet}} \\
\midrule
P ({\rm True}) &  0.686&0.660 &\dotuline{0.800} &\dotuline{0.719} &0.556  &0.766 \\
SC (Ecc)     & 0.671 & 0.821 & 0.711 & 0.702 & 0.548 & 0.766 \\
SC (Degree)   & 0.674 & 0.855& 0.713 & 0.704 & 0.550 & 0.771 \\
SC (EigV)     & 0.673 & 0.851& 0.713 & 0.703 & 0.550 & 0.771 \\
SC (KLE) & 0.684 & 0.883 &0.756 & 0.692 & 0.563 & 0.791 \\
\dotuline{SC (SE)}       & \dotuline{0.693}& \dotuline{0.887} & 0.753 & 0.694 & \dotuline{0.568} &  \dotuline{0.792} \\
\underline{Agent}        & \textbf{0.725} & \textbf{0.947} & \textbf{0.802} & \textbf{0.836} & \textbf{0.624} & \textbf{0.833} \\
\midrule
\multicolumn{7}{c}{\textit{Llama-3-70b-Instruct}} \\
\midrule
P ({\rm True})&0.618& 0.641 & \textbf{0.698} &  0.642&  0.542 &0.601 \\
SC (Ecc)      & 0.594& 0.716 & 0.628 & 0.660 & 0.488 & 0.644 \\
SC (Degree)  & 0.595 & 0.704 & 0.629 & 0.662 & 0.486 & 0.645 \\
SC (EigV)    & 0.595& 0.707  & 0.629 & \dotuline{0.664} & 0.486 & 0.645 \\
\dotuline{SC (KLE)} & \dotuline{0.632} & \dotuline{0.738} &  0.664& 0.632 & 0.537 &\dotuline{0.694} \\
\dotuline{SC (SE)}     & 0.624 & 0.737 & \dotuline{0.673} & 0.632 & \dotuline{0.545} &  \dotuline{0.694} \\
\underline{Agent}        & \textbf{0.651} & \textbf{0.753} & \dotuline{0.673} & \textbf{0.697} & \textbf{0.592} & \textbf{0.713} \\
\bottomrule
\end{tabular}
}
\caption{Comparison of AUROC scores across different QA datasets. We highlight the \underline{\textbf{best}} and \dotuline{suboptimal} methods.Our method is more calibrated.}
\label{tab:auroc}
\end{table}
\vpara{Models' susceptibility to contextual biases results in the poor retrieval.} We conduct a qualitative analysis by sampling 45 instances from the same pool.
We observe that the model tends to generate a different response, even when it knows the correct answer to the target knowledge, under these conditions: 1) 42\% of cases occur when changes in the varied query significantly diverges in wording, 2) 22\% when an incorrect answer is more popular within the context of the target knowledge, and 3) 20\% when the additional context is more closely associated with a different possible answer to the target knowledge. Examples of each scenario are provided in \S\ref{appendix:Retrivability_error}. These findings highlight the need for systematic research into contextual biases.

\subsection{Evaluation of \framework{}}

\vpara{\framework{} is more calibrated than self-consistency-based uncertainty estimation.}
Table~\ref{tab:auroc} reports AUROC scores across different methods. Our method achieves the highest AUROC, indicating better calibration. Notably, it shows larger gains on long-tailed data (+3.8\% average on less popular PopQA) and realistic benchmarks (+7.5\% on FreshQA, +5.2\% on TruthfulQA) compared to the best baseline, with especially strong performance on the more advanced Claude model. These results highlight \framework{}’s promise for high-stakes scenarios and scalable oversight.

\vpara{\framework{}-based abstention policy effectively detects hallucinations.} Table~\ref{tab:main} shows that our method more effectively identifies model hallucinations. It achieves the highest correctness and truthfulness scores. Dataset-level results are provided in \S\ref{appendix:hallucination_numbers}. Fig.~\ref{fig:pr_curves} presents Accuracy-Recall (AR)-curves across all data with per-dataset results in Appendix Fig.~\ref{fig:appnedix_main}. Our method consistently outperforms all baselines, achieving the highest accuracy at comparable recall levels on non-abstained answers (+ 2.5\% average increase). Appendix Table~\ref{tab:wrong_correct} shows that on average, 18.0\%/ 27.0\% of incorrect instances were completely corrected (\textbf{Correct}), while only 2.5\%/8.9\% of correct instances became partially incorrect (\textbf{Wrong}) for Claude/Llama. This highlights that \framework{} significantly improves accuracy with minimal degradation and detailed analysis is in \S\ref{appendix:hallucination_numbers}.

\begin{table}[t]
    \centering
    \scalebox{0.65}{
    \begin{tabular}{l@{\hspace{5pt}}c@{\hspace{5pt}}c@{\hspace{5pt}}c@{\hspace{5pt}}c@{\hspace{5pt}}c@{\hspace{5pt}}c@{\hspace{5pt}}c@{\hspace{5pt}}c}

        \toprule
        & \multicolumn{4}{c}{Claude-3-Sonnet} & \multicolumn{4}{c}{Llama-3-70b-Instruct} \\
        \cmidrule(lr){2-5} \cmidrule(lr){6-9}
        Method & Acc & Ab-R & Correct & TruthF & Acc & Ab-R & Correct & TruthF \\
        \midrule
        \dotuline{Greedy}                              & 0.808 & 0.126 & 0.707 & 0.832 & 0.775 & 0.008 & 0.769 & 0.777 \\
        \underline{Self-Refine}              &    0.826&  0.131 & \dotuline{0.718}&   0.849  & 0.783 & 0.030 & 0.760 & 0.790 \\
        Self-Eval  & 0.814 & 0.141 & 0.700 & 0.840 & 0.754 & 0.020 & 0.739 & 0.759 \\
        MultiRecite &0.779& 0.114&  0.690&  0.804&0.715&  0.010 &0.708& 0.717 \\
         \midrule
        \dotuline{SC (3/5)}              & 0.823 & 0.129 & 0.717 & 0.846 & 0.794 & 0.035 & 0.766 & 0.801 \\
        SeQ    &   0.815& 0.149&  0.693&  0.842 &0.818  &0.084& 0.749&  0.833 \\
        \underline{DiverseQ}  & 0.858 & 0.342 & 0.564 & \dotuline{0.906} & 0.811 & 0.121 & 0.713 & 0.834 \\
         \midrule
         Agent\_L                  & 0.852 & 0.142 & \textbf{0.731} & 0.873 & 0.826 & 0.055 & \textbf{0.780} & \dotuline{0.835} \\
        Agent                  & 0.883 & 0.216 & 0.692 & \textbf{0.908} & 0.841 & 0.084 & \dotuline{0.770} & \textbf{0.854} \\
        \bottomrule
    \end{tabular}}
    \caption{Performance evaluation of different methods on all data with the best setup for each method. Acc is accuracy. Ab-R is abstention rate. Correct is correctness score. TruthF is truthfulness score.}
    \label{tab:main}

\end{table}

\subsection{Analysis of \framework{}}
\vpara{Both diverse query generation and agent interaction are key for performance boost.} 
In Fig.~\ref{fig:pr_curves}, we compare \framework{} with two ablations: using self-consistency with 5 questions without agent interaction (SC with 5 questions) and agent interaction with only the target query (Agent w single question). The results show agent interaction and diverse query generation are important. 

\vpara{The quality of the diverse queries.} We evaluate question generation quality using Llama-3.1-70b-Instruct by sampling 500 queries per model (20 instances per dataset). Queries are assessed for 1) knowledge preservation, whether they required answering the target query, and 2) diversity, whether any two queries from the same instance differed by at least 20\% in wording. Claude and Llama achieve 95.0\% and 95.6\% in knowledge preservation, and 97\% and 96\% in diversity, respectively. These results demonstrate high-quality and varied question generation across models. Details in \S\ref{appnedix:format}.

\vpara{The number of agents.} In Appendix Fig.~\ref{fig:num_agents} and Appendix Fig.~\ref{fig:appnedix_num_agent}, we increase the number of agents, limiting interactions to 4 rounds. Performance improves with more agents but gains become smaller beyond 4 agents for Llama model, suggesting 5 agents are sufficient practically.

\vpara{The rounds of interactions.} As shown in Appendix Fig.~\ref{fig:num_rounds} and Appendix Fig.~\ref{fig:appnedix_num_round}, with the number of agents fixed at 5, increasing the rounds of interaction generally leads to improved performance.

\vpara{Format of agent interaction.}
We examine whether agents should engage in one-on-one interaction or group interactions, where each agent can view the unique questions and answers of all other agents. Our findings in Appendix Fig.~\ref{fig:interaction_format} and Appendix Fig.~\ref{fig:appnedix_interaction} indicate that one-on-one interactions outperform group interactions. We present a human error analysis in \S~\ref{appnedix:format_human}.

\vpara{Robustness of agent interaction.}
We evaluate agent susceptibility to misleading signals when one agent consistently gives either the most plausible incorrect answer or repeatedly responds with "I don't know". Appendix Fig.~\ref{fig:interaction_robustness} and Fig.~\ref{fig:appnedix_perturb} show that performance slightly declines in both cases, indicating sensitivity to consistent misleading input, while a better model, e.g., Claude, is more robust. 
\section{Conclusion}
Accurately determining the uncertainty of black-box LLMs is vital for reliable responses. We show that existing methods fall short because models suffer from suboptimal retrieval of parametric knowledge due to susceptibility to contextual bias. In this paper, we propose a novel method with a theoretical framework, \framework{}, to quantify an LLM's uncertainty after multi-agent interaction. Our method overcomes the limitations of self-consistency-based uncertainty estimation and delivers superior performance in detecting hallucinations.

\section*{Limitations}
We acknowledge that our method is more resource-intensive, primarily due to the use of multi-agent interactions, which, like similar approaches in the literature, naturally incur higher computational costs. A detailed analysis is provided in \S\ref{appendix:cost}. Importantly, our method more accurately recovers the model’s true uncertainty, a crucial advantage especially in high-stakes applications where reliability and correctness are prioritized.

Our proof is developed under relatively strict assumptions, which are primarily intended to provide intuition and can be relaxed or extended in future work.

We only evaluate two representative models with different base capacities due to the budget constraint, given the scale of our ablation studies. Although not the latest versions, they offer a meaningful comparison and consistently demonstrate the same trends observed across our ablation studies.

While prior work on LLM uncertainty estimation largely adopts the same evaluation setting as ours to analyze short-form, query-based uncertainty, we recognize the need to explore more complex scenarios in future work. We view the current setting as a necessary foundation for addressing more challenging tasks that require specialized study beyond the scope of this paper, such as: (1) long-form QA, which is difficult to evaluate due to challenges in assessing semantic equivalence and the limitations of current evaluation techniques, particularly in black-box settings; and (2) response-based uncertainty~\cite{qiu2024semantic}, which builds upon query-based uncertainty in black-box scenarios and thus depends on its accurate estimation (with the exception of verbalization, which is often unreliable). Moreover, quantifying response-based uncertainty becomes more difficult when query-based uncertainty is high, indicating that the model lacks reliable knowledge about the query, in which case none of the candidate responses should be trusted.

For further direction of broader impact, we believe our method has the potential to generalize to retrieval-augmented generation (RAG) settings. Its ability to generate questions from diverse perspectives and leverage multi-agent interactions makes it well-suited for effectively integrating heterogeneous information sources. In the RAG context, generating perspective-varied questions encourages the retriever to surface a broader range of documents, thereby enriching the informational basis for downstream tasks. Moreover, the multi-agent interaction mechanism enables self-reflection and deliberation, which is particularly valuable in QA scenarios involving ambiguous or conflicting retrieved content.

\section*{Acknowledgments}
This work was conducted during Yu Feng’s internship at Amazon Web Services (AWS) and was fully funded by AWS. We sincerely appreciate the valuable feedback provided by the AWS team and extend our gratitude to the ARR reviewers and editors for their insightful and constructive comments.

\bibliography{acl_latex.bbl}
\bibliographystyle{acl_natbib}
\clearpage
\appendix
\section{Appendix}
\subsection{Existing Self-consistency Based Uncertainty Quantification Methods}
\label{appendix:other_workds}

Existing self-consistency-based uncertainty estimation methods in the black-box setting~\citep{kuhn2023semantic_1,kuhn2023semantic, lin2024generatingconfidenceuncertaintyquantification, aichberger2024how, nikitin2024kernel}
follow a similar procedure: 1) For a given input $q$, generate $N$ response samples. 2) Calculate  similarity scores for these $N$ responses. 3) Compute an uncertainty estimate $U (q)$ using the similarity values. 

Specifically, \citet{kuhn2023semantic} introduces semantic entropy to calculate $p(y_i |q)$ in Eq.\ref{eq_main_approx} as a frequency-based probability derived from repeated sampling of the target query $q$. Assume we find the semantic clusters for the sampled answers, and let each query return a possible semantically different answer $y_i \in C$. The count of times a particular $y_i$ appears as the output for the input $q$ over these $N$ queries is denoted as $c(y_i)$. Therefore, $p(y_i|q) = \frac{c(y_i)}{n}
\label{frequency_entropy}$.

\citet{lin2024generatingconfidenceuncertaintyquantification} calculates uncertainty using a weighted adjacency graph built upon semantic affinities. An affinity model $e$ maps pairs of responses to values in $[0,1]$. Given $n$ independent samples, the model induces a symmetric adjacency matrix $W = [w_{i,j}]_{i,j=1}^n$, where $w_{i,j}$ is the mean of the pairwise similarities between response $i$ and $j$. 
The degree matrix is $D = [\mathbbm{1}[j=i]\sum_{a=1}^n w_{n,j}]_{i,j=1}^n$, and the Laplacian $L = I - D^{-1/2}WD^{-1/2}$ has eigenvalues $\{\lambda_n\}_{a=1}^n$. The following uncertainty measures are then defined:
$U_{EigV}(x) = \sum_{a=1}^n \max\{0, 1 - \lambda_a\}, U_{Degree}(x) = 1 - \frac{trace(D)}{n^2}, U_{Ecc}(x) = \|[v_1, v_2, \dots, v_n]\|_2$ where $\{v_a\}_{a=1}^n$ are vectors associated with $L$.

\citet{nikitin2024kernel} proposes Kernel Language Entropy (KLE), which quantifies uncertainty using the von Neumann entropy. After generating $N$ response samples, KLE constructs a positive semidefinite, unit-trace kernel matrix $K$ that encodes the semantic similarities between these responses. This kernel captures fine-grained semantic dependencies. The uncertainty is then calculated as the von Neumann entropy of this kernel matrix, given by $U_{\text{KLE}}(q) = -\text{Tr}(K \log K)$ where $\text{Tr}$ denotes the matrix trace and the logarithm of the matrix is defined through its eigendecomposition.

\subsection{Observation of Agent Behaviors When Knowledge is Unknown }\label{appendix:pilot}
We  conduct a pilot experiment using \framework{} to observe agent behaviors for the scenario when the model doesn't know the answer and therefore the model's true distribution $\tilde{p}_{\theta}(\cdot \mid q^*)$ does not exist. We randomly sample 50 false-assumption queries where the model definitely doesn't know the answer from FalseQA and freshQA. We observe three distinct behaviors after agent interaction\footnote{Note that the behaviors are consistent on both Claude and Llama.}:

\begin{enumerate}
  \item \textbf{Arbitrary Shifts and Inconsistency (36\%)}  Even after many interaction rounds, queries $q_i$ and $q_j$ maintain 
    \emph{inconsistent} distributions 
    $\tilde{p}_{\theta}^{\,R}(\cdot \mid q_i)$ 
    and 
    $\tilde{p}_{\theta}^{\,R}(\cdot \mid q_j)$,
    disagreeing about which answer is most plausible. 
  Instead, they \emph{shift arbitrarily} depending on interaction's context. 

  \item \textbf{Trivial Convergence to ``I don't know''(54\%).} 
    After several rounds of disagreement, all queries $q_i$ converge to ``I don't know''(or similarly agnostic) answers, indicating the model's inability to pin down the knowledge. 
    
   \item \textbf{Convergence to An Arbitrarily Wrong Answer Distribution(10\%).} 
    After several rounds of interaction, a query's strong belief may enforce all the queries to converge to an arbitrarily wrong answer distribution. 

\end{enumerate}

\begin{figure*}[t] 
    \centering
    \scalebox{1}{
    \begin{minipage}{0.32\textwidth}
        \centering
        \includegraphics[width=\linewidth]{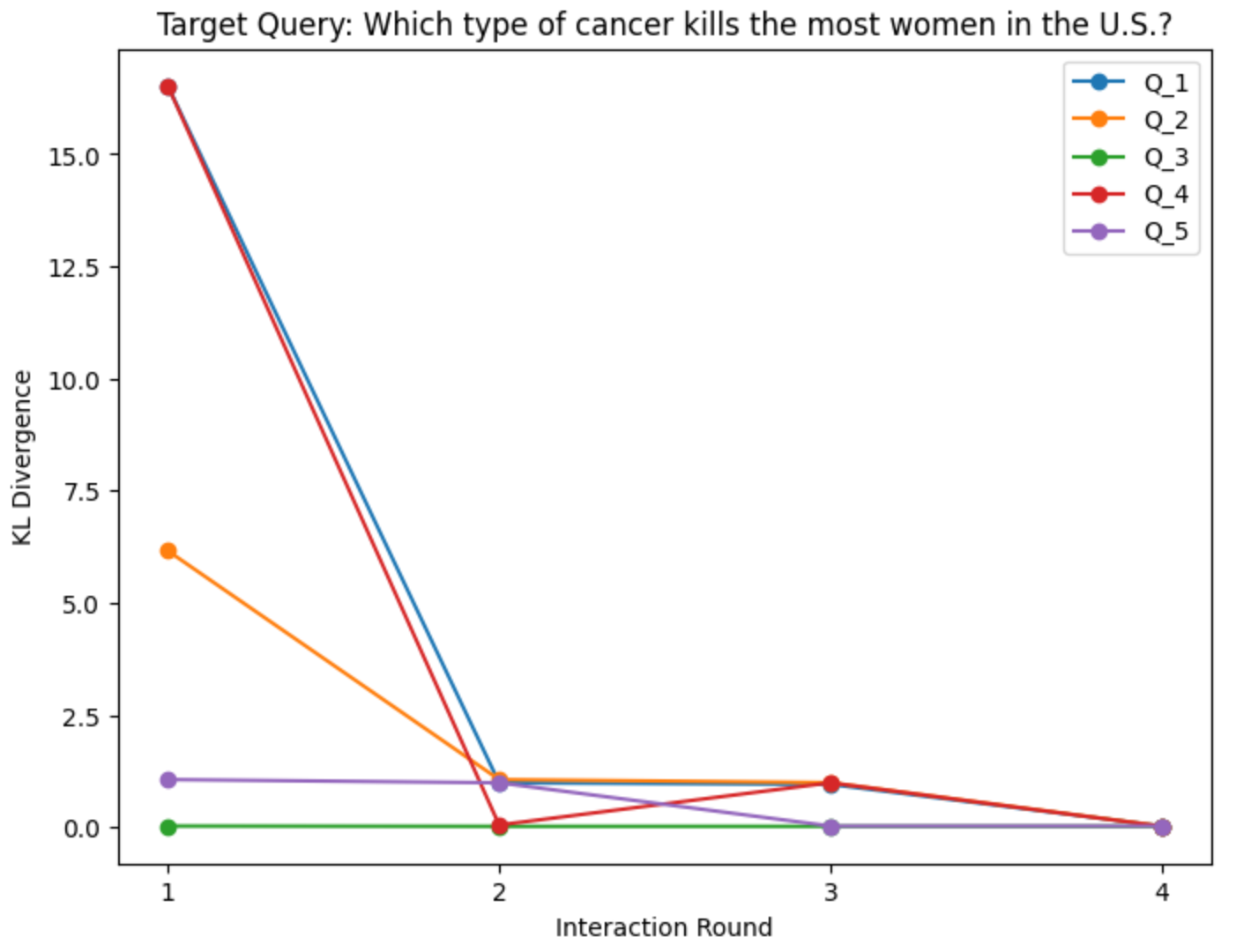}
    \end{minipage}\hfill
    \begin{minipage}{0.32\textwidth}
        \centering
        \includegraphics[width=\linewidth]{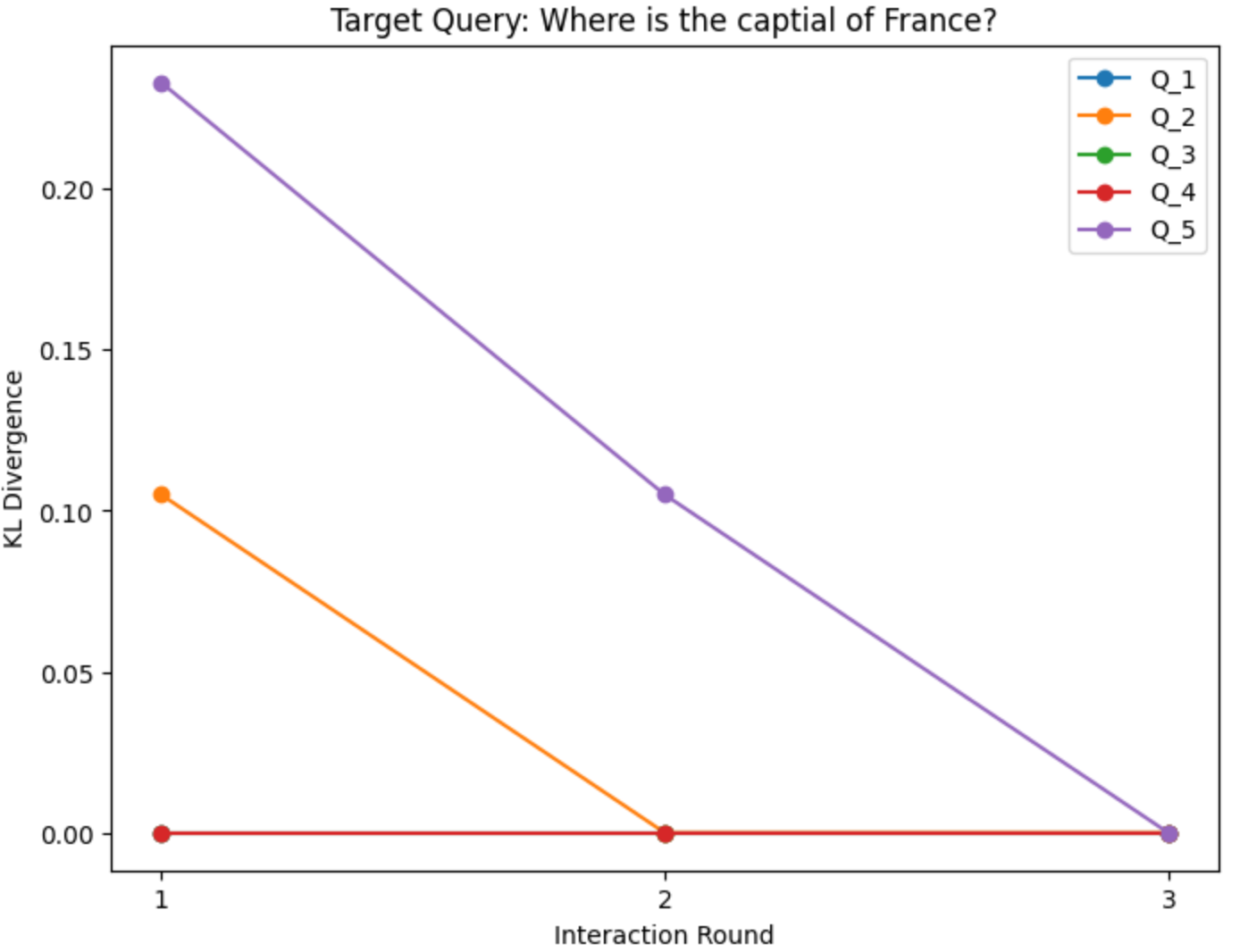}
    \end{minipage}\hfill
    \begin{minipage}{0.32\textwidth}
        \centering
        \includegraphics[width=\linewidth]{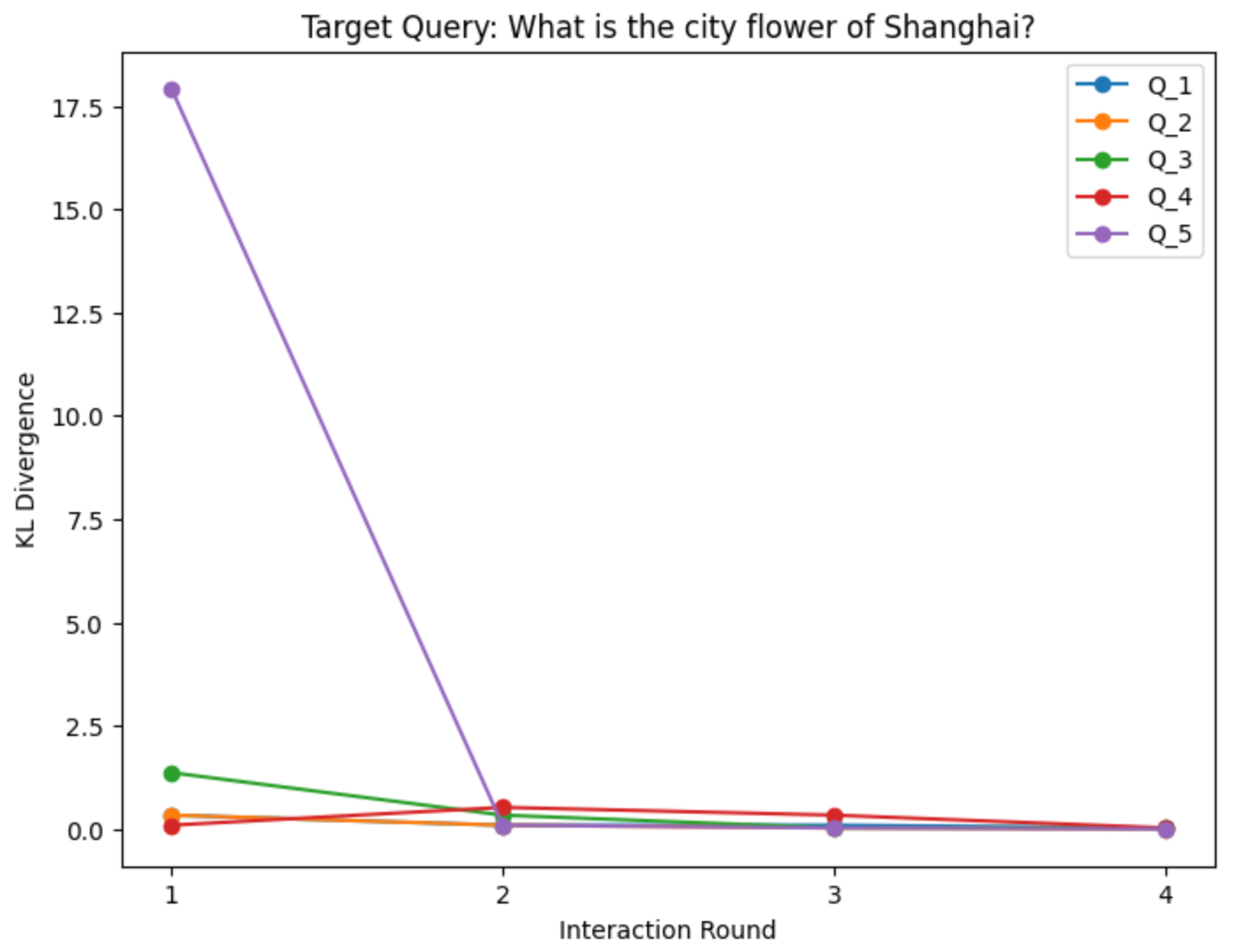}   
    \end{minipage}\hfill
    }
    \caption{KL-Divergence with regard to interaction rounds for three examples where the model knows the answer. }
    \label{fig:assumption}
\end{figure*}

\subsection{Implementation of Diverse Query Generation}
\label{sec:implementation}
Given a target query $x$, we use the same model to be tested to generate varied knowledge-perserving queries that require knowledge of the target query, ensuring that these questions are both representative and comprehensive. The question-generation process is completely automated and the detailed question-generation prompts can be found in \S\ref{appendix:prompt}. Specifically, we first conceptualize the target query and then sample various perspectives to ensure a comprehensive understanding. For each perspective, we generate $m$ questions that build upon the target query, tailored to that particular perspective. We filter these generated questions to ensure they strictly require knowledge of the target query to answer while avoiding the inclusion of the direct answer. We also generate $m$ semantically equivalent questions for the target query. 

We select $n$ questions from the generated pool to form the final candidate set $Q$ for the agents. In order to maximize diversity, the optimal set includes the target query $x$, one semantically equivalent question, and $n-2$ questions each targeting a unique perspective. If there are insufficient unique perspectives with qualified questions, we repeat the perspective question selection process to select from existing perspectives. If not, we supplement with additional semantically equivalent questions.

\subsection{Analysis of Assumption~\ref{assump:contractive-updates}}
\label{appendix:assump}
We now give a detailed explanation for the assumption. Interacting queries provide diverse contextual hints through varying background knowledge~\cite{sun2023recitationaugmented}. Transformer-based models, as shown by~\cite{wang2024a}, perform in-context gradient descent to enable dynamic self-correction. If the model knows the answer, each update is expected to \emph{reduce} the gap between $\tilde{p}^{r}$ and the true distribution $\tilde{p}^{*}$. While some bad updates may strongly increase divergence, the overall trend is an improvement. In later rounds, under Assumption~\ref{assump:connected}, connectivity allows queries to influence one another, propagating and averaging information. This prevents errors from persisting or amplifying~\cite{du2024improving}.  Thus, over $R$ rounds, at most a constant number $C$ of updates may fail to reduce divergence.

We test whether the agent’s interactions guide the KL divergence toward convergence with the (approximated) true distribution when the model knows the answer. We select 45 examples for evaluation from Experiment~\ref{exp:retrievability}. We allow 4 rounds of interactions. For each interaction round, we sample 10 times with a temperature of 0.7 to estimate the agent distribution. Manual inspection of the interaction update trend figures shows that 89\% of these examples consistently converge toward the true distribution. We also test interactions using the top-1 distribution, where all examples converge to the ground truth answer. To better illustrate this behavior, we present three representative examples in Appendix Figure~\ref{fig:assumption}.

For query (left): What type of cancer kills the most women in the U.S.? 
Q1: What type of cancer kills the most women in the U.S.?	Q2: Which organ does the cancer that kills the most women in the U.S. affect?	Q3: List the types of cancer and their impact on women in the U.S. based on mortality rates	
Q4: What public health initiatives exist to prevent and treat the type of cancer kills the most women in the U.S.?	
Q5: What’s the most common symptom of the deadliest cancer for women in the U.S.?
Gold distribution for Llama-3-70b-Instruct (approximated): Lung cancer: 0.75, Breast cancer: 0.2, Colorectal Cancer: 0.05.

For query (middle): Where is the capital of France? 
Q1:  Where is the capital of France? 
Q2: Which is the capital city of France?
Q3: How does the capital of France influence its Gen Z culture?
Q4: How has the location of the capital of France influenced its technological advancements?
Q5: How does the history of France shape its capital city? Gold distribution for Claude-3-sonnet (approximated): Paris: 1.

For query (right): What is the city flower of Shanghai?
Q1: What is the city flower of Shanghai?
Q2: Which bloom is honored as the city flower of Shanghai?
Q3: What role does the city flower of Shanghai play in the city's efforts to promote ecological conservation and environmental sustainability?
Q4: In what ways has the city flower of Shanghai been incorporated into the city's branding and tourism marketing strategies?
Q5: How does the city flower of Shanghai reflect the city's historical and cultural connection to the Yangtze River Delta region? Gold distribution for Llama-3-70b-Instruct (approximated):White Magnolia: 0.8, White Peony: 0.2.

\subsection{Theorem Proof Sketch for Theorem~\ref{Theorem}}
\label{appendix:proof}
Since\footnote{The proof is developed under relatively strict assumptions and is primarily intended to provide intuition.} $\tilde{p}^{\,r}_\theta(z \mid q_j) = \tilde{p}_\theta(z \mid q^*) + \epsilon_j^r(z)$, the KL divergence becomes:
\begin{equation}
\begin{split}
D_{\text{KL}}\bigl(\tilde{p}_\theta(\cdot \mid q^*) \parallel \tilde{p}^{\,r}_\theta(\cdot \mid q_j)\bigr)
= \\ \sum_{z \in \mathcal{Y}_{target}} \tilde{p}_\theta(z \mid q^*) 
\log\left(\frac{\tilde{p}_\theta(z \mid q^*)}{\tilde{p}^{\,r}_\theta(z \mid q_j)}\right).
\end{split}
\end{equation}

We can further decompose the KL divergence into: 
\begin{equation}
\begin{split}
D_{\text{KL}}\bigl(\tilde{p}_\theta(\cdot \mid q^*) \parallel \tilde{p}^{\,r}_\theta(\cdot \mid q_j)\bigr)
= \\\sum_{z} \tilde{p}_\theta(z \mid q^*) \log \tilde{p}_\theta(z \mid q^*)\\
- \sum_{z} \tilde{p}^{\,}_\theta(z \mid q^*) \log \tilde{p}^{\,r}_\theta(z \mid q_j).
\end{split}
\end{equation}
For the logarithmic term $
\log\left(\tilde{p}^{\,r}_\theta(z \mid q_j)\right)$, using a Taylor expansion of \(\log(x)\) around \(x_0 = \tilde{p}_\theta(z \mid q^*)\), we have:
\begin{equation} 
\begin{split}
\log\left(\tilde{p}^{\,r}_\theta(z \mid q_j)\right)
\approx \log\left(\tilde{p}_\theta(z \mid q^*)\right) + \\ \frac{\epsilon_j^r(z)}{\tilde{p}_\theta(z \mid q^*)} -\frac{1}{2}(\frac{\epsilon_j^r(z)}{\tilde{p}_\theta(z \mid q^*)})^2.
\end{split}
\end{equation}
We only assume the second-order approximation and therefore,
\begin{equation} 
\begin{split}
\mathbb{E}[D_{\text{KL}}\bigl(\tilde{p}_\theta(\cdot \mid q^*) \parallel \tilde{p}^{\,r}_\theta(\cdot \mid q_j)\bigr)]
\approx \frac{1}{2} \sum_{z} \frac{\mathbb{E} [ \epsilon_j^r(z)^2]}{\tilde{p}_\theta(z \mid q^*)}
\\ z \in \{z|\tilde{p}_\theta(z \mid q^*) \neq 0\}
\end{split}
\end{equation}

Let \(\Delta_j(r+1)= \mathbb{E}\bigl[
      D\bigl(
        \tilde{p}_\theta(\cdot \mid q^*),\;
        \tilde{p}^{\,r+1}_\theta(\cdot \mid q_j)
      \bigr)
    \bigr] - \mathbb{E}\bigl[
      D\bigl(
        \tilde{p}_\theta(\cdot \mid q^*),\;
        \tilde{p}^{\,r}_\theta(\cdot \mid q_j)
      \bigr)
    \bigr] \), therefore:
\begin{equation} 
\begin{split}
    \Delta_j(r+1) = \sum_{z} \tilde{p}_\theta(z \mid q^*) \mathbb{E}\bigl[\log \tilde{p}^{\,r}_\theta(z \mid q_j)] \\- \sum_{z} \tilde{p}_\theta(z \mid q^*) \mathbb{E}\bigl[\log \tilde{p}^{\,r+1}_\theta(z \mid q_j)]
\end{split}
\end{equation}
Substituting this approximation back into the \(\Delta_j(r+1)\), we get:
\begin{equation} 
\begin{split}
\Delta_j(r+1) = \frac{1}{2} \sum_{z} \frac{\mathbb{E} [\epsilon_j^{r+1}(z)^2 - \epsilon_j^r(z)^2]}{\tilde{p}_\theta(z \mid q^*)}.
\end{split}
\end{equation}
Here, \(\epsilon_j^r(z)\) can be positive, negative, or zero for different rounds \(r\). Since \(\tilde{p}_\theta(z \mid q^*) \in [\eta, 1]\) and \(\sum_{z}\tilde{p}_\theta(z \mid q^*)=1\), \(\epsilon_j^r(z)\) is bounded as \(\epsilon_j^r(z) \in \bigl[-\tilde{p}_\theta(z \mid q^*), 1 - \tilde{p}_\theta(z \mid q^*)\bigr]\). Therefore:
\begin{equation} 
\begin{split}
-\frac{|\mathcal{Z}|}{2\eta}  \leq \Delta_j(r+1) \leq  \frac{|\mathcal{Z}|}{2\eta},
\end{split}
\end{equation}
We assume $\eta \geq 0.1\%$ as we observe through common data, 
For \(r\) such that \(\Delta_j(r+1) \geq 0\), the sequence increases. However, the boundedness of \(\Delta_j(r+1)\) ensures that these increases are finite for the C total rounds.

Based on preliminary results that is shown in Appendix~\ref{appendix:assump} and Figure~\ref{fig:assumption}, we assume for \(r\) such that \(\Delta_j(r+1) < 0\), the sequence 
\begin{equation} 
\begin{split}
\mathbb{E}
[D_{\text{KL}}\bigl(\tilde{p}_\theta(\cdot \mid q^*) \parallel \tilde{p}^{\,r+1}_\theta(\cdot \mid q_j)\bigr)]  = \\e^{-\lambda_{q_{j}} } 
\mathbb{E}[D_{\text{KL}}\bigl(\tilde{p}_\theta(\cdot \mid q^*) \parallel \tilde{p}^{\,r}_\theta(\cdot \mid q_j)\bigr)]
\end{split}
\end{equation}
where $\lambda_{q_{j}}$ is a constant relevant to the query $q_{j}$.

Due to the abovementioned converge rate and nearly monotonic convergence except for bounded finitely many bad steps, $\mathbb{E}[D_{\text{KL}}\bigl(\tilde{p}_\theta(\cdot \mid q^*) \parallel \tilde{p}^{\,r}_\theta(\cdot \mid q_j)\bigr)]$ converges to a limit \(L\) at the unique global minimum of KL divergence as it's strictly convex, so:
\begin{equation}
L = 0 \implies \lim_{r \to \infty} \mathbb{E}[\epsilon_j^r(z)] = 0, \forall z\in\, \mathcal{Y}_{\text{target}}.
\end{equation}

In terms of variance, since we have each answer's error satisfying $\epsilon_j^r(z)
  \;\in\;
  \bigl[
    -\tilde{p}_\theta(z\mid q^*),\;1-\tilde{p}_\theta(z\mid q^*)
  \bigr]$ for every round \(r\), \(\epsilon_j^r(z)\) is a random variable with
\begin{equation}
\begin{split}
      \bigl|\epsilon_j^r(z)\bigr|
  \;\le\;
  \max\!\Bigl\{|\tilde{p}_\theta(z\mid q^*)|,\,\bigl|1-\tilde{p}_\theta(z\mid q^*)\bigr|\Bigr\} \\
  \;=\; M_z
  \quad
  (\text{some finite constant }M_z>0).
\end{split}
\end{equation}
Thus \(\epsilon_j^r(z)\) is \emph{surely} bounded, taking values in a finite interval \([-M_z,M_z]\). It follows immediately that
\begin{equation}
  \mathrm{Var}[\epsilon_j^r(z)]
  \;\le\;
  \bigl(M_z\bigr)^2
  \quad
  \text{for every round $r$.}
\end{equation}
Thus \(\text{Var}[\epsilon_j^r(z)]\) is bounded. 

$\epsilon_j^{r}(z)$ is weakly dependent due to the diversity of the queries and individual reflection traces. Putting these conditions together, the law of large number for weakly dependent processes implies $
\sum_{j=0}^n w_j \epsilon_j^{r}(z) \xrightarrow[n \to \infty]{\text{in probability}} 0.$ Therefore,
\begingroup
\setlength{\abovedisplayskip}{7pt}
\setlength{\belowdisplayskip}{7pt}
\begin{equation}
\begin{split}
\lim_{\substack{n \to \infty \\ r \to \infty}}\sum_{j=0}^n w_j \tilde{p}^{\,r}_\theta(z \mid q_j) 
&= \\ \underbrace{\sum_{j=0}^n w_j}_{=1} \tilde{p}_\theta(z \mid q^*)
&\quad + \underbrace{\sum_{j=0}^n w_j \epsilon_j^{r}(z)}_{\to 0}.
\end{split}
\end{equation}
\endgroup

\begin{table*}[ht]
\centering
\small
\scalebox{1}{
\begin{tabular}{lcl}
\toprule
\textbf{Dataset Type} & \textbf{\#Data} & \textbf{Example} \\
\midrule
PopQA\_less\_popular & 459 & What is Geeling Ng's occupation? \\
PopQA\_popular & 452 & What is the capital of Hungary? \\
\midrule
TruthfulQA & 219 & What type of cancer kills the most women in the U.S.? \\
FreshQA & 283 & What's the city flower of Shanghai? \\
\midrule
FalseQA & 1867 & Which planet is larger, Mars or Moon? \\
\bottomrule
\end{tabular}}
\caption{Detailed statistics of the datasets used in the paper.}
\label{tab:data}
\end{table*}

\subsection{Cost Analysis}
\label{appendix:cost}
We present a detailed cost analysis on our proposed ~\framework{}. In a self-consistency-based method, we typically sample a simple query 5 times, requiring 5 API calls. For our method, starting with a single query, we need 1 API call for question conceptualization, 1 for perspective generation, and 5 for generating questions under different perspectives. During agent interaction (with 5 agents), each agent answers their assigned question, followed by multiple interaction rounds. Assuming an average of 2.5 rounds, agent interaction requires 20 calls. In total, our method averages 25 API calls, making it 5 times more expensive than the base self-consistency method. 

However, we would like to emphasize that in high-stakes applications where correctness is prioritized over cost, our calibrated uncertainty score can provide users with a reliable measure of how much they can trust the model's output. As shown in Argument~\ref{claim:distribution-shift} and Experiment~\ref{exp:retrievability}, single-context methods like self-consistency suffer from contextual bias. Multi-context enabled by multi-agent interactions can help the model recognize and correct for these biases and recover its true internal belief. We further present the cost, i.e., the number of inference calls for all the baselines in Appendix Table~\ref{tab:cost}.

Additionally, The varied queries generated by our method can also be utilized for fine-tuning or training purposes. The intermediate results generated by our method, including varied queries and the self-reflection interaction processes, can be further leveraged to create synthetic data for finetuning or training LLMs. 

Future work can explore ways to maintain the same level of performance while reducing costs. This could involve using fewer but higher-quality questions from diverse perspectives and minimizing the number of interaction rounds.

\subsection{Datasets Statistics}
\label{appendix:data}

We consider five different datasets under three categories. We present the detailed dataset statistics in Appendix Table~\ref{tab:data}. All the datasets have the license to be publicly used.  \textbf{Entity-centric QA}: we randomly sample data from PopQA
~\citep{mallen-etal-2023-trust} where each question is created by converting a knowledge tuple retrieved from Wikidata using a template. 1) PopQA popular. We sample questions where the entity's popularity is larger than $10^4$ as the criteria in the original paper. 2) PopQA less popular. We sample questions where the entity's popularity is lower than $10^4$. \textbf{General QA}: 3) TruthfulQA~\citep{lin-etal-2022-truthfulqa}. Note that not all data in TruthfulQA tests about factual questions. We choose examples only from categories: Law, Sociology, Health, History, and Language, focusing on clear facts instead of opinions. 4) FreshQA~\citep{vu2023freshllmsrefreshinglargelanguage}. We adopt the 07112024 version and select one-hop slow-changing or never-changing data points where the effective year is before 2022 to avoid the temporal influence. \textbf{False assumption QA}: 5) FalseQA~\citep{hu-etal-2023-wont}. All questions in the dataset contain false assumptions and we remove all the WHY questions. 

\begin{table}[t]
\centering
\small
\scalebox{1}{
\begin{tabular}{lc}
\toprule
\textbf{Model} & \textbf{Cost} \\
\midrule
\multicolumn{2}{l}{uncertainty estimation methods} \\
\midrule
SC (Ecc)	&5\\
SC (Degree)	&5	\\
SC (EigV)	&5	\\
SC (KLE) &16  \\
SC (SE) &6  \\
\midrule
\multicolumn{2}{l}{hallucination detection/ direct inferernce methods} \\
\midrule
Greedy& 1\\
Self-Reflect& 2\\
Self-Eval w Samples &  6\\
Multiple-Recite & 2\\
\midrule
\multicolumn{2}{l}{hallucination detection/ sampling-based methods} \\
\midrule
SC (3/5)  & 6\\
SeQ    &  7\\
diverseQ  & 13 \\
\midrule
\framework{} & 25\\
\bottomrule
\end{tabular}}
\caption{Comparison of cost across all the methods. Specifically, we present the number of API calls.}
\label{tab:cost}
\end{table}

\subsection{Thresholds for the Abstention Policy}
\label{appendix:threshold}
We adopt two variants of \framework{} where we have 5 agents, i.e., 5 varied querys: 1) Agent (Loose Majority Vote): We abstain when the uncertainty score exceeds the threshold, calculated as the entropy of 3 answers with probabilities of 0.6 (3/5), 0.2 (1/5), and 0.2 (1/5). This setup implies that at least one answer still has a majority (60\%, 3/5 chance).  2) Agent: We use a stricter majority vote, abstaining when the uncertainty score exceeds the threshold, calculated as the entropy of 2 answers with probabilities of 0.6 (3/5) and 0.4 (2/5). This is the strictest majority vote threshold. The two variants balance flexibility and conservatism in decision-making: the loose majority vote allows for more uncertainty with three answer probabilities, making it suitable for situations with acceptable disagreement but where one answer is still dominant. In contrast, the stricter majority vote, using two-answer probabilities, ensures abstention in cases where only minor uncertainty can be tolerated.

\subsection{ Error Analysis for Suboptimal Retrieval of Parametric Knowledge for the Models}
\label{appendix:Retrivability_error}
We conduct the error analysis for the retrieval of parametric knowledge for the models with 45 examples, 23 sampled from Llama-3-70b-Instruct and 22 sampled from Claude-3-Sonnet. We observe similar behaviors in both models. As we discuss we observe that the model is more likely to generate a different response, even when it knows the correct answer to the target knowledge, under these three conditions: 1) Example 1 sampled from Llama-3-70b-Instruct in Appendix Table~\ref{tab:appendix_error}: the changes in the diverse query, either phrasing or the added context, significantly diverges in wording. The varied queries are the least two similar queries to the target query among the 5 varied queries according to the score of SentenceBert. 2) In Example 2, sampled from Llama-3-70b-Instruct in Appendix Table~\ref{tab:appendix_error}, an incorrect answer is more popular within the context of the target knowledge. For instance, Cristiano Ronaldo is a more well-known football player compared to Ali Daei. 3) Example 3, sampled from Claude-3-Sonnet in Appendix Table~\ref{tab:appendix_error}, illustrates a case where the additional context is more closely related to a different possible answer. In this instance, the model is distracted by the ``Yangtze River Delta region'' mentioned in the varied query as the white peony (wrong answer) is native to the Yangtze River Delta region, where Shanghai is located. Note that after agent interaction, the models answer all the questions correctly.

\subsection{Performance Evaluation for Hallucination Detection on Individual Datasets}
\label{appendix:hallucination_numbers}

We present the individual dataset performance on the two models in Appendix Table~\ref{tab:claude}
 and Appendix Table~\ref{tab:llama} respectively. We present Accuracy-Recall (AR)-curves for both the baselines and the proposed method on individual datasets in Appendix Fig.~\ref{fig:appnedix_main}.

Table~\ref{tab:wrong_correct} further shows the proportion of initially incorrect responses that are corrected after interaction (\textbf{Correct}) and initially correct responses that become incorrect after interaction (\textbf{Wrong}) across datasets. A response is considered \textbf{Correct} if at least 3 out of 5 sampled answers to the target query from the diverse queries are incorrect before interaction, and all 5 answers are correct after interaction. Conversely, a response is considered \textbf{Wrong} if at least 3 out of 5 sampled answers are correct before interaction, and at least 3 out of 5 are incorrect afterward. Note that we adopt a stricter criterion for \textbf{Correct}. \framework{} significantly improves correctness while introducing minimal errors.

 \begin{table}[t]
    \centering
    \setlength{\tabcolsep}{3.5pt}
    \scalebox{0.75}{
    \begin{tabular}{lcccccc}
        \toprule
        Dataset & \multicolumn{2}{c}{Claude-3-Sonnet} & \multicolumn{2}{c}{Llama-3-70b-Instruct} \\
        \cmidrule(lr){2-3} \cmidrule(lr){4-5}
         & Wrong & Correct & Wrong & Correct \\
        \midrule
        PopQA pop     & 0.052 & 0.487 & 0.100 & 0.545 \\
        PopQA less pop  & 0.061 & 0.179 & 0.055 & 0.300 \\
        FalseQA             & 0.000 & 0.042 & 0.088 & 0.140 \\
        TruthfulQA          & 0.035 & 0.568 & 0.150 & 0.605 \\
        FreshQA             & 0.089 & 0.302 & 0.086 & 0.381 \\
        \bottomrule
    \end{tabular}
    }
    \caption{The ratios of correct-to-incorrect and incorrect-to-correct response changes.} 
    \label{tab:wrong_correct}
\end{table}

\begin{table*}[h!]
    \centering
    \scalebox{0.47}{
    \begin{tabular}{lcccccccccccccccccccc}
        \toprule
        & \multicolumn{4}{c}{TruthfulQA} & \multicolumn{4}{c}{FreshQA} & \multicolumn{4}{c}{FalseQA} & \multicolumn{4}{c}{PopQA popular} & \multicolumn{4}{c}{PopQA less popular} \\
        \cmidrule(lr){2-5} \cmidrule(lr){6-9} \cmidrule(lr){10-13} \cmidrule(lr){14-17} \cmidrule(lr){18-21}
        Method & Acc & Ab-R & TruthF & Correct & Acc & Ab-R & TruthF & Correct & Acc & Ab-R & TruthF & Correct & Acc & Ab-R & TruthF & Correct & Acc & Ab-R & TruthF & Correct \\
        \midrule
        Greedy                              & 0.723 & 0.059 & 0.680 & 0.739 & 0.777 & 0.064 & 0.727 & 0.791 & 0.891 & 0.093 & 0.809 & 0.901 & 0.824 & 0.037 & 0.793 & 0.830 & 0.344 & 0.420 & 0.199 & 0.619 \\
        Self-Reflect              & 0.731 & 0.082 & 0.671 & 0.753 & 0.770 & 0.032 & 0.746 & 0.777 & 0.888 & 0.066 & 0.829 & 0.895 & 0.839 & 0.098 & 0.768 & 0.866 & 0.470 & 0.520 & 0.226 & 0.746 \\
        Self-Eval w Samples           & 0.725 & 0.087 & 0.662 & 0.749 & 0.709 & 0.064 & 0.728 & 0.664 & 0.879 & 0.077 & 0.812 & 0.889 & 0.812 & 0.059 & 0.773 & 0.832 & 0.482 & 0.562 & 0.212 & 0.774 \\
        Recitation & 0.724 & 0.073 & 0.671 & 0.744 & 0.743 & 0.049 & 0.707 & 0.707 & 0.839 & 0.071 & 0.780 & 0.851 & 0.828 & 0.039 & 0.795 & 0.834 & 0.366 & 0.431 & 0.208 & 0.639 \\
         \midrule
        SC (3/5)              & 0.682 & 0.037 & 0.658 & 0.694 & 0.777 & 0.028 & 0.755 & 0.783 & 0.887 & 0.063 & 0.831 & 0.894 & 0.833 & 0.059 & 0.784 & 0.843 & 0.440 & 0.577 & 0.186 & 0.763 \\
        SeQ      & 0.782 & 0.183 & 0.639 & 0.822 & 0.814 & 0.163 & 0.681 & 0.844 & 0.888 & 0.099 & 0.800 & 0.899 & 0.852 & 0.061 & 0.800 & 0.861 & 0.309 & 0.420 & 0.186 & 0.606 \\
        diveseQ  & 0.739 & 0.261 & 0.545 & 0.807 & 0.856 & 0.216 & 0.671 & 0.887 & 0.874 & 0.302 & 0.610 & 0.912 & 0.891 & 0.193 & 0.730 & 0.923 & 0.714 & 0.777 & 0.159 & 0.936 \\
         \midrule
        Agent (Loose Majority Vote)                  & 0.740 & 0.078 & 0.683 & 0.761 & 0.826 & 0.085 & 0.756 & 0.841 & 0.907 & 0.080 & 0.834 & 0.914 & 0.852 & 0.059 & 0.814 & 0.873 & 0.537 & 0.546 & 0.243 & 0.790 \\
        Agent                & 0.753 & 0.128 & 0.656 & 0.784 & 0.879 & 0.184 & 0.717 & 0.901 & 0.924 & 0.139 & 0.795 & 0.935 & 0.883 & 0.144 & 0.768 & 0.911 & 0.611 & 0.670 & 0.201 & 0.872 \\
        \bottomrule
    \end{tabular}}
    \caption{Performance comparison on various datasets for Claude-3-Sonnet. Acc refers to accuracy, Ab-R refers to abstention rate, TruthF refers to truthfulness, Correct refers to correctness.}
    \label{tab:claude}
\end{table*}

\begin{table*}[h!]
    \centering
    \scalebox{0.47}{
    \begin{tabular}{lcccccccccccccccccccc}
        \toprule
        & \multicolumn{4}{c}{TruthfulQA} & \multicolumn{4}{c}{FreshQA} & \multicolumn{4}{c}{FalseQA} & \multicolumn{4}{c}{PopQA popular} & \multicolumn{4}{c}{PopQA less popular} \\
        \cmidrule(lr){2-5} \cmidrule(lr){6-9} \cmidrule(lr){10-13} \cmidrule(lr){14-17} \cmidrule(lr){18-21}
        Method & Acc & Ab-R & TruthF & Correct & Acc & Ab-R & TruthF & Correct & Acc & Ab-R & TruthF & Correct & Acc & Ab-R & TruthF & Correct & Acc & Ab-R & TruthF & Correct \\
        \midrule
        Greedy                              & 0.709 & 0.027 & 0.690 & 0.717 & 0.784 & 0.000 & 0.784 & 0.784 & 0.858 & 0.003 & 0.855 & 0.859 & 0.856 & 0.002 & 0.854 & 0.856 & 0.367 & 0.029 & 0.356 & 0.385 \\
        Self-Reflect              & 0.702 & 0.018 & 0.689 & 0.708 & 0.748 & 0.018 & 0.735 & 0.753 & 0.871 & 0.011 & 0.861 & 0.872 & 0.826 & 0.009 & 0.832 & 0.841 & 0.386 & 0.146 & 0.330 & 0.476 \\
        Self-Eval w Samples           & 0.670 & 0.046 & 0.639 & 0.685 & 0.721 & 0.000 & 0.721 & 0.721 & 0.853 & 0.022 & 0.834 & 0.856 & 0.819 & 0.002 & 0.817 & 0.819 & 0.336 & 0.033 & 0.325 & 0.358 \\
        Recitation & 0.707 & 0.018 & 0.694 & 0.712 & 0.705 & 0.018 & 0.693 & 0.710 & 0.785 & 0.009 & 0.778 & 0.787 & 0.782 & 0.002 & 0.780 & 0.782 & 0.363 & 0.013 & 0.358 & 0.372 \\
         \midrule
        SC (3/5)              & 0.619 & 0.018 & 0.607 & 0.626 & 0.791 & 0.018 & 0.777 & 0.795 & 0.880 & 0.012 & 0.869 & 0.881 & 0.848 & 0.013 & 0.837 & 0.850 & 0.408 & 0.170 & 0.338 & 0.509 \\
        SeQ          & 0.681 & 0.116 & 0.602 & 0.718 & 0.769 & 0.066 & 0.718 & 0.784 & 0.915 & 0.064 & 0.857 & 0.921 & 0.828 & 0.034 & 0.800 & 0.834 & 0.437 & 0.215 & 0.343 & 0.558 \\
        diverseQ  & 0.676 & 0.155 & 0.571 & 0.763 & 0.798 & 0.088 & 0.728 & 0.813 & 0.865 & 0.071 & 0.803 & 0.874 & 0.869 & 0.065 & 0.825 & 0.891 & 0.489 & 0.389 & 0.299 & 0.688 \\
         \midrule
        Agent (Loose Majority Vote)                  & 0.750 & 0.050 & 0.712 & 0.763 & 0.806 & 0.035 & 0.777 & 0.813 & 0.894 & 0.026 & 0.870 & 0.897 & 0.868 & 0.011 & 0.872 & 0.883 & 0.471 & 0.235 & 0.361 & 0.595 \\
        Agent                & 0.752 & 0.078 & 0.694 & 0.772 & 0.831 & 0.078 & 0.767 & 0.845 & 0.899 & 0.037 & 0.865 & 0.903 & 0.875 & 0.026 & 0.865 & 0.891 & 0.508 & 0.343 & 0.334 & 0.677 \\
        \bottomrule
    \end{tabular}}
    \caption{Performance comparison on Llama-3-70b-Instruct for multiple datasets. Acc refers to accuracy, Ab-R refers to abstention rate, TruthF refers to truthfulness, Correct refers to correctness.}
    \label{tab:llama}
\end{table*}

\subsection{Additional Analysis for Query Diversity and Agent Interaction}
\label{appnedix:format}

We conduct an analysis using Llama-3.1-70b-Instruct to evaluate both the quality of question generation. For this, we randomly sample 20 instances per dataset per model, totaling 500 diverse queries per model.

To assess the quality of the diverse queries, we use two criteria: (1) Knowledge preservation, whether the generated query explicitly requires answering the target query, thereby preserving the necessary knowledge; and (2) Diversity, whether the queries for a given instance differ in wording. We consider a query set diverse if, when sampling a pair of queries from an instance (including the original target query, semantically equivalent, and compositional queries), at least 20\% of the wording differs. The prompts can be referred to in Appendix Figs.~\ref{fig:prompt-ana-k} and~\ref{fig:prompt-ana-d}. 

Based on these criteria, we see that 95.0\% of Claude's queries and 95.6\% of LLaMA's are knowledge-preserving. Furthermore, 97\% of Claude’s instances and 96\% of LLaMA’s instances meet the diversity criterion, indicating high-quality question generation.

For the small number of imperfect queries, those that are diverse and topically related to the target query but do not elicit direct answers to the target query (e.g., "What role does the respiratory system play in the progression and severity of AIDS?" for the target "Is AIDS a deadly respiratory disease?"), we manually inspect the interaction trajectories. We see that, despite minor inaccuracies, these queries still add value. Because the interaction prompt includes both the target query and a peer agent's answer, even imperfect queries and responses provided useful background, aiding the agent in reflecting on peer responses and improving its final answer.

Regarding instruction-following during interaction, we conduct a manual review of interaction trajectories. We observe that agents made very few errors: 99.2\% of interaction rounds adhered to the given instructions, and the clustering of semantically equivalent answers achieved 96.3\% accuracy.

\subsection{Human Error Analysis on Interaction Format}
\label{appnedix:format_human}
We also conduct a human error analysis for group interaction. In our analysis of 30 incorrect examples from each model, we identified two primary error types: (1) 50\% of errors occurred when agents were influenced by the majority’s incorrect answer, and (2) 15\% occurred when agents concluded that the question had no valid answer or was based on a false premise due to conflicting responses. This analysis further demonstrates that agents are more easily influenced by dominant incorrect information, reinforcing the importance of using one-on-one interactions for single-query uncertainty checking, as it allows the model to be exposed to diverse information while maintaining its ability to apply independent reasoning.

\subsection{Prompts for the Proposed \framework{}}
\label{appendix:prompt}
\clearpage

\begin{table*}[h!]
\centering
\scalebox{0.9}{%
\setlength{\tabcolsep}{4pt} 
\renewcommand{\arraystretch}{1.4} 
\begin{tabular}{|c|p{7cm}|p{7cm}|}
\hline
\rowcolor{gray!20}
\textbf{Example} & \textbf{Target Query / Gold Answer} & \textbf{Full Answer (Varied Query) / Extracted Answer} \\ \hline

\textbf{Example 1} 
& \textbf{Target Knowledge:} What film starred Brad Pitt's first leading role? &  \textbf{Gold Answer:} The Dark Side of the Sun 
\\ \hline
& \textbf{Varied Query:} Which film marked Brad Pitt's debut as the primary protagonist? &  \textbf{Full Answer:} Thelma \& Louise \newline \textbf{Extracted Answer:} Thelma \& Louise \\ \hline
& \textbf{Varied Query:} What was the influence of the film starring Brad Pitt's first leading role on the development of the Gothic horror genre in cinema? & \textbf{Full Answer:} You’re likely referring to the 1988 film "Dark Side of the Sun", which was not a Gothic horror film. Brad Pitt’s first leading role was actually in the 1988 film "Dark Side of the Sun". \newline \textbf{Extracted Answer:} The Dark Side of the Sun \\ \hline

\textbf{Example 2} 
& \textbf{Target Knowledge:} Who was the first football player to score one hundred international goals? &  \textbf{Gold Answer:} Ali Daei 
 \\ \hline
& \textbf{Varied Query:} Who was the first football player to score one hundred international goals? &  \textbf{Full Answer:} The first football player to score one hundred international goals is Cristiano Ronaldo . \newline \textbf{Extracted Answer:} Cristiano Ronaldo \\\hline

\textbf{Example 3} 
& \textbf{Target Knowledge:} What is the city flower of Shanghai? & \textbf{Gold Answer:} Magnolia denudata \\ \hline
& \textbf{Varied Query:} How does the city flower of Shanghai reflect the city's historical and cultural connection to the Yangtze River Delta region? & \textbf{Full Answer:} The city flower of Shanghai is the White Peony (Paeonia lactiflora), which reflects the city’s historical and cultural connection to the Yangtze River Delta region in several ways: 1. Native origin: The White Peony is native to the Yangtze River Delta region, where Shanghai is located. This emphasizes the city's deep roots in the region's natural environment. .. \newline \textbf{Extracted Answer:} White Peony (Paeonia lactiflora) \\ \hline

\end{tabular}}
\caption{Error analysis with 3 examples for the retrieval of parametric knowledge for models. Note that after agent interaction, the models give the correct answer to the target query. Here, the target knowledge is a simplified wording proxy for the underlying target knowledge. We do not know how this knowledge is actually represented within the model.}
\label{tab:appendix_error}
\end{table*}

\clearpage
\begin{figure*}[t]
    \centering
    \hspace*{-0.1cm} 
    \begin{minipage}{1\textwidth}
        \centering
        \begin{minipage}{0.49\textwidth}
            \centering
            \scalebox{1}{\includegraphics[width=\linewidth]{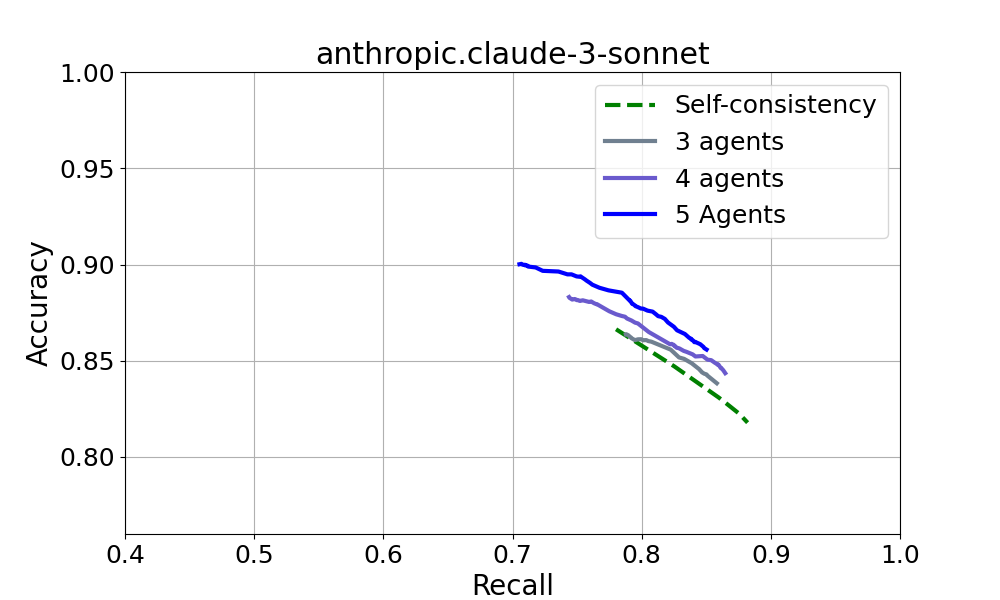}}
        \end{minipage}\hfill
        \begin{minipage}{0.49\textwidth}
            \centering
            \scalebox{1}{\includegraphics[width=\linewidth]{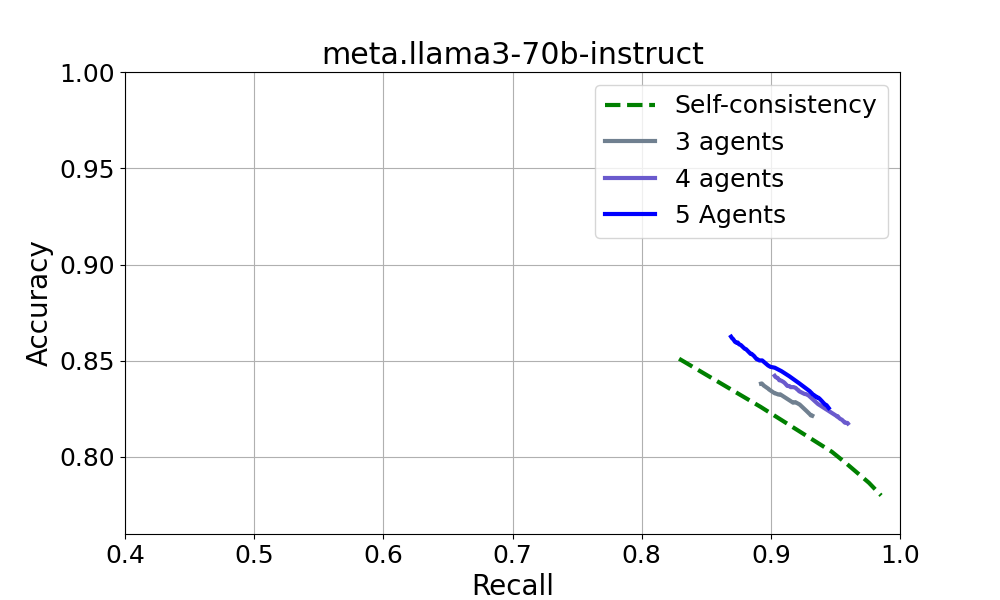}}   
        \end{minipage}
        \caption{Effect of $\#$agents on performance.}
        \label{fig:num_agents}
    \end{minipage}
\end{figure*}
\vspace{-0.1cm}

\begin{figure*}[t]
    \centering
    \hspace*{-0.1cm} 
    \begin{minipage}{1\textwidth}
        \centering
        \begin{minipage}{0.49\textwidth}
            \centering
            \scalebox{1}{\includegraphics[width=\linewidth]{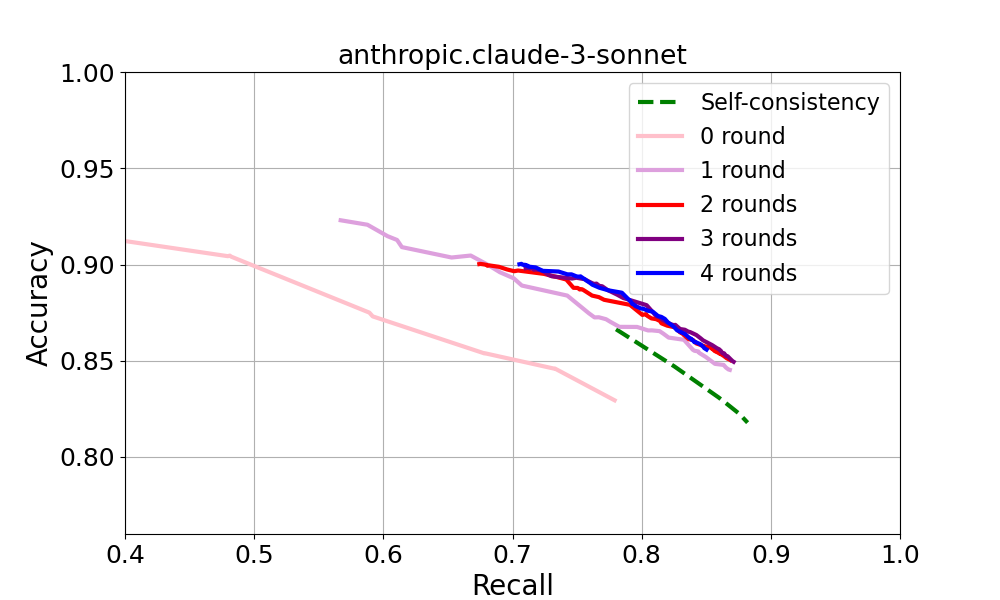}}
        \end{minipage}\hfill
        \begin{minipage}{0.49\textwidth}
            \centering
            \scalebox{1}{\includegraphics[width=\linewidth]{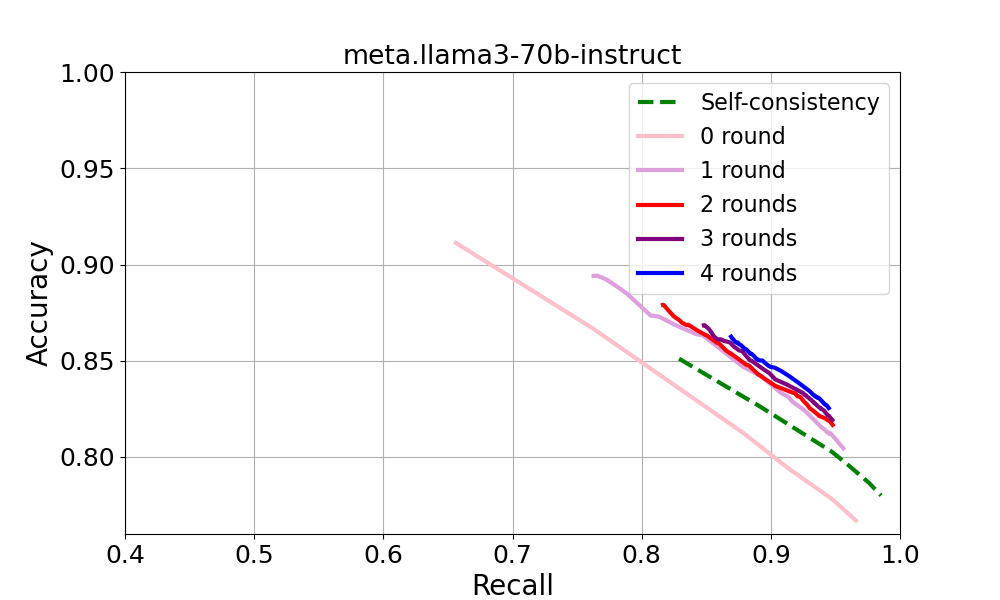}}   
        \end{minipage}
        \caption{Effect of $\#$interaction rounds on performance.}
        \label{fig:num_rounds}
    \end{minipage}
\end{figure*}

\begin{figure*}[t]
    \centering
    \hspace*{-0.1cm} 
    \begin{minipage}{1\textwidth}
        \centering
        \begin{minipage}{0.49\textwidth}
            \centering
            \scalebox{1}{\includegraphics[width=\linewidth]{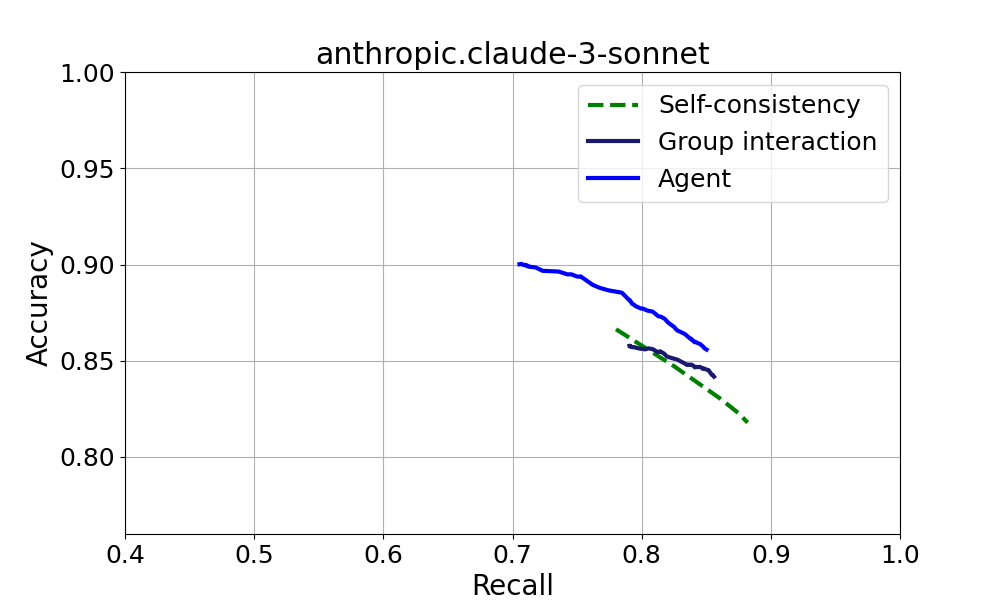}}
        \end{minipage}\hfill
        \begin{minipage}{0.49\textwidth}
            \centering
            \scalebox{1}{\includegraphics[width=\linewidth]{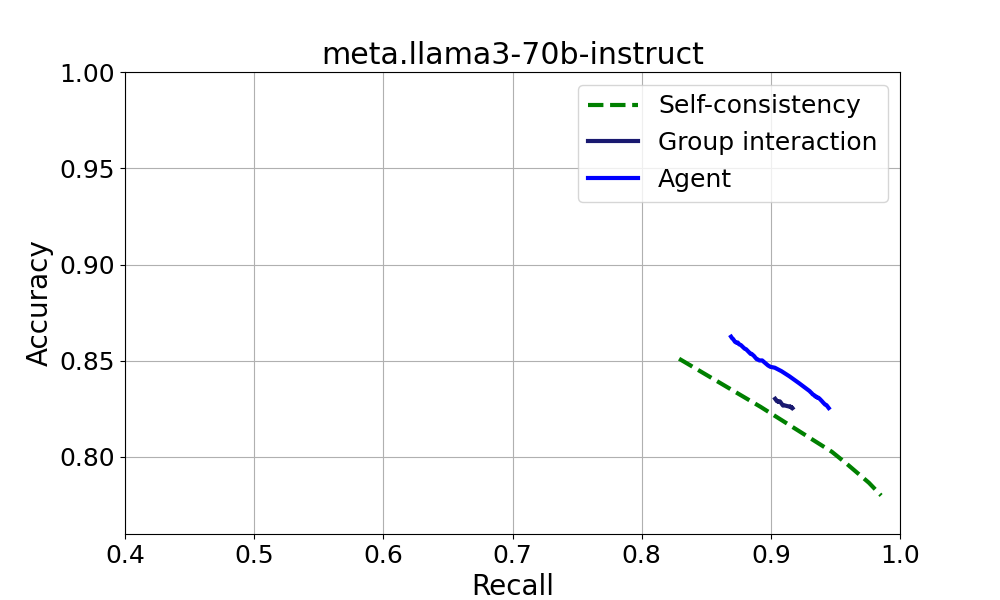}}   
        \end{minipage}
        \caption{Effect of interaction format on agent performance.}
        \label{fig:interaction_format}
    \end{minipage}
\end{figure*}
\vspace{-0.1cm}
\begin{figure*}[t]
    \centering
    \hspace*{-0.1cm} 
    \begin{minipage}{1\textwidth}
        \centering
        \begin{minipage}{0.49\textwidth}
            \centering
            \scalebox{1}{\includegraphics[width=\linewidth]{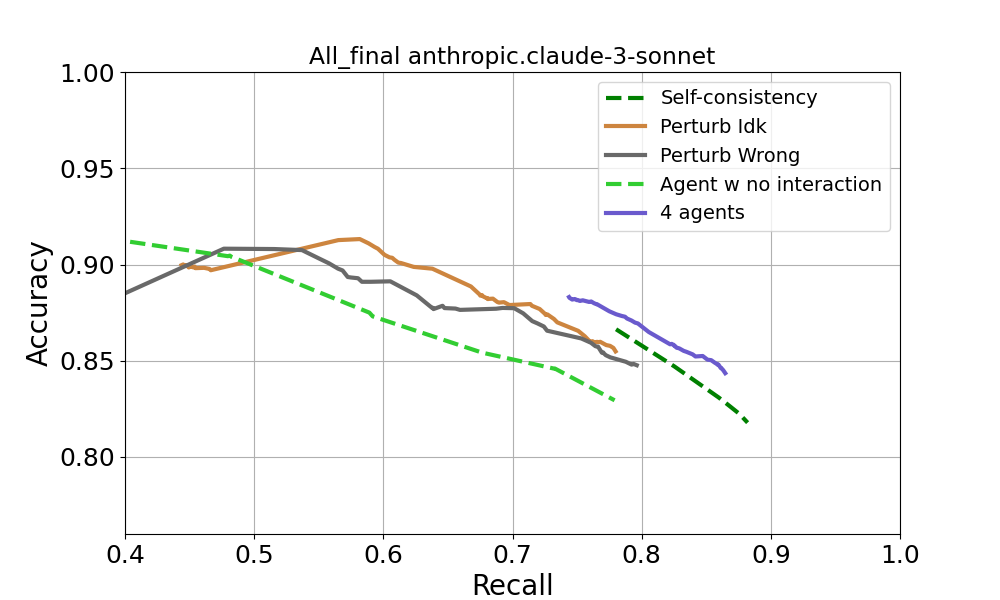}}
        \end{minipage}\hfill
        \begin{minipage}{0.49\textwidth}
            \centering
            \scalebox{1}{\includegraphics[width=\linewidth]{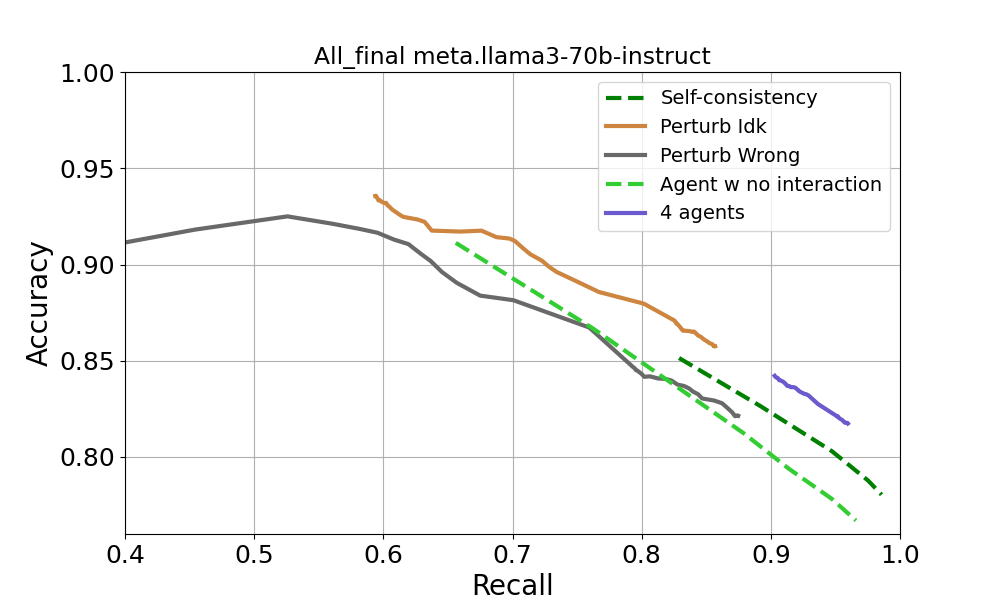}}   
        \end{minipage}
        \caption{Analysis of robustness in agent interactions.}
        \label{fig:interaction_robustness}
    \end{minipage}
\end{figure*}

\begin{figure*}[ht]
    \centering
    \begin{minipage}{0.45\textwidth}
        \centering
        \includegraphics[width=\linewidth]{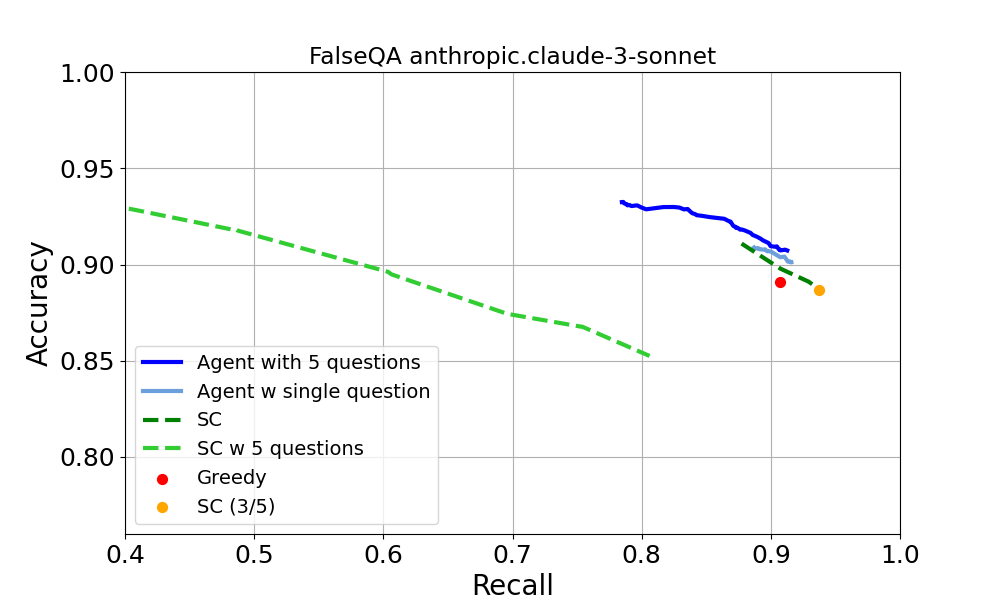}
    \end{minipage}\hfill
    \begin{minipage}{0.45\textwidth}
        \centering
        \includegraphics[width=\linewidth]{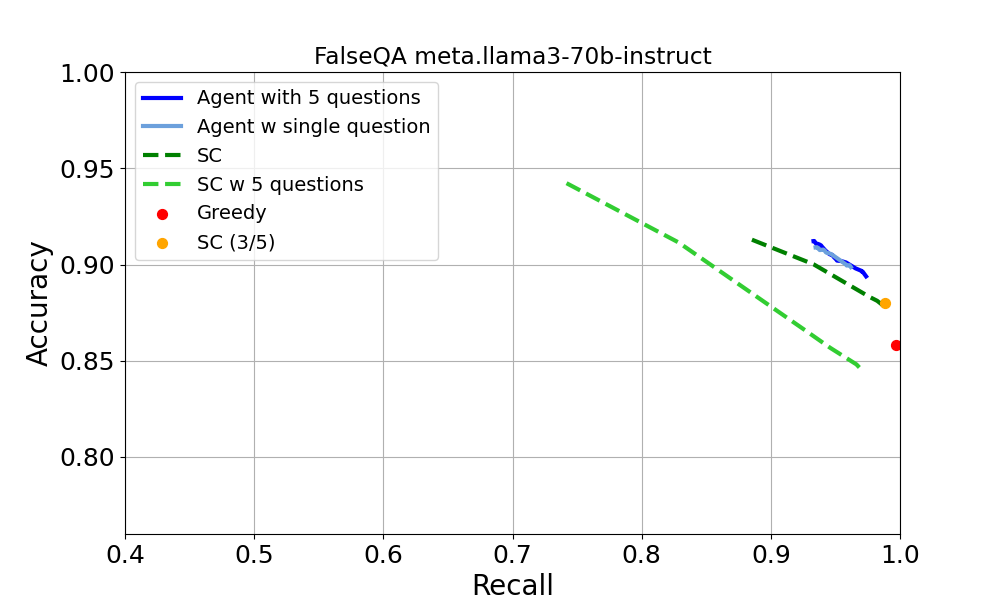}   
    \end{minipage}\hfill
     \begin{minipage}{0.45\textwidth}
        \centering
        \includegraphics[width=\linewidth]{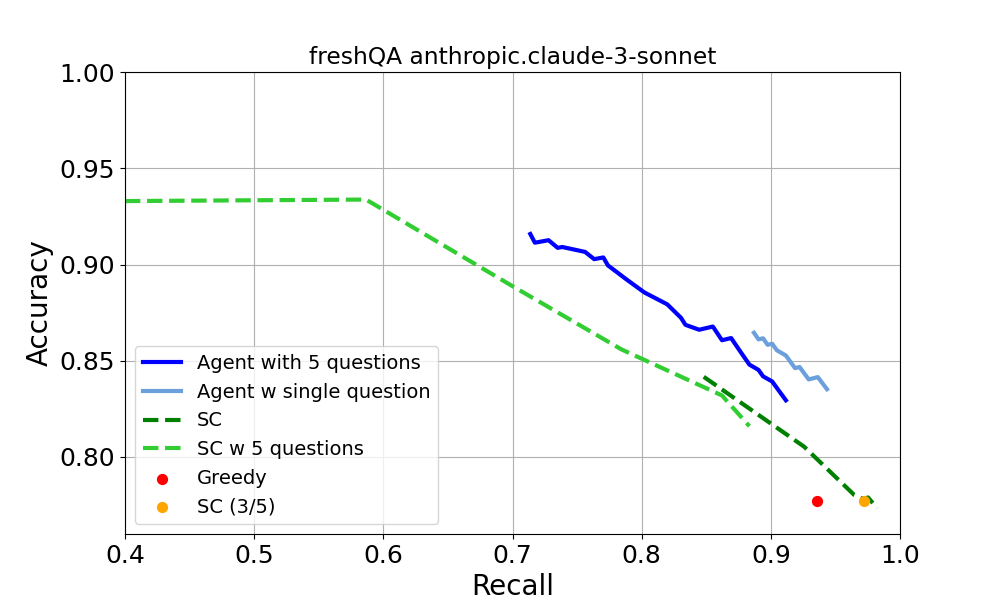}   
    \end{minipage}\hfill
     \begin{minipage}{0.45\textwidth}
        \centering
        \includegraphics[width=\linewidth]{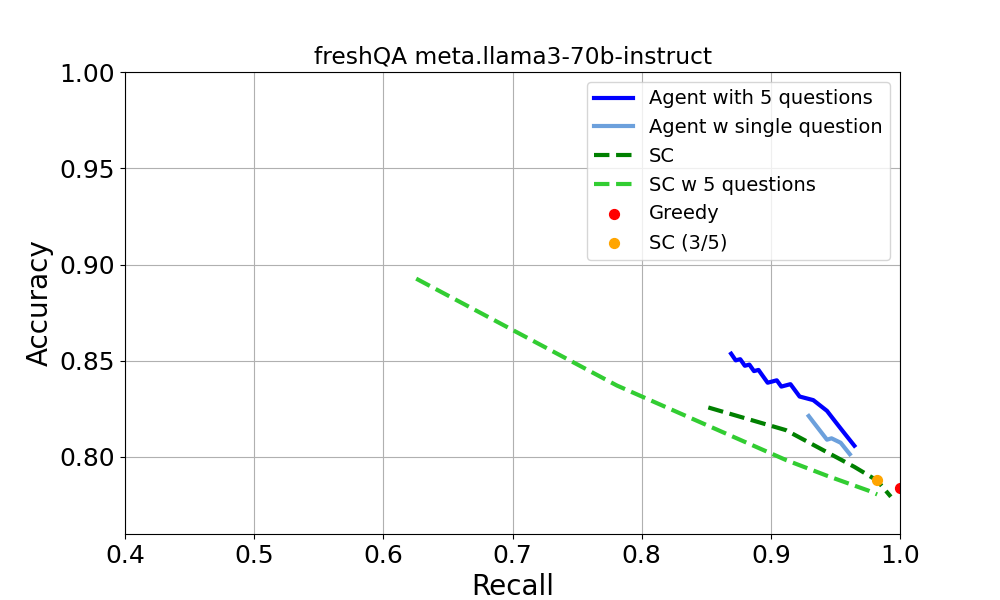}   
    \end{minipage}\hfill
    \begin{minipage}{0.45\textwidth}
        \centering
        \includegraphics[width=\linewidth]{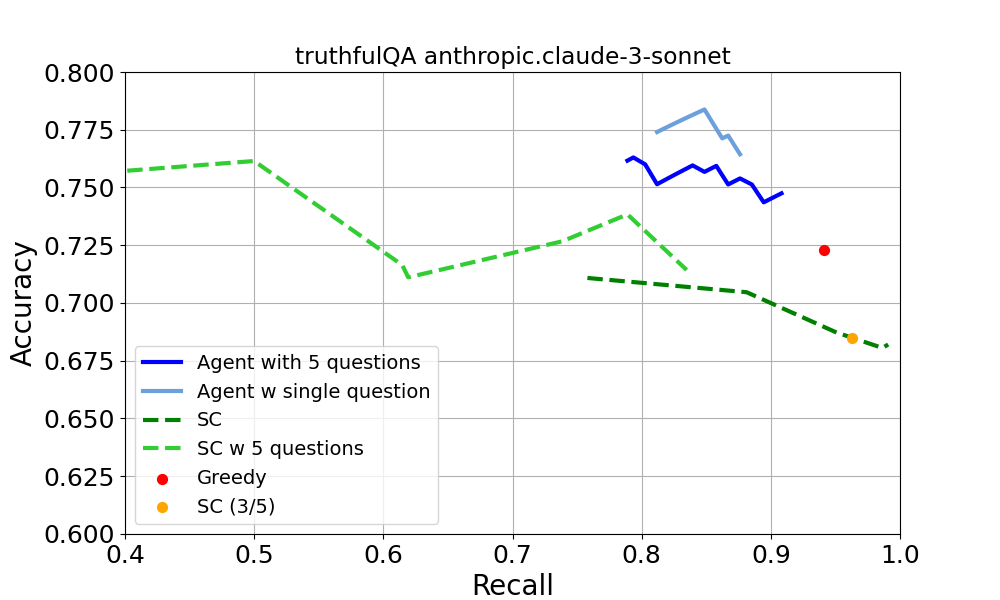}   
    \end{minipage}\hfill
     \begin{minipage}{0.45\textwidth}
        \centering
        \includegraphics[width=\linewidth]{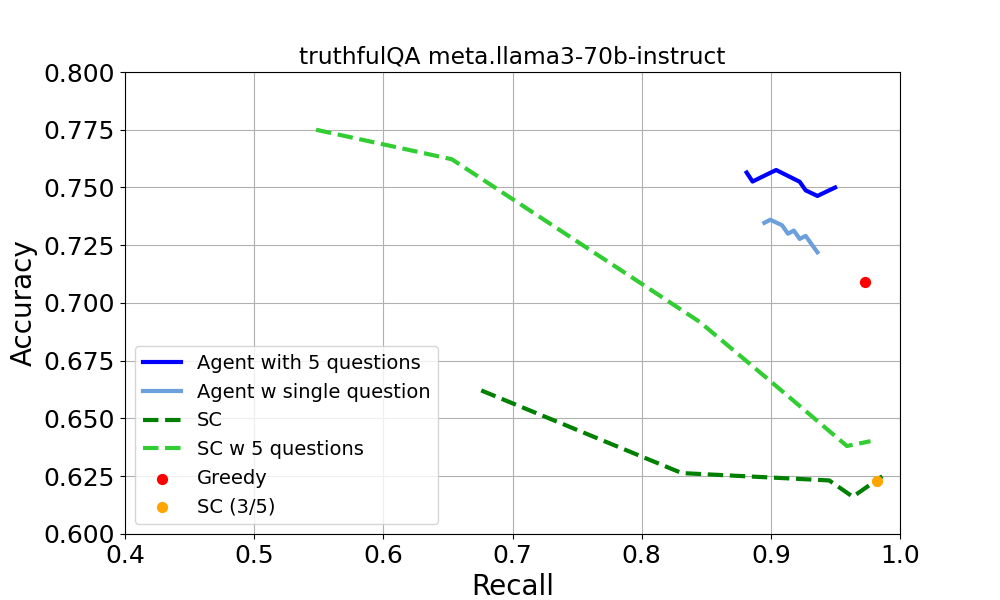}   
    \end{minipage}\hfill
    \begin{minipage}{0.45\textwidth}
        \centering
        \includegraphics[width=\linewidth]{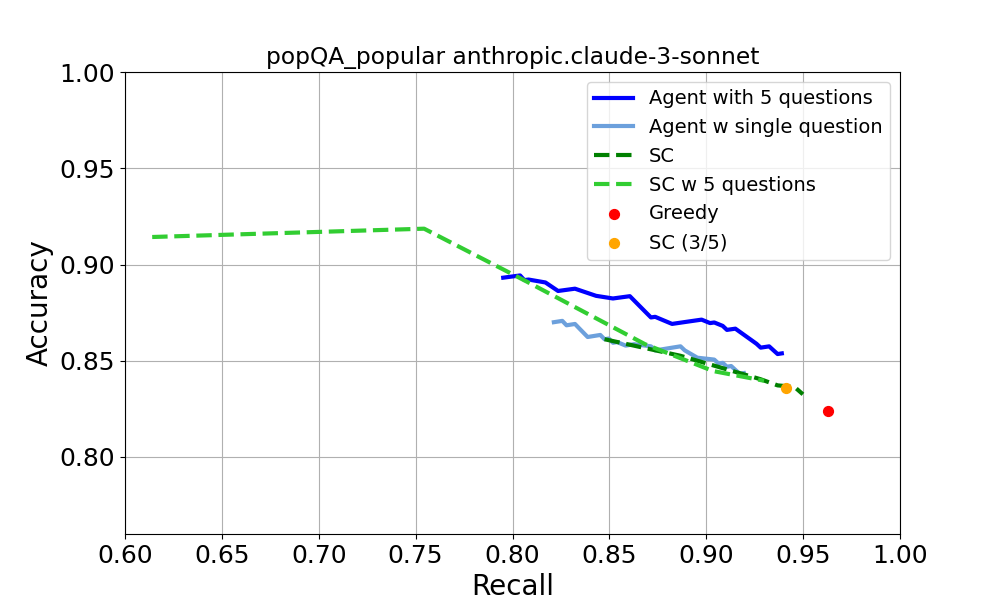}   
    \end{minipage}\hfill
     \begin{minipage}{0.45\textwidth}
        \centering
        \includegraphics[width=\linewidth]{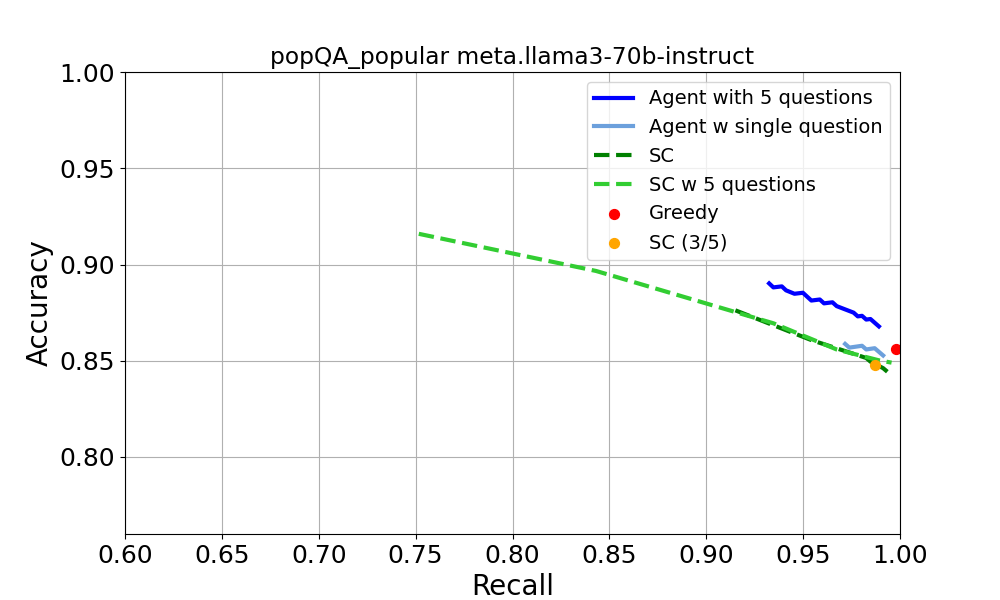}   
    \end{minipage}\hfill
    \begin{minipage}{0.45\textwidth}
        \centering
        \includegraphics[width=\linewidth]{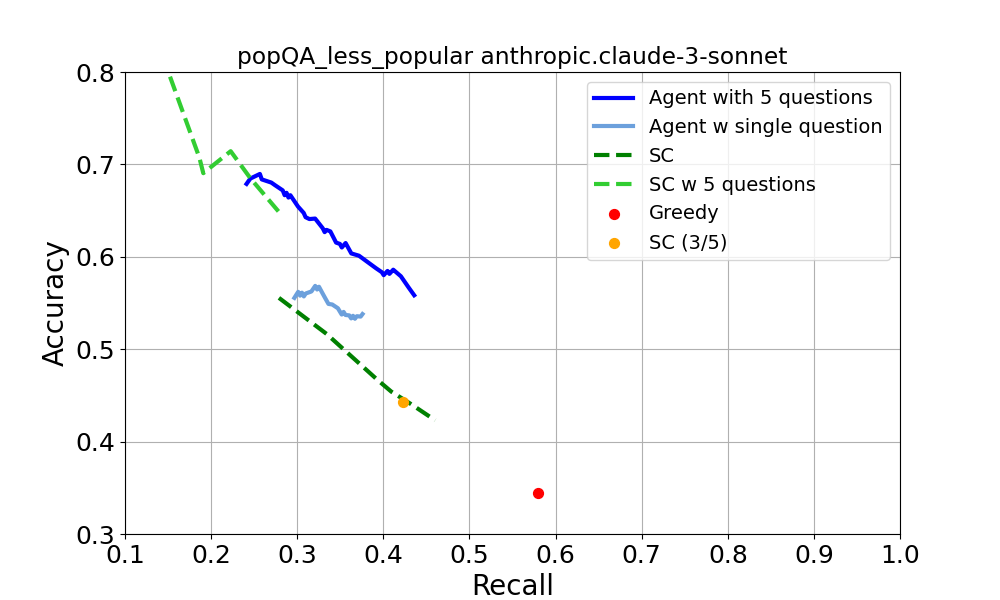}   
    \end{minipage}\hfill
     \begin{minipage}{0.45\textwidth}
        \centering
        \includegraphics[width=\linewidth]{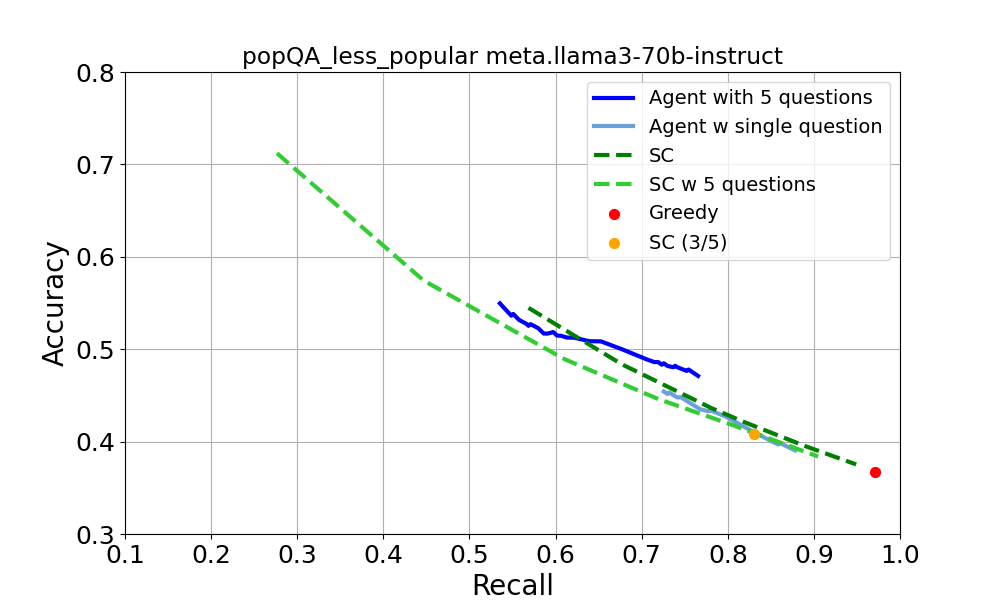}   
    \end{minipage}\hfill
    \caption{AR-curves for both the baselines and the proposed method on individual datasets. SC refers to self-consistency-based entropy. SC w 5 questions refers to the baseline using the agents' questions without agent interaction.}
    \label{fig:appnedix_main}
\end{figure*}

\begin{figure*}[ht]
    \centering
    \begin{minipage}{0.45\textwidth}
        \centering
        \includegraphics[width=\linewidth]{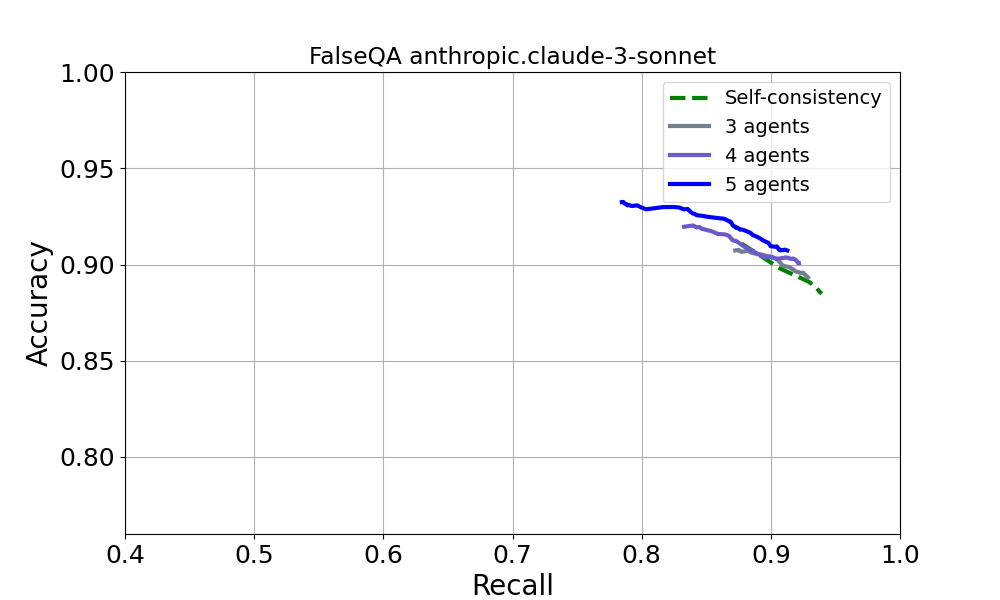}
    \end{minipage}\hfill
    \begin{minipage}{0.45\textwidth}
        \centering
        \includegraphics[width=\linewidth]{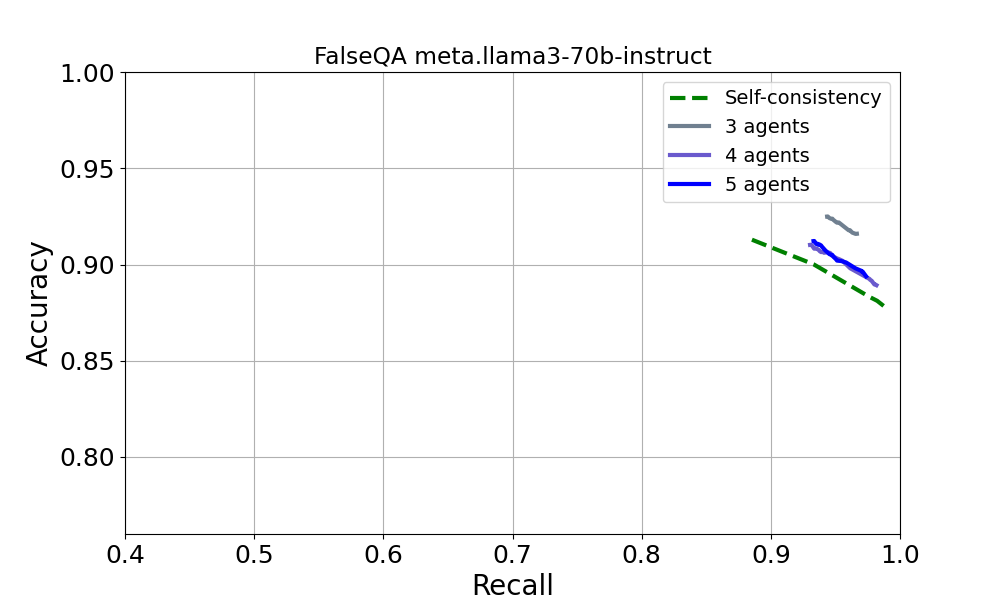}   
    \end{minipage}\hfill
     \begin{minipage}{0.45\textwidth}
        \centering
        \includegraphics[width=\linewidth]{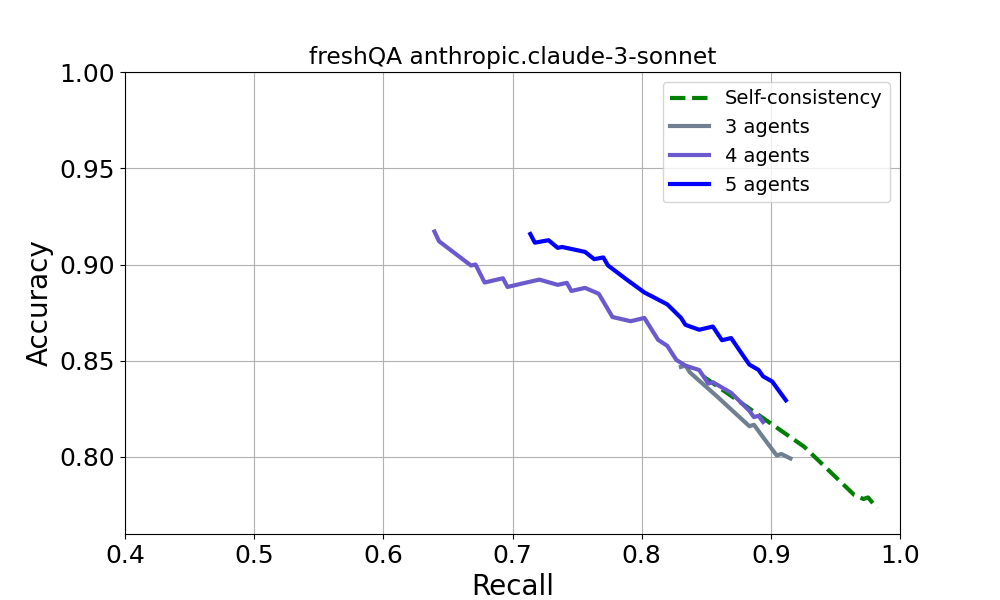}   
    \end{minipage}\hfill
     \begin{minipage}{0.45\textwidth}
        \centering
        \includegraphics[width=\linewidth]{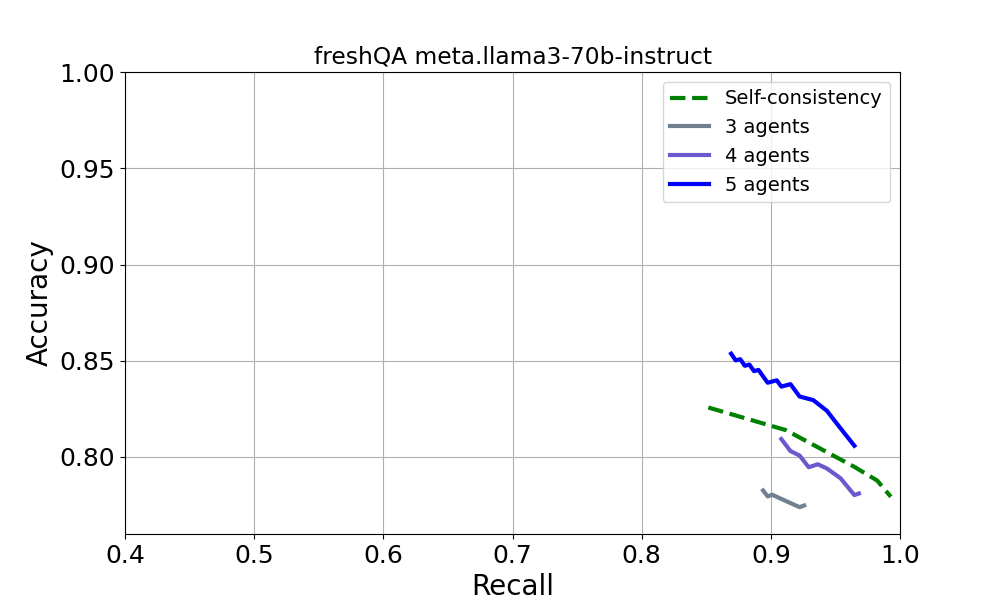}   
    \end{minipage}\hfill
    \begin{minipage}{0.45\textwidth}
        \centering
        \includegraphics[width=\linewidth]{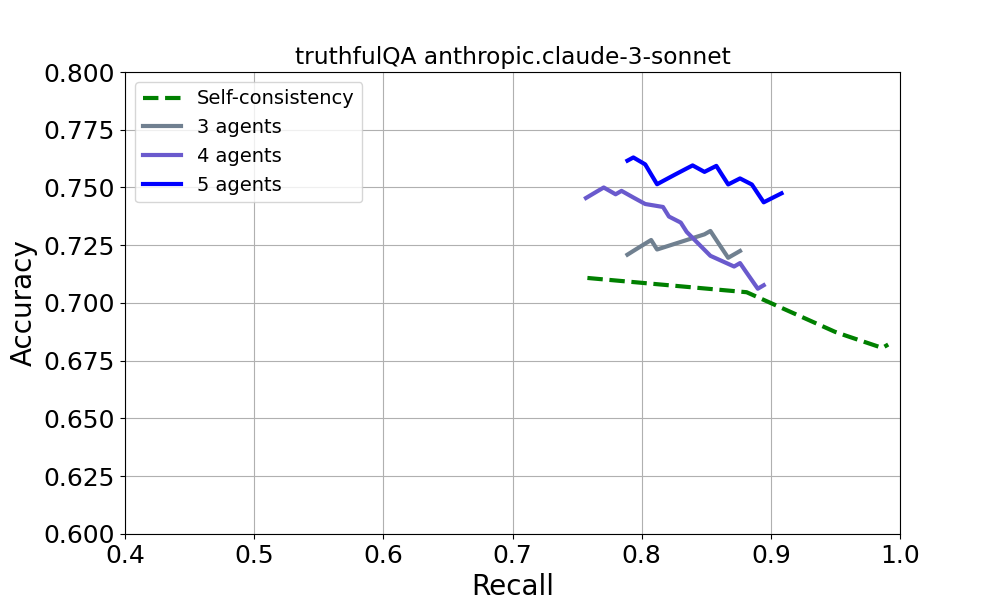}   
    \end{minipage}\hfill
     \begin{minipage}{0.45\textwidth}
        \centering
        \includegraphics[width=\linewidth]{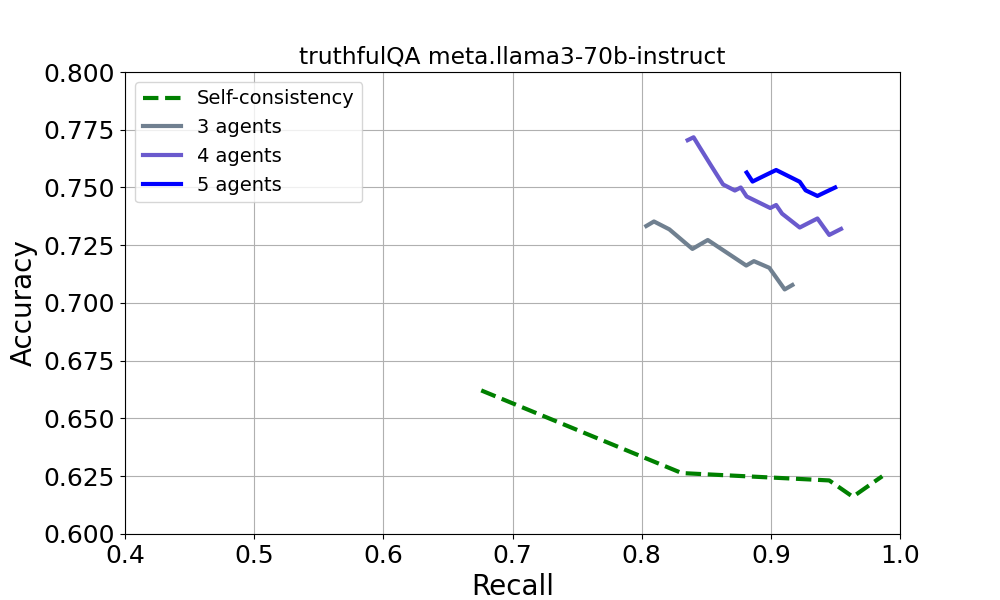}   
    \end{minipage}\hfill
    \begin{minipage}{0.45\textwidth}
        \centering
        \includegraphics[width=\linewidth]{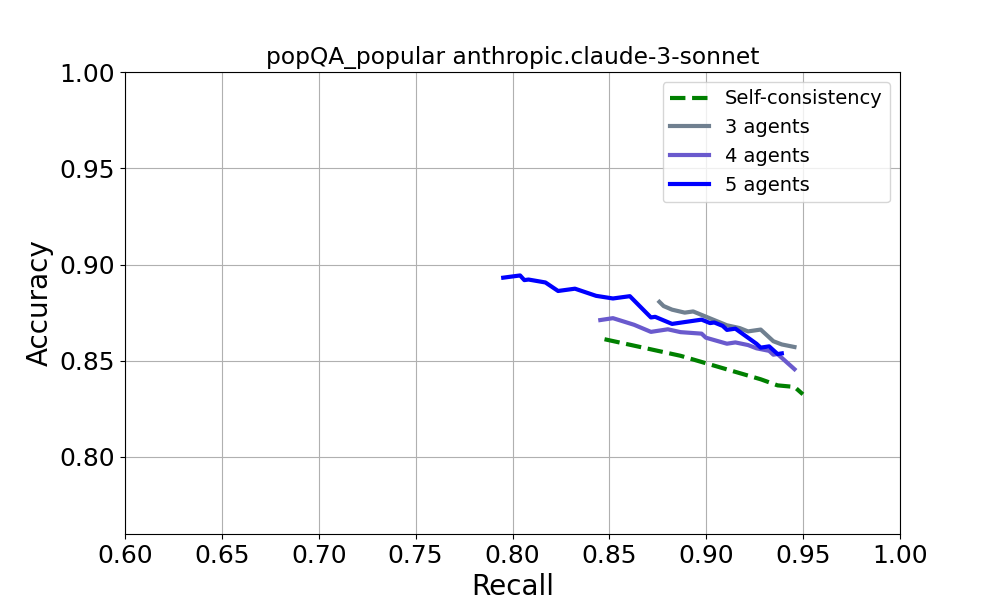}   
    \end{minipage}\hfill
     \begin{minipage}{0.45\textwidth}
        \centering
        \includegraphics[width=\linewidth]{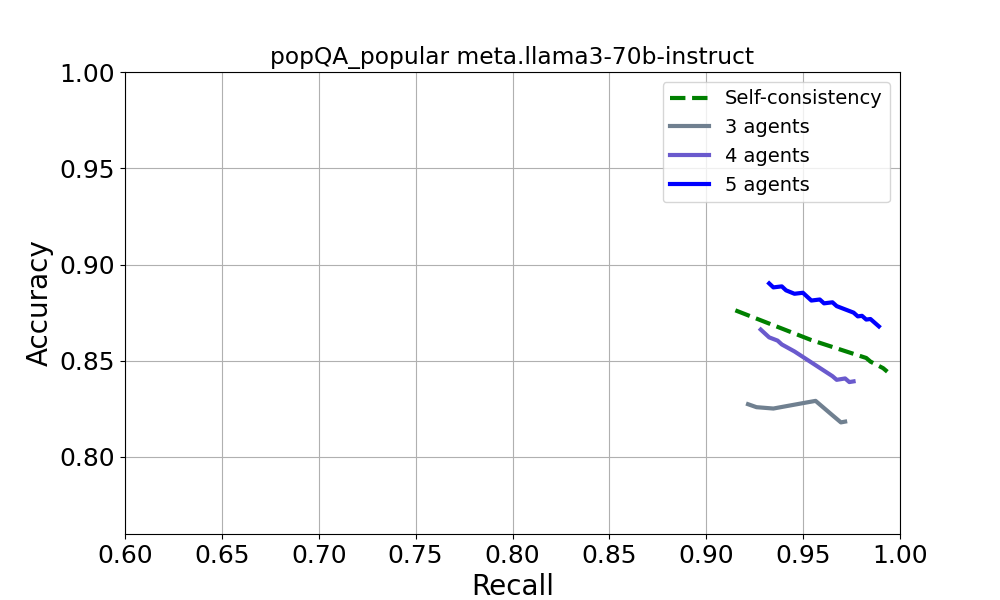}   
    \end{minipage}\hfill
    \begin{minipage}{0.45\textwidth}
        \centering
        \includegraphics[width=\linewidth]{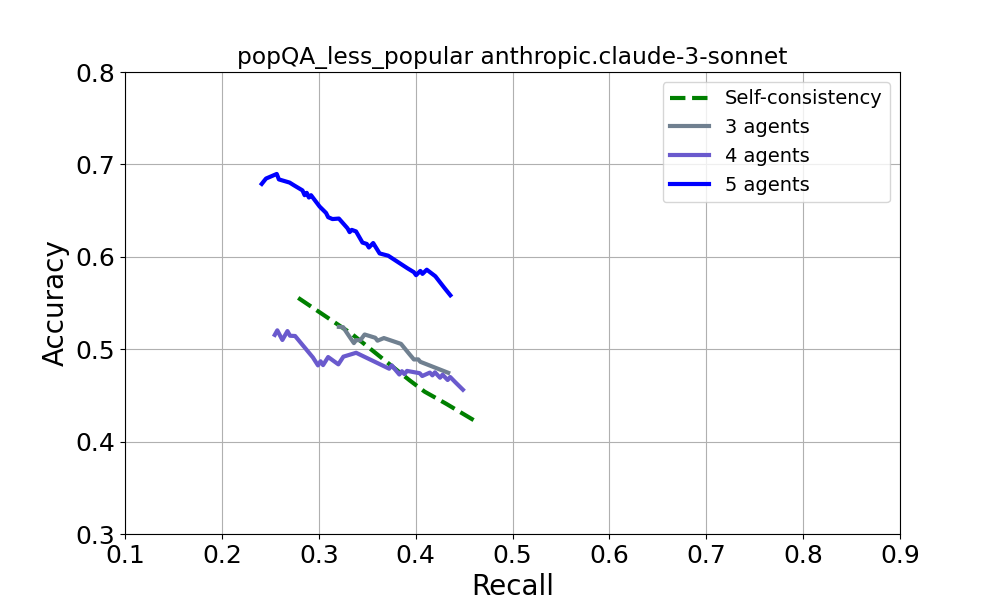}   
    \end{minipage}\hfill
     \begin{minipage}{0.45\textwidth}
        \centering
        \includegraphics[width=\linewidth]{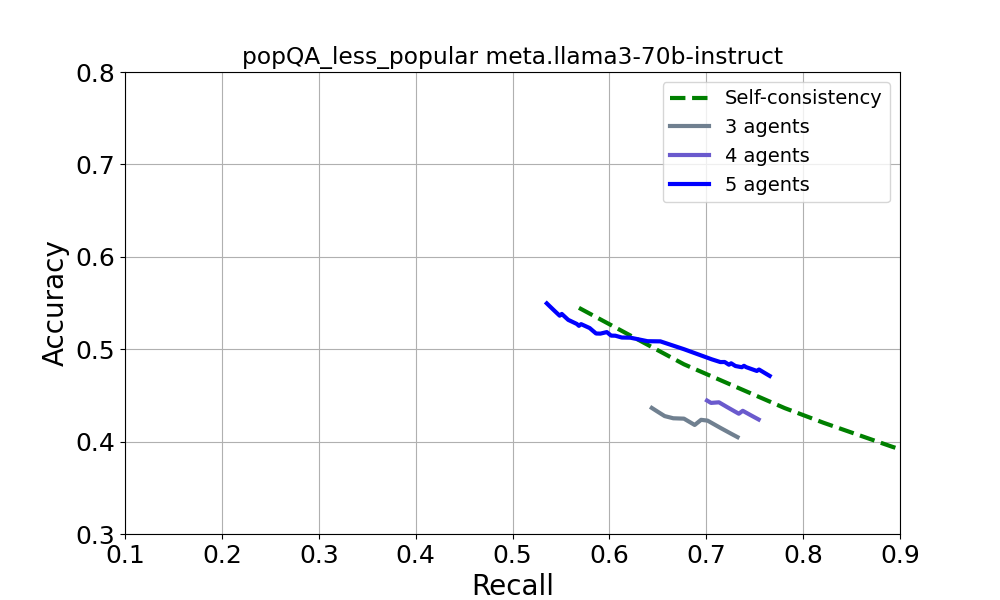}   
    \end{minipage}\hfill
    \caption{We present the effect of the number of agents on agent performance for each dataset.}
    \label{fig:appnedix_num_agent}
\end{figure*}

\begin{figure*}[ht]
    \centering
    \begin{minipage}{0.45\textwidth}
        \centering
        \includegraphics[width=\linewidth]{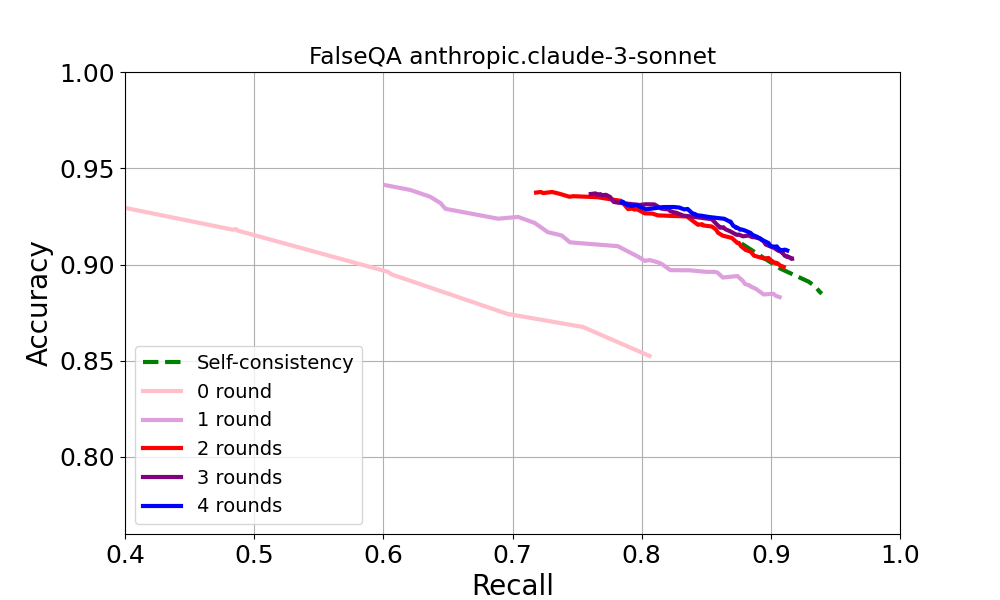}
    \end{minipage}\hfill
    \begin{minipage}{0.45\textwidth}
        \centering
        \includegraphics[width=\linewidth]{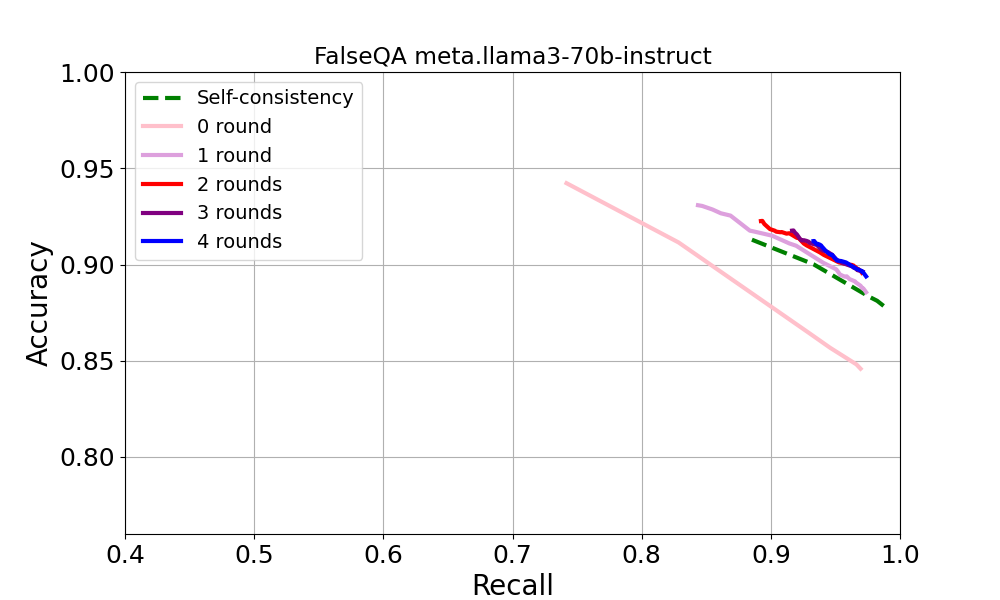}   
    \end{minipage}\hfill
     \begin{minipage}{0.45\textwidth}
        \centering
        \includegraphics[width=\linewidth]{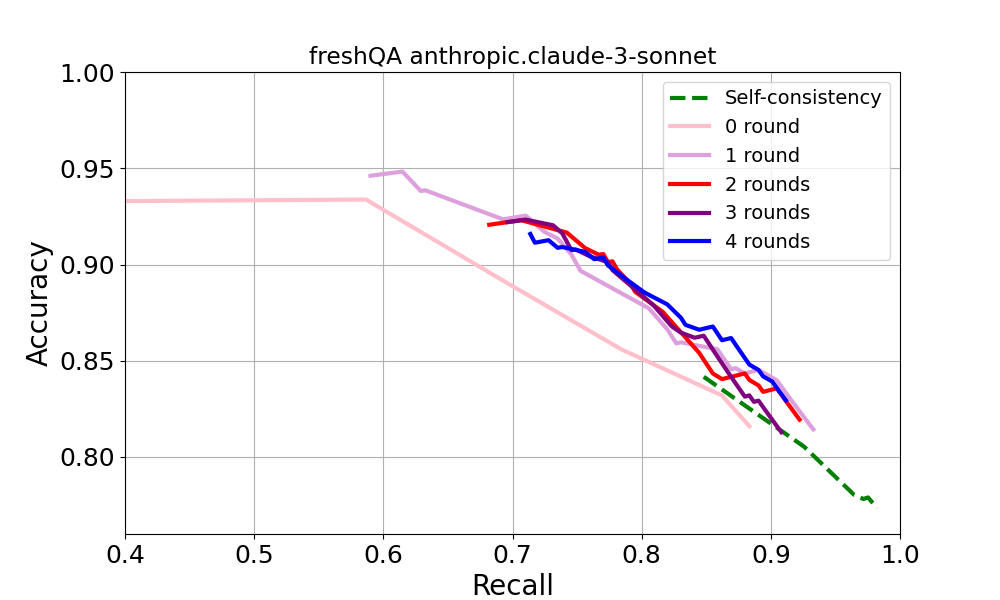}   
    \end{minipage}\hfill
     \begin{minipage}{0.45\textwidth}
        \centering
        \includegraphics[width=\linewidth]{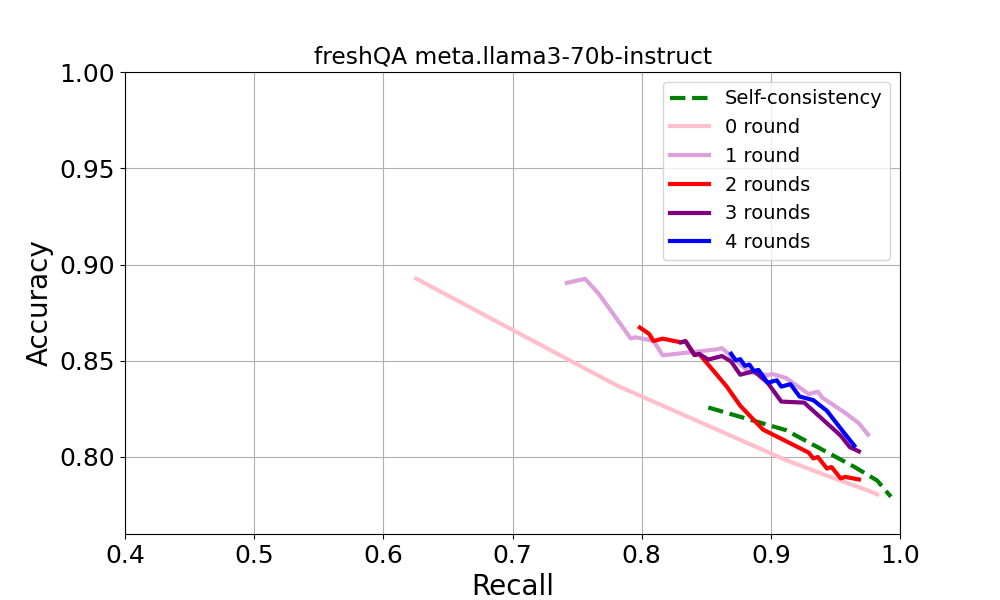}   
    \end{minipage}\hfill
    \begin{minipage}{0.45\textwidth}
        \centering
        \includegraphics[width=\linewidth]{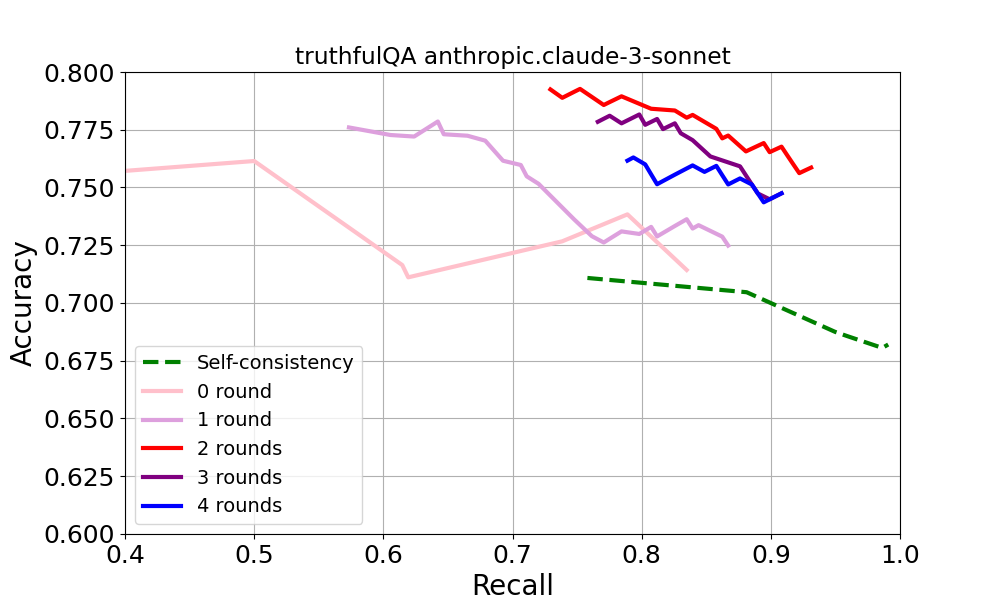}   
    \end{minipage}\hfill
     \begin{minipage}{0.45\textwidth}
        \centering
        \includegraphics[width=\linewidth]{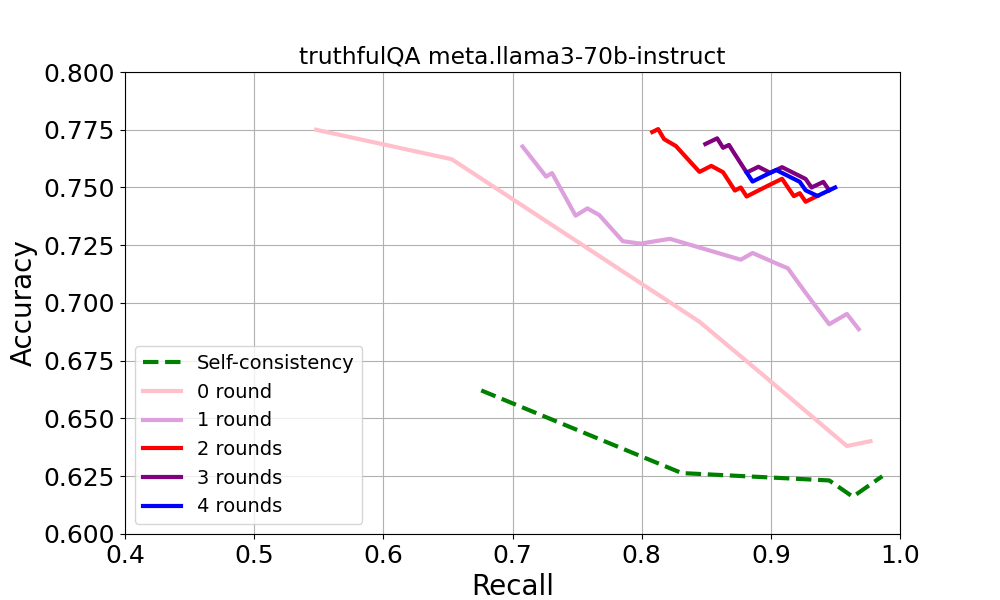}   
    \end{minipage}\hfill
    \begin{minipage}{0.45\textwidth}
        \centering
        \includegraphics[width=\linewidth]{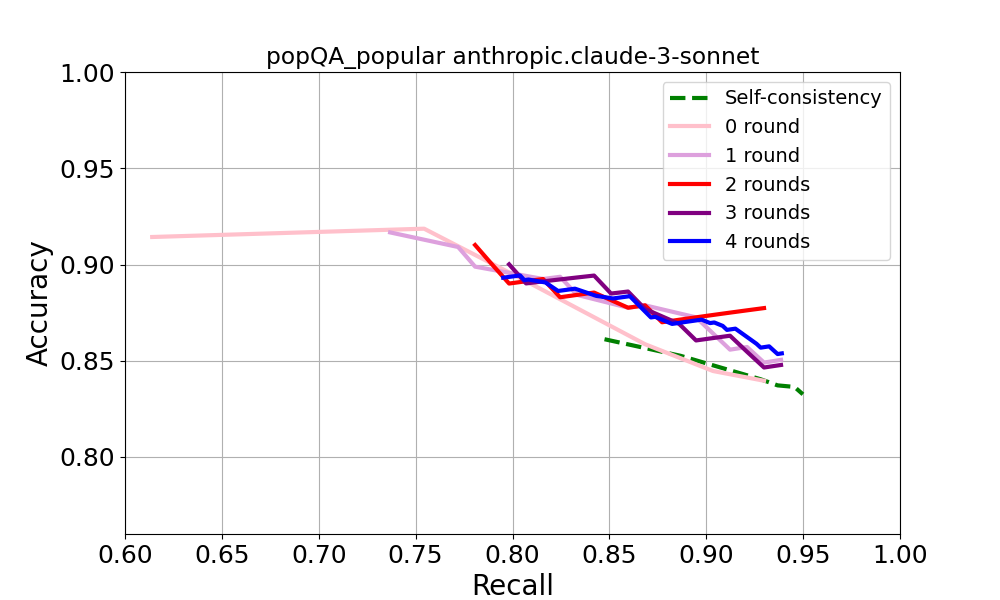}   
    \end{minipage}\hfill
     \begin{minipage}{0.45\textwidth}
        \centering
        \includegraphics[width=\linewidth]{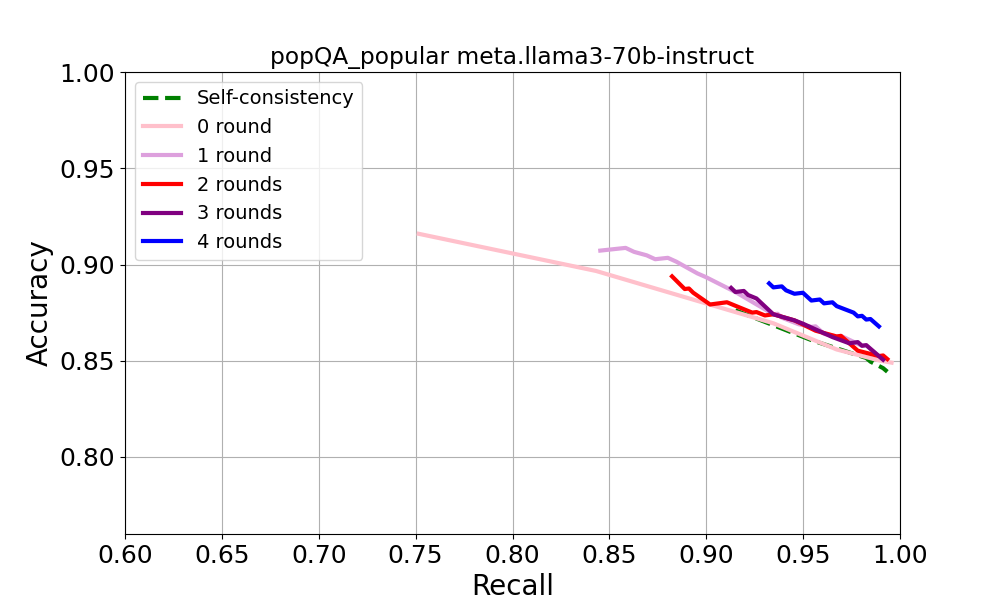}   
    \end{minipage}\hfill
    \begin{minipage}{0.45\textwidth}
        \centering
        \includegraphics[width=\linewidth]{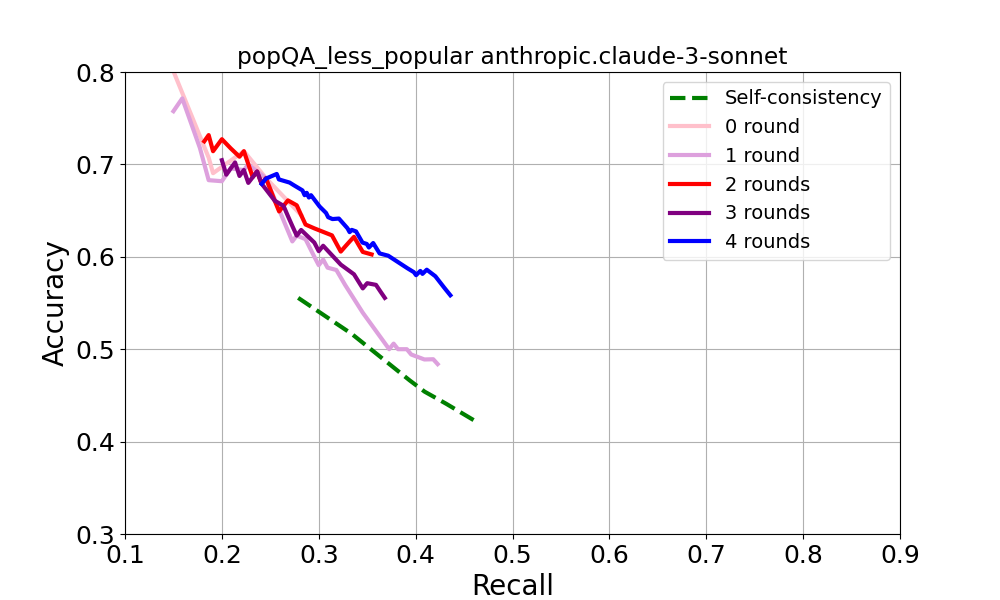}   
    \end{minipage}\hfill
     \begin{minipage}{0.45\textwidth}
        \centering
        \includegraphics[width=\linewidth]{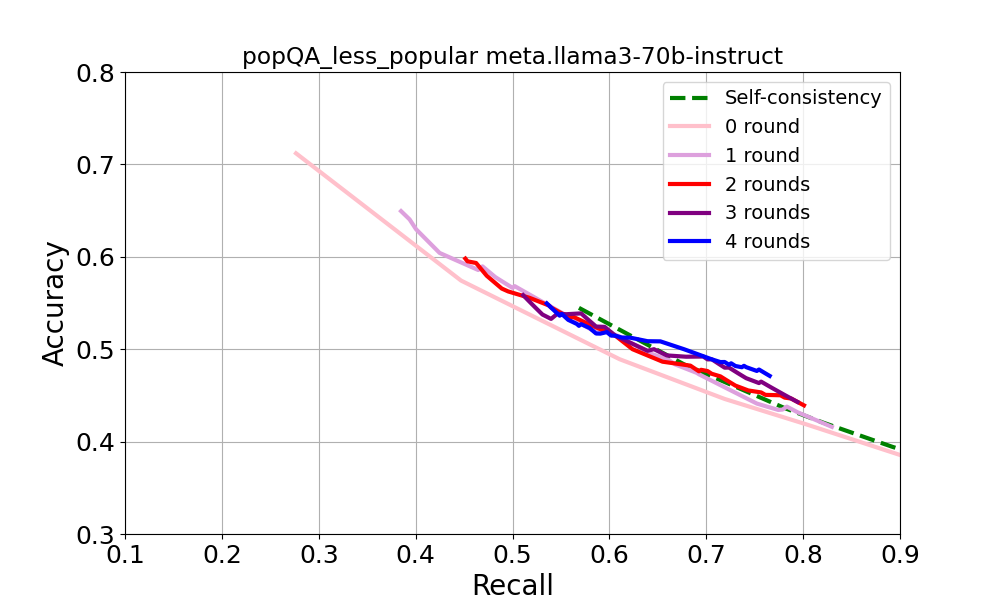}   
    \end{minipage}\hfill
    \caption{We present the effect of the number of interactions on agent performance on each dataset.}
    \label{fig:appnedix_num_round}
\end{figure*}

\begin{figure*}[ht]
    \centering
    \begin{minipage}{0.45\textwidth}
        \centering
        \includegraphics[width=\linewidth]{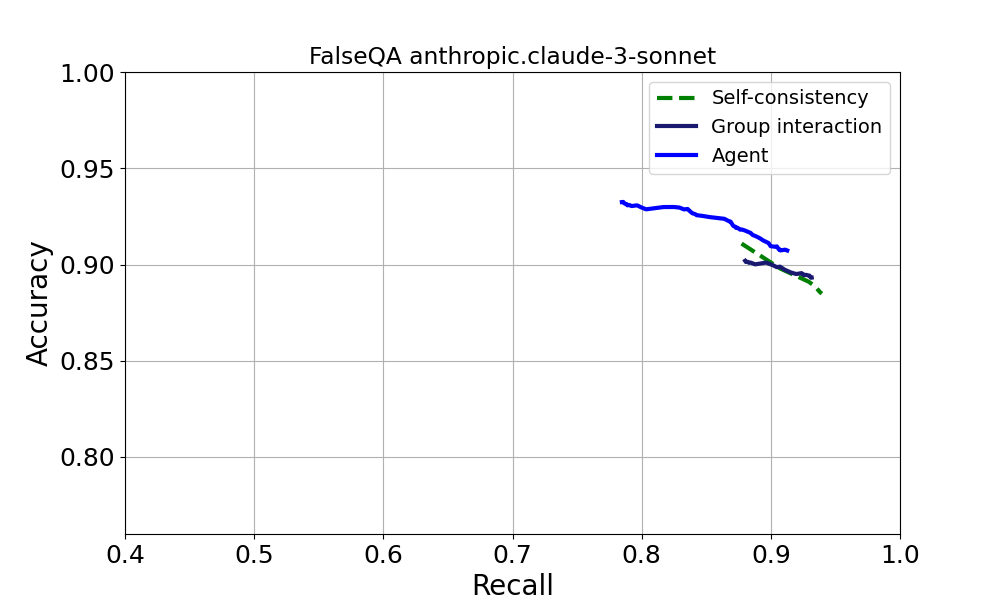}
    \end{minipage}\hfill
    \begin{minipage}{0.45\textwidth}
        \centering
        \includegraphics[width=\linewidth]{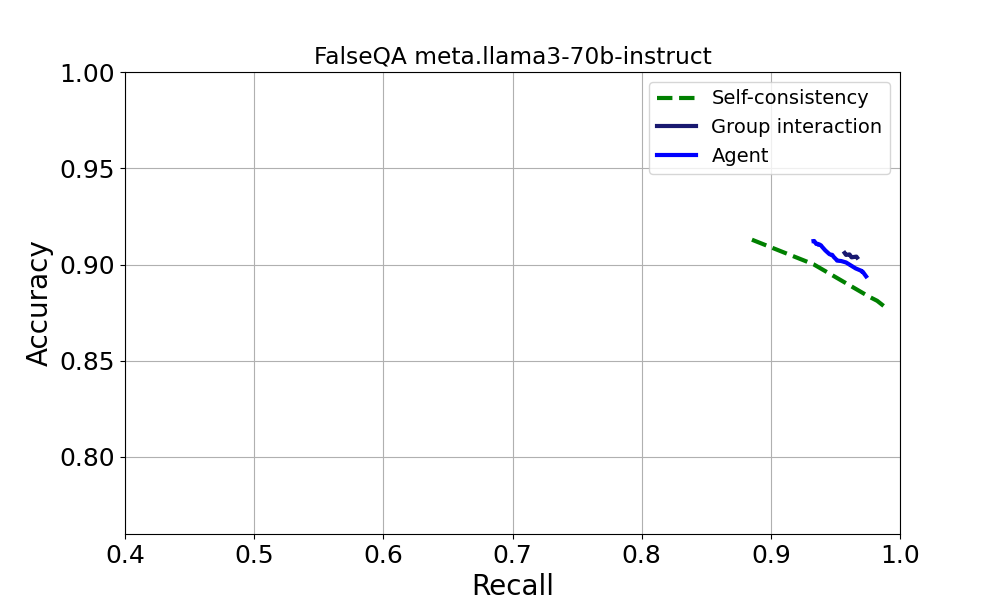}   
    \end{minipage}\hfill
     \begin{minipage}{0.45\textwidth}
        \centering
        \includegraphics[width=\linewidth]{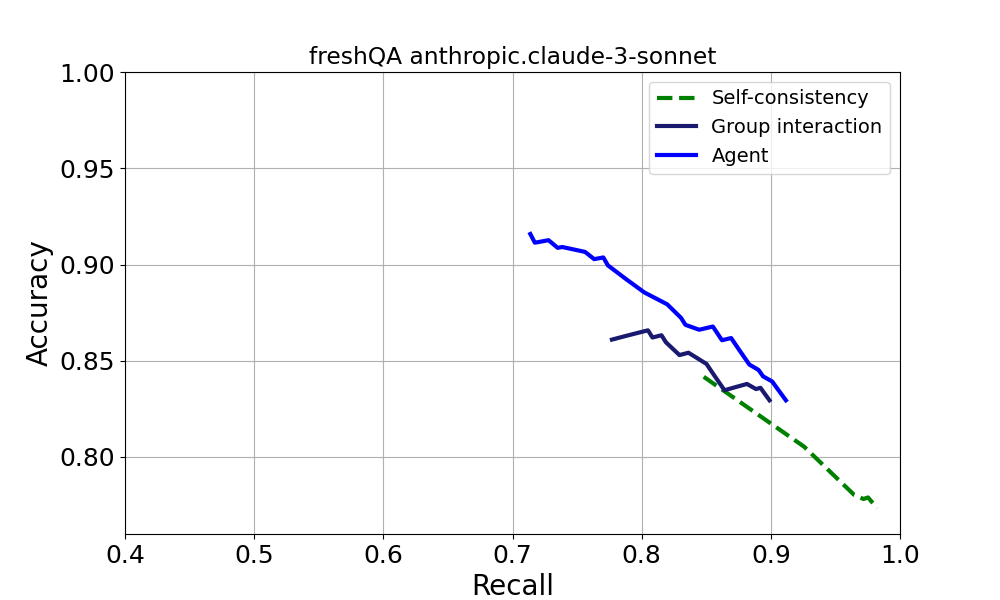}   
    \end{minipage}\hfill
     \begin{minipage}{0.45\textwidth}
        \centering
        \includegraphics[width=\linewidth]{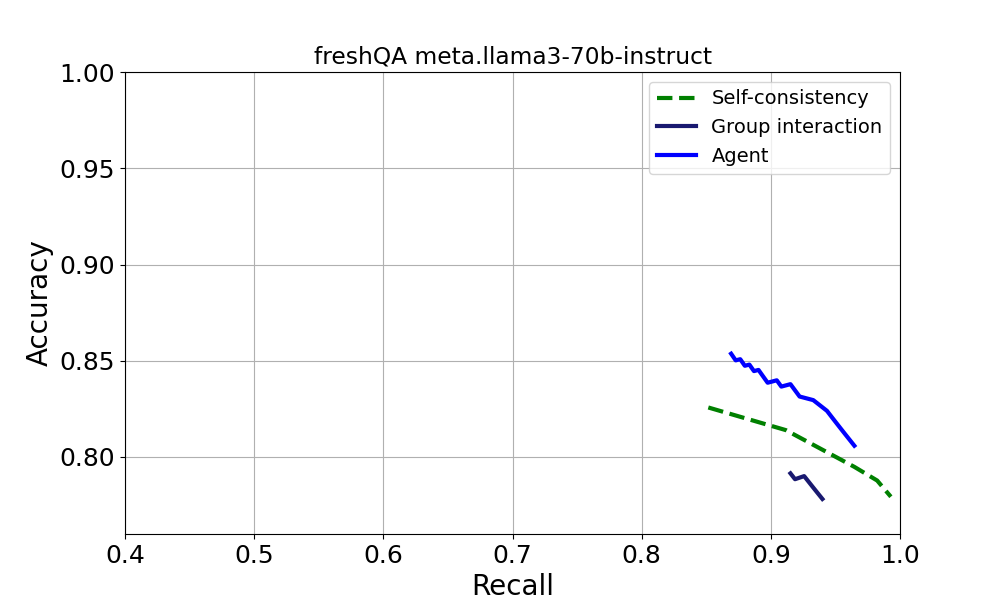}   
    \end{minipage}\hfill
    \begin{minipage}{0.45\textwidth}
        \centering
        \includegraphics[width=\linewidth]{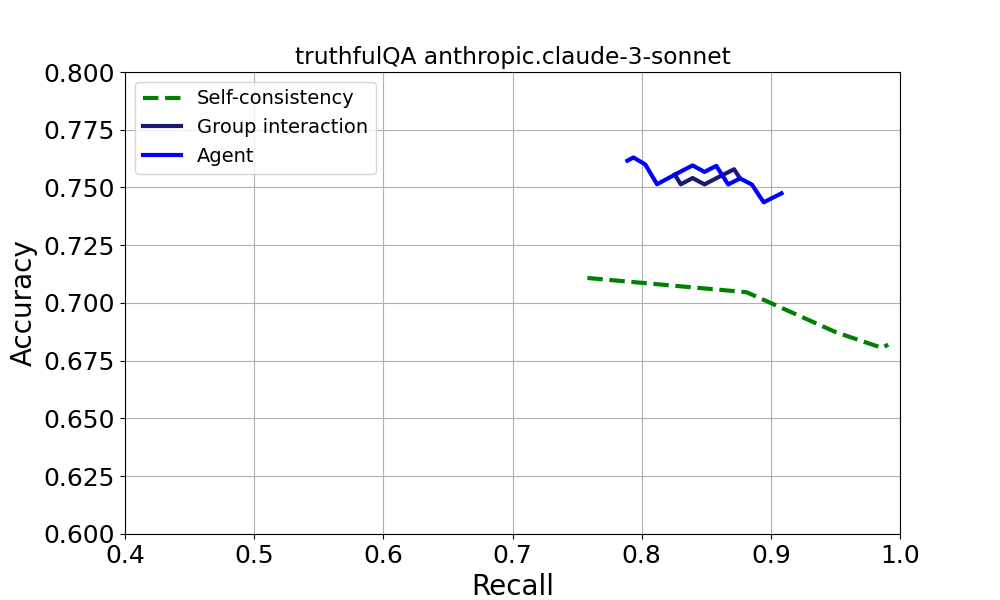}   
    \end{minipage}\hfill
     \begin{minipage}{0.45\textwidth}
        \centering
        \includegraphics[width=\linewidth]{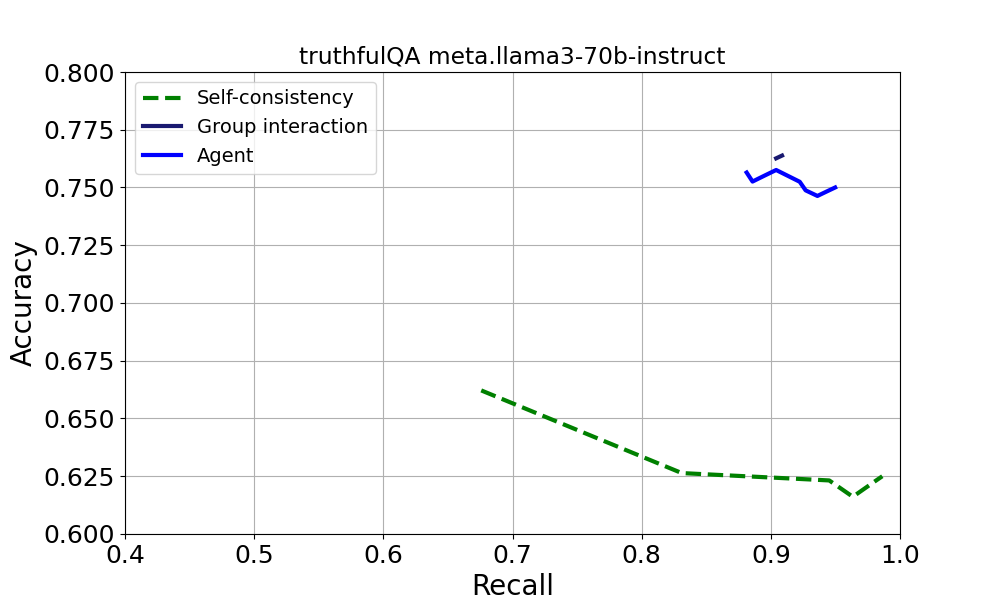}   
    \end{minipage}\hfill
    \begin{minipage}{0.45\textwidth}
        \centering
        \includegraphics[width=\linewidth]{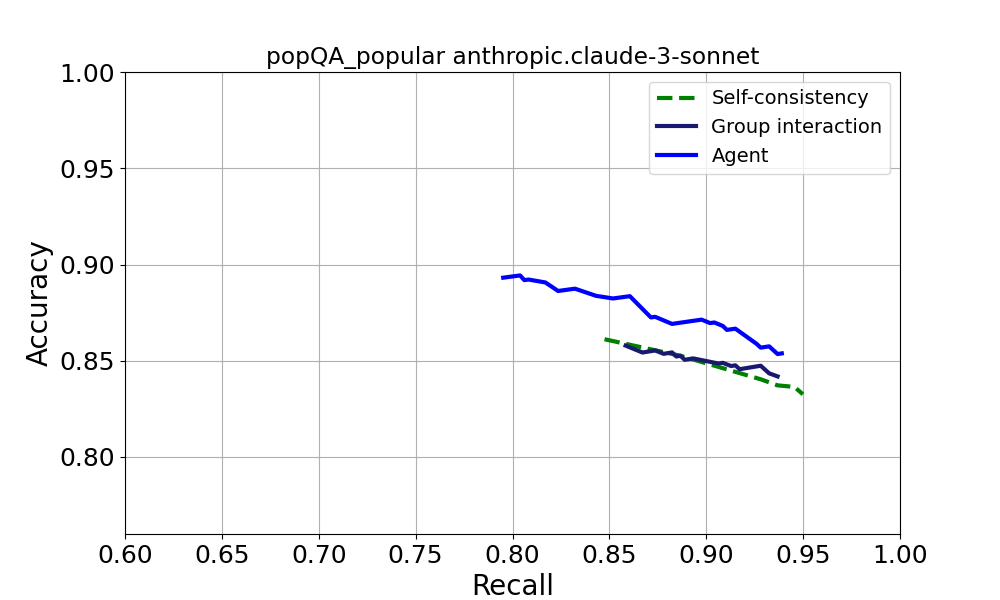}   
    \end{minipage}\hfill
     \begin{minipage}{0.45\textwidth}
        \centering
        \includegraphics[width=\linewidth]{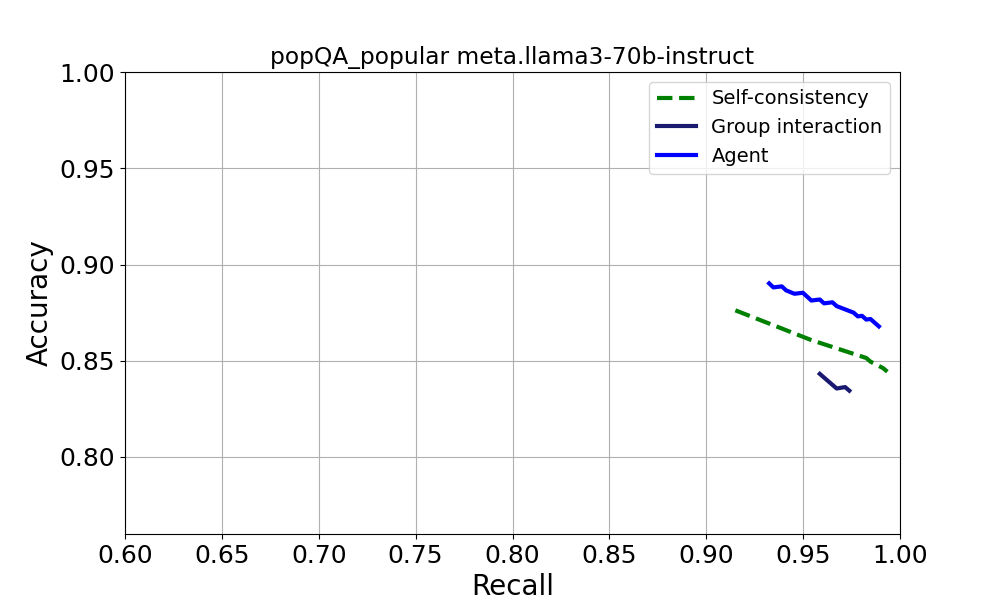}   
    \end{minipage}\hfill
    \begin{minipage}{0.45\textwidth}
        \centering
        \includegraphics[width=\linewidth]{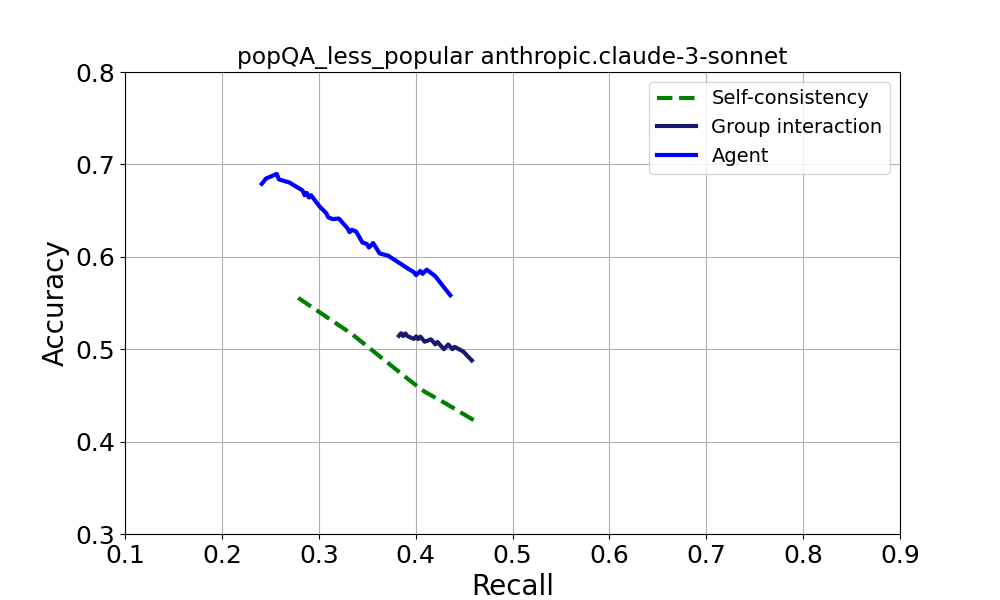}   
    \end{minipage}\hfill
     \begin{minipage}{0.45\textwidth}
        \centering
        \includegraphics[width=\linewidth]{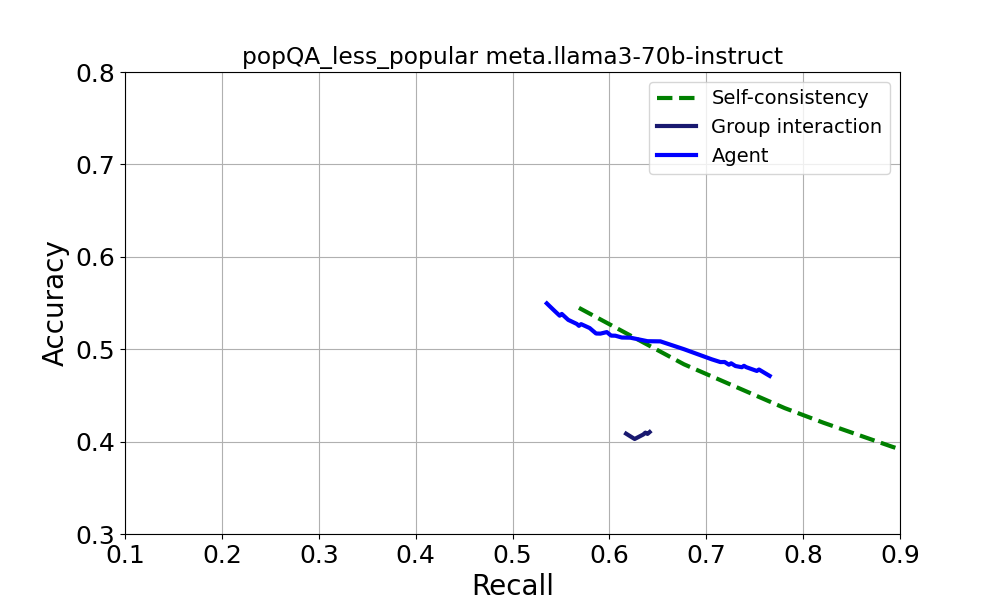}   
    \end{minipage}\hfill
    \caption{We present the effect of interaction format on each dataset.}
    \label{fig:appnedix_interaction}
\end{figure*}

\begin{figure*}[ht]
    \centering
    \begin{minipage}{0.45\textwidth}
        \centering
        \includegraphics[width=\linewidth]{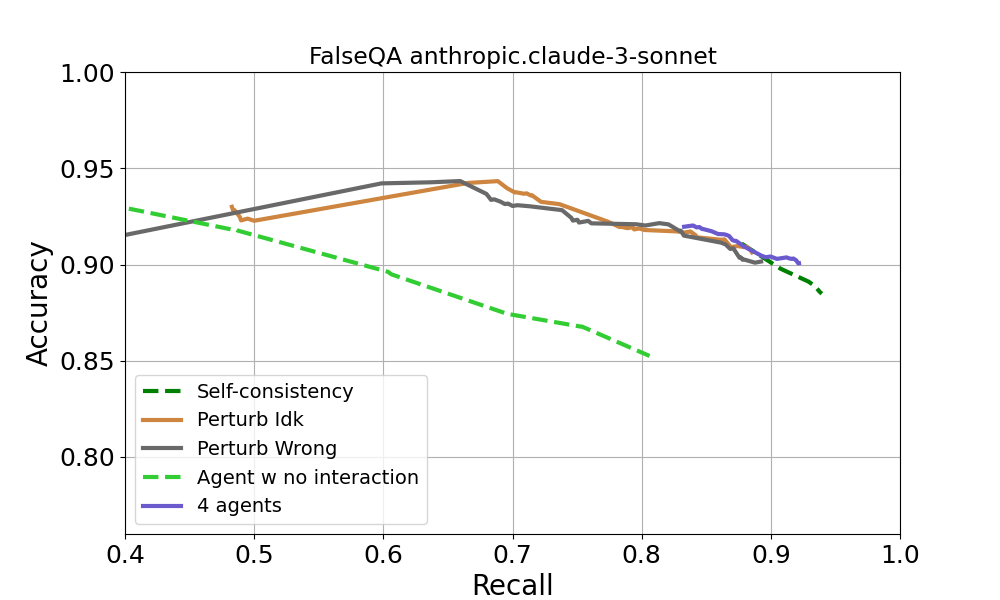}
    \end{minipage}\hfill
    \begin{minipage}{0.45\textwidth}
        \centering
        \includegraphics[width=\linewidth]{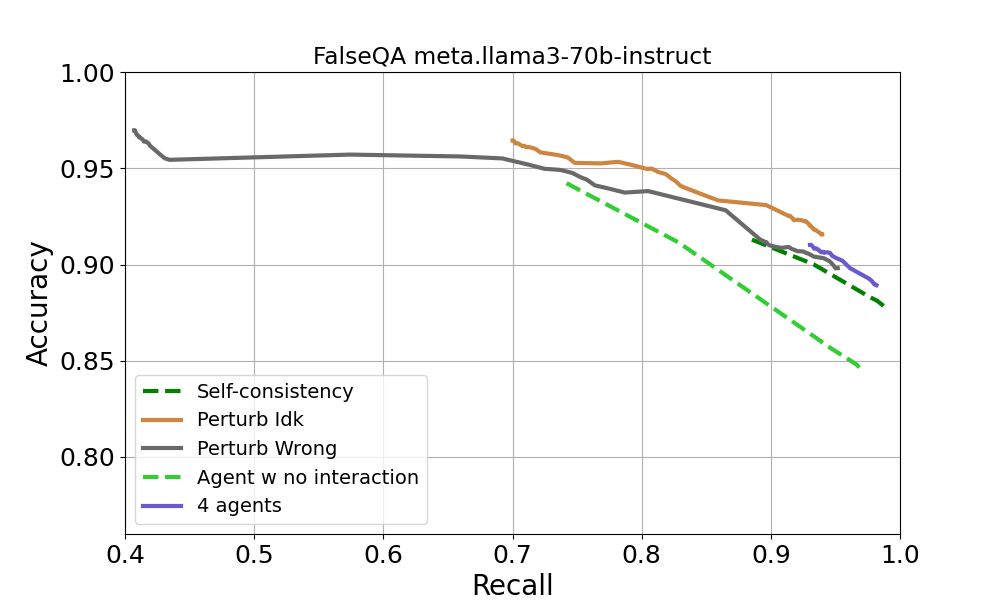}   
    \end{minipage}\hfill
     \begin{minipage}{0.45\textwidth}
        \centering
        \includegraphics[width=\linewidth]{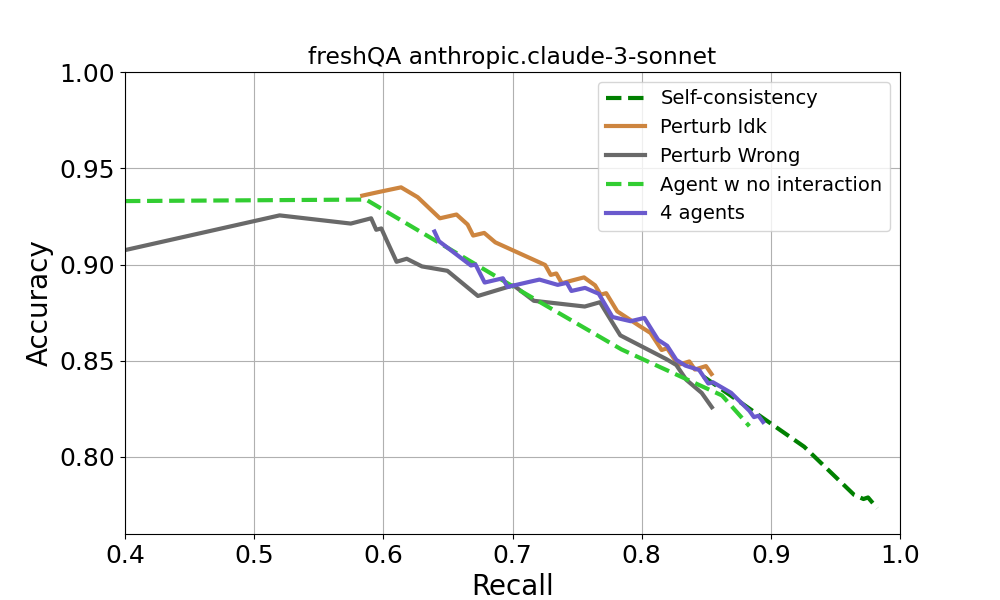}   
    \end{minipage}\hfill
     \begin{minipage}{0.45\textwidth}
        \centering
        \includegraphics[width=\linewidth]{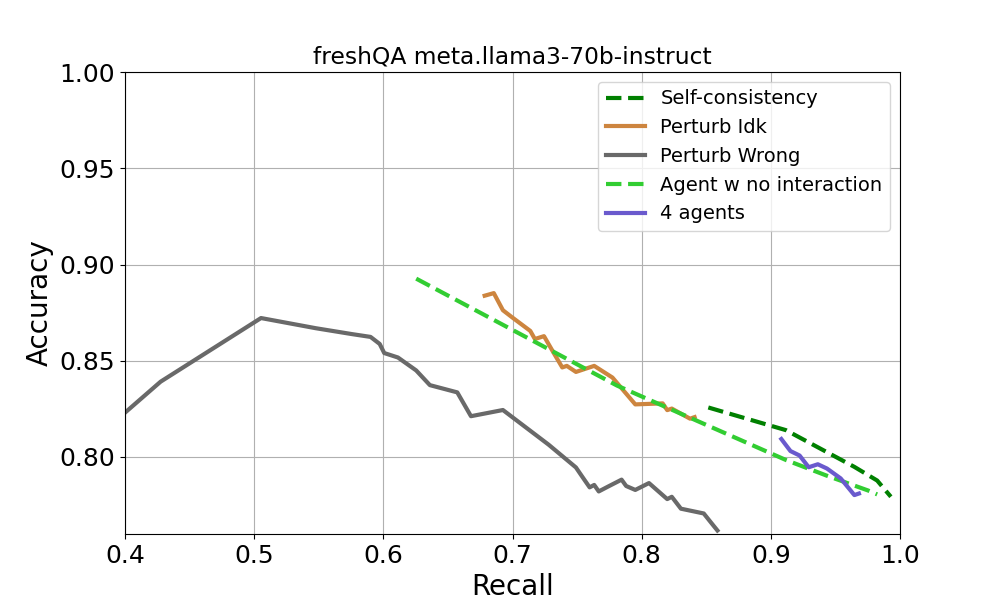}   
    \end{minipage}\hfill
    \begin{minipage}{0.45\textwidth}
        \centering
        \includegraphics[width=\linewidth]{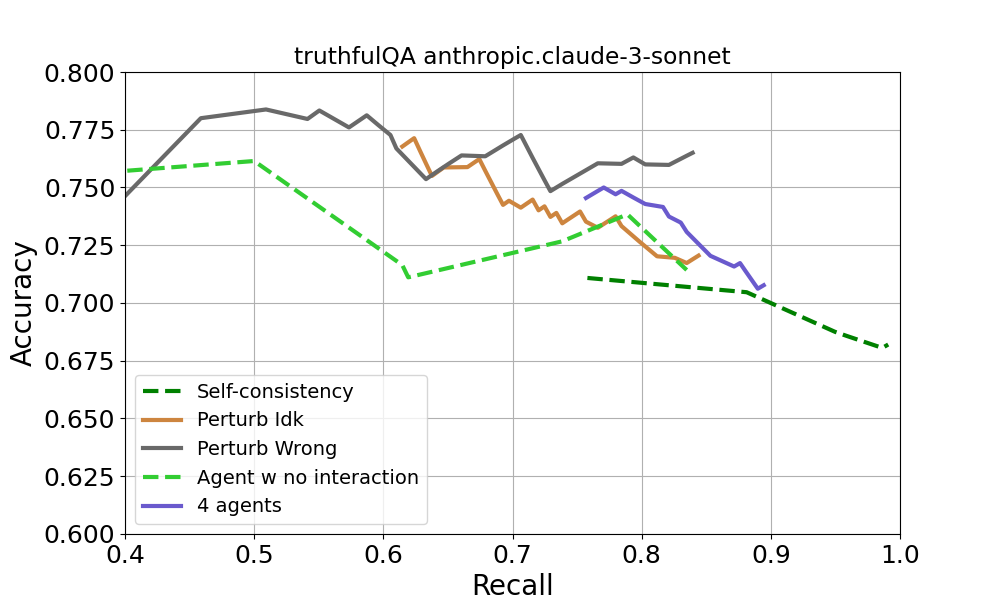}   
    \end{minipage}\hfill
     \begin{minipage}{0.45\textwidth}
        \centering
        \includegraphics[width=\linewidth]{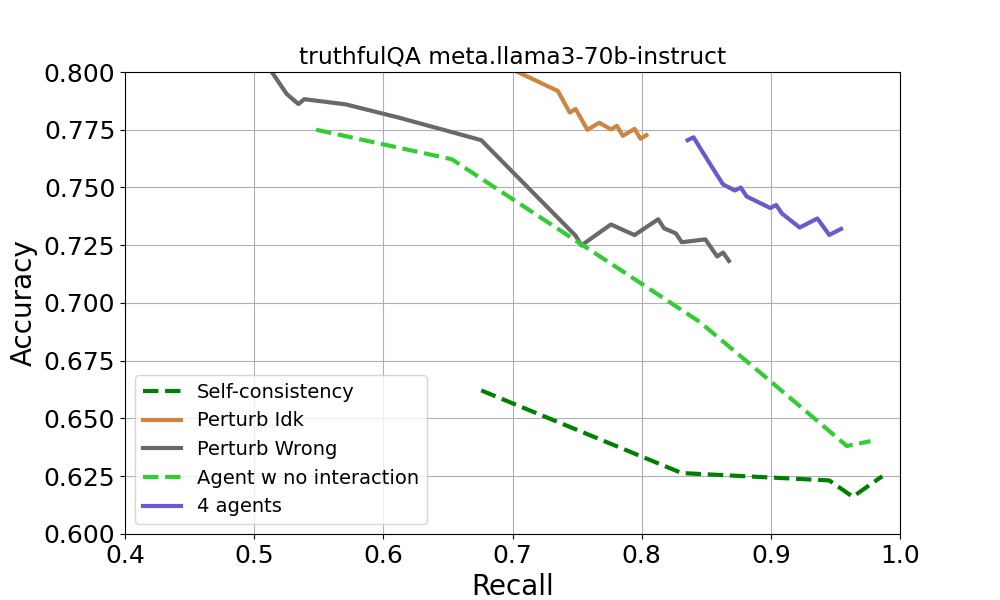}   
    \end{minipage}\hfill
    \begin{minipage}{0.45\textwidth}
        \centering
        \includegraphics[width=\linewidth]{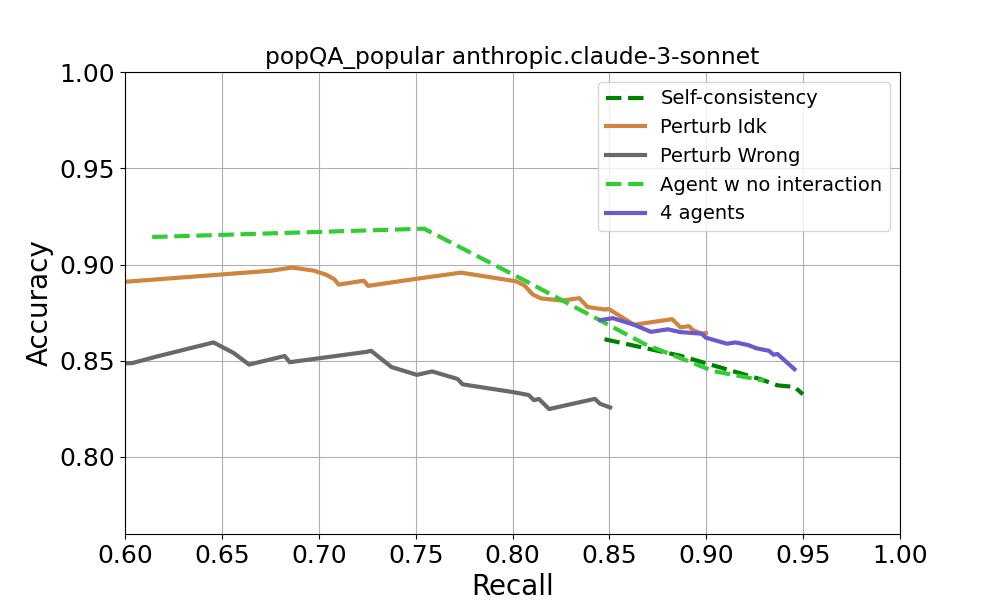}   
    \end{minipage}\hfill
     \begin{minipage}{0.45\textwidth}
        \centering
        \includegraphics[width=\linewidth]{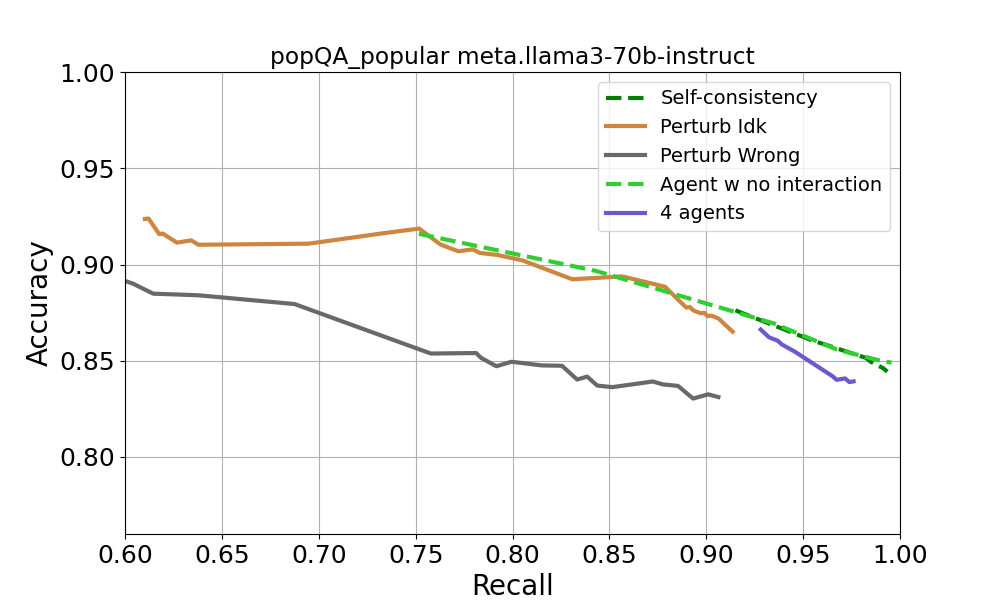}   
    \end{minipage}\hfill
    \begin{minipage}{0.45\textwidth}
        \centering
        \includegraphics[width=\linewidth]{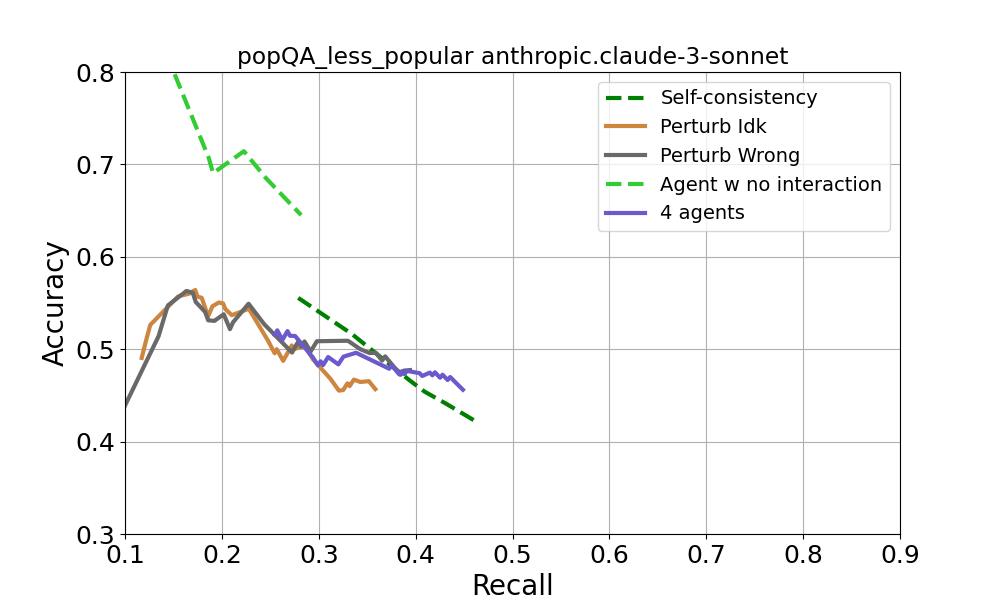}   
    \end{minipage}\hfill
     \begin{minipage}{0.45\textwidth}
        \centering
        \includegraphics[width=\linewidth]{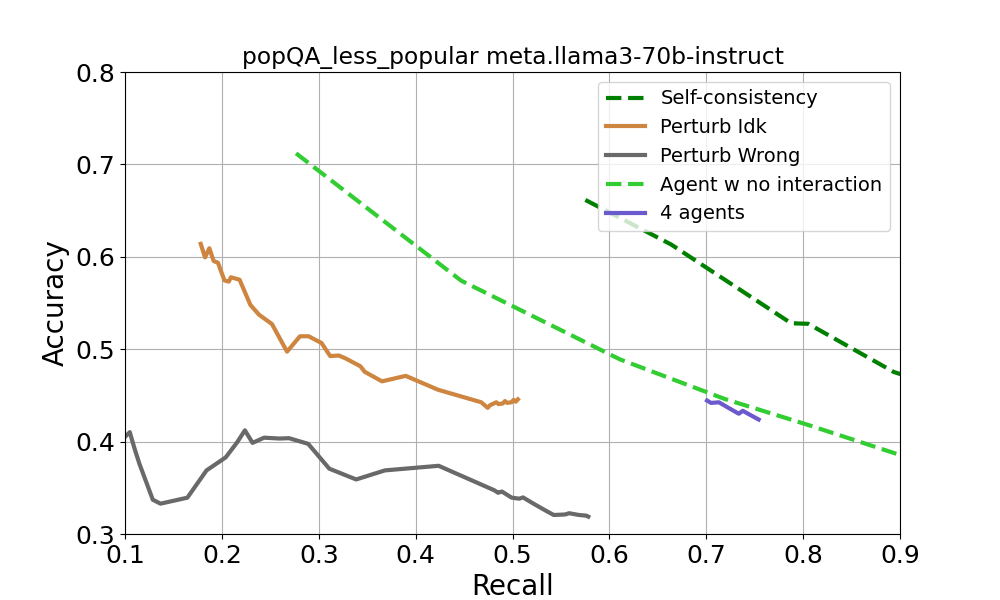}   
    \end{minipage}\hfill
    \caption{We present the robustness of interaction on each dataset.}
    \label{fig:appnedix_perturb}
\end{figure*}

\label{appendix:prompt}
\newtcolorbox{CoTAgBox}[1][]{%
  colframe=blue!60!black,  %
  colback=blue!5!white,
  title=\textsc{Example \textbf{Question Conceptualiztion} Prompt},
  #1
}

\begin{figure*}[htbp]
\centering
\begin{CoTAgBox}
\vspace{1mm}
{

\tt \small  
\textsc{\textbf{\underline{System}}}

Can you identify the broader category of the specific entity referenced in the question? 
If there is a specific entity, you MUST CHANGE it to a general category, e.g., a person, a item, a place, a object. If there is no specific entity, you MUST KEEP the original question.

\tt \small  
\textsc{\textbf{\underline{User}}}

What is the most spoken language in the world?

\tt \small  
\textsc{\textbf{\underline{Assistant}}}

What is the most spoken language in the world?

\tt \small  
\textsc{\textbf{\underline{User}}}

What is a Joe Biden's occupation?

\tt \small  
\textsc{\textbf{\underline{Assistant}}} 

What is a person's occupation?

}
\captionof{figure}{Example \textbf{Question Conceptualization} Prompt}\label{fig:prompt-cot-ag}
\vspace{0.5mm}
\end{CoTAgBox}
\end{figure*}

\newtcolorbox{CoTAgBox_1}[1][]{%
  colframe=blue!60!black,  %
  colback=blue!5!white,
  title=\textsc{Example \textbf{Aspect Generation} Prompt},
  #1
}

\begin{figure*}[]
\centering
\begin{CoTAgBox_1}
\vspace{1mm}
{

\tt \small  
\textsc{\textbf{\underline{System}}}

Can you identify up to 5 key conceptual perspectives that are as varied and diverse as possible, ensuring a comprehensive and multifaceted understanding of the question?
Given ONLY the conceptual aspect name, no other words or explanation. The aspect SHOUlD NOT indicate the answer to the question.
Each aspect is a line <as short as possible; not a complete sentence!>

\tt \small  
\textsc{\textbf{\underline{User}}}

What is the most spoken language in the world?

\tt \small  
\textsc{\textbf{\underline{Assistant}}}

demographic statistics

education policy

cultural influence

technology and media

globalization effects
}
\captionof{figure}{Example \textbf{Aspect Generation} Prompt}\label{fig:prompt-cot-ag}
\vspace{0.5mm}
\end{CoTAgBox_1}
\end{figure*}

\newtcolorbox{CoTAgBox_2}[1][]{%
  colframe=blue!60!black,  %
  colback=blue!5!white,
  title=\textsc{Example \textbf{Aspect Question Generation} Prompt},
  #1
}

\begin{figure*}[htbp]
\centering
\begin{CoTAgBox_2}
\vspace{1mm}
{

\tt \small  
\textsc{\textbf{\underline{System}}}

Generate 5 questions that build upon a given question based on a given aspect. Ensure that each question STRICTLY requires knowledge of this original question to answer but DO NOT INCLUDE the direct answer to the original question in your generated questions but MUST INCLUDE the exact content of the original question. These questions should encourage a deeper exploration of the underlying themes or concepts introduced in the original question.
Given ONLY the question, no other words or explanation.
For example:
Q1: <the generated question should be SIMPLE and DO NOT INCLUDE the direct answer to the original question>

\tt \small  
\textsc{\textbf{\underline{User}}}

Question: What is the most spoken language in the world?

Aspect: cultural influence

\tt \small  
\textsc{\textbf{\underline{Assistant}}}

Q1: How does the prevalence of the most spoken language in the world influence global media and entertainment?

Q2: In what ways has the most spoken language in the world shaped international business and trade practices?

Q3: How has the education system in non-native speaking countries adapted to teach the most spoken language in the world?

Q4: What role does the most spoken language in the world play in diplomatic relations and international policy-making?

Q5: How has the cultural heritage of the most spoken language in the world influenced global cuisine and fashion trends?
}
\captionof{figure}{Example \textbf{Aspect Question Generation} Prompt}\label{fig:prompt-cot-ag}
\vspace{0.5mm}
\end{CoTAgBox_2}
\end{figure*}

\newtcolorbox{CoTAgBox_3}[1][]{%
  colframe=blue!60!black,  %
  colback=blue!5!white,
  title=\textsc{Example \textbf{Semantically Equivalent Question Generation} Prompt},
  #1
}

\begin{figure*}[htbp]
\centering
\begin{CoTAgBox_3}
\vspace{1mm}
{

\tt \small  
\textsc{\textbf{\underline{System}}}

For the given question, provide 5 semantically equivalent questions. Do not answer the question.
STRICTLY follow the structure that each generated question is a line.

\tt \small  
\textsc{\textbf{\underline{User}}}

What is the most spoken language in the world?

\tt \small  
\textsc{\textbf{\underline{Assistant}}}

Which language has the highest number of speakers globally?

What language is spoken by most people worldwide?

Which language tops the list of the world's most widely spoken languages?

What is the world's dominant language by number of speakers?

Globally, which language is spoken by the greatest number of people?

}
\captionof{figure}{Example \textbf{Semantically Equivalent Question Generation} Prompt}\label{fig:prompt-cot-ag}
\vspace{0.5mm}
\end{CoTAgBox_3}
\end{figure*}

\newtcolorbox{CoTAgBox_4}[1][]{%
  colframe=blue!60!black,  %
  colback=blue!5!white,
  title=\textsc{Example \textbf{1-1 Interaction} Prompt},
  #1
}

\begin{figure*}[htbp]
\centering
\begin{CoTAgBox_4}
\vspace{1mm}
{

\tt \small  
\textsc{\textbf{\underline{System}}}

You are an AI assistant that helps people answer questions. Ensure your responses are concise and strictly relevant to the queries presented, avoiding any unrelated content to the question. Do not change your answer unless you think you are absolutely wrong.

<previous interaction conversations…>

\tt \small  
\textsc{\textbf{\underline{User}}}

 When I asked you in another api call that + selection\_agent\_question + You mentioned that + selection\_agent\_answer\_to\_original\_query + Which is your actual answer to + original\_query?

}
\captionof{figure}{Example \textbf{1-1 Interaction} Prompt}\label{fig:prompt-cot-ag}
\vspace{0.5mm}
\end{CoTAgBox_4}
\end{figure*}

\newtcolorbox{CoTAgBox_5}[1][]{%
  colframe=blue!60!black,  %
  colback=blue!5!white,
  title=\textsc{Example \textbf{Answer Extraction} Prompt},
  #1
}

\begin{figure*}[htbp]
\centering
\begin{CoTAgBox_5}
\vspace{1mm}
{

\tt \small  
\textsc{\textbf{\underline{System}}}

You will extract the answer to the given question using ONLY the information provided in the "Response" section. You will identify the answer directly without using any additional knowledge or explanation. If the response includes a negation to the question, use those as the answer.

\tt \small  
\textsc{\textbf{\underline{User}}}

Response: The prevalence of the most spoken language in the world, which is Mandarin Chinese, has a significant influence on global media and entertainment in several ways:

                1. **Content creation**: Many Chinese production companies and studios create content specifically for the massive Chinese-speaking audience, which often gets distributed globally. This leads to a increase in Chinese-language content in international markets.
                2. **Dubbing and subtitles**: To cater to the large Chinese-speaking population, many international films and TV shows are dubbed or subtitled in Mandarin, making them more accessible to Chinese audiences.
                3. **Global market appeal**: The massive Chinese market has become a crucial factor in the success of global films, TV shows, and music. Creators often tailor their content to appeal to Chinese audiences, incorporating Chinese themes, actors, or storylines.
                4. **Influence on global trends**: Chinese social media platforms, such as WeChat and Weibo, have become essential channels for promoting global entertainment content. Trends and memes that originate in China can quickly spread globally.
                5. **Cultural exchange**: The prevalence of Mandarin Chinese has facilitated cultural exchange between China and other countries. International artists and creators are increasingly collaborating with Chinese counterparts, leading to a fusion of cultural influences in media and entertainment.

                Overall, the dominance of Mandarin Chinese has reshaped the global media and entertainment landscape, with creators and distributors adapting to cater to the vast and influential Chinese-speaking audience.
        
Based solely on the response, What is the most spoken language in the world?

\tt \small  
\textsc{\textbf{\underline{Assistant}}}

The most spoken language in the world is Mandarin Chinese.

}
\captionof{figure}{Example \textbf{Answer Extraction} Prompt}\label{fig:prompt-cot-ag}
\vspace{0.5mm}
\end{CoTAgBox_5}
\end{figure*}

\newtcolorbox{CoTAgBox_6}[1][]{%
  colframe=blue!60!black,  %
  colback=blue!5!white,
  title=\textsc{Prompt for Analysis of Diverse Queries - Knowledge Preservation},
  #1
}

\begin{figure*}[htbp]
\centering
\begin{CoTAgBox_6}
\vspace{1mm}
{

\tt \small  
\textsc{\textbf{\underline{System}}}

You are evaluating whether a generated query preserves the knowledge required to answer a target query.

Definition: A query preserves knowledge if answering it requires the same knowledge as answering the original query, even if it is phrased differently.

Return your output in the following JSON format:
\{
  "explanation": "<brief explanation of whether and why the generated query preserves the knowledge required to answer the target query>",
  "valid": <true | false>
\}

\tt \small  
\textsc{\textbf{\underline{User}}}

Target Query: "<INSERT TARGET QUERY HERE>"

Generated Query: "<INSERT GENERATED QUERY HERE>"

}
\captionof{figure}{Prompt for Analysis of Diverse Queries - Knowledge Preservation}\label{fig:prompt-ana-k}
\vspace{0.5mm}
\end{CoTAgBox_6}
\end{figure*}

\newtcolorbox{CoTAgBox_7}[1][]{%
  colframe=blue!60!black,  %
  colback=blue!5!white,
  title=\textsc{Prompt for Analysis of Diverse Queries - Diversity},
  #1
}

\begin{figure*}[htbp]
\centering
\begin{CoTAgBox_7}
\vspace{1mm}
{

\tt \small  
\textsc{\textbf{\underline{System}}}

You are evaluating whether two queries are sufficiently diverse in their surface.

Definition: Two queries meet the diversity criterion if they differ in at least 20\% of their wording (measured by word variation).

Return your output in the following JSON format:
{
  "explanation": "<brief explanation of whether and why the two queries are semantically equivalent and differ by at least 20\% in wording>",
  "valid": <true | false>
}

\tt \small  
\textsc{\textbf{\underline{User}}}

Query 1: "<INSERT QUERY 1 HERE>"

Query 2: "<INSERT QUERY 2 HERE>"

}
\captionof{figure}{Prompt for Analysis of Diverse Queries - Diversity}\label{fig:prompt-ana-d}
\vspace{0.5mm}
\end{CoTAgBox_7}
\end{figure*}

\end{document}